\documentclass[conf]{new-aiaa}  % for conference papers
\usepackage[utf8]{inputenc}

\usepackage{graphicx}
\usepackage{amsmath}
\usepackage[version=4]{mhchem}
\usepackage{siunitx}
\usepackage{longtable, tabularx}
\usepackage{color}
\usepackage{algorithm}
\usepackage[noend]{algpseudocode}
\usepackage{soul}

\newcommand{\mtx}[0]{\mathbf}  % Matrix
\newcommand{\vc}[0]{\boldsymbol}  % Vector
\newcommand{\mc}[0]{\mathcal}
\newcommand{\reals}[0]{\mathbb R}
\newcommand{\transpose}[0]{\intercal}

\newcommand{\expectation}[2]{\langle #2 \rangle_{#1}}

\DeclareMathOperator*{\argmin}{arg\,min}

\newcommand{\scriptD}[0]{{\mathcal D}}

\newcommand{\scriptM}[0]{{\mathcal M}}
\newcommand{\scriptX}[0]{{\mathcal X}}

\newcommand{\G}[3]{\mathcal N ({#1} | {#2}, {#3})}

\newcommand{\kff}[0]{\mtx K_{ff}}
\newcommand{\kyy}[0]{\mtx K_{yy}}
\newcommand{\kyyinv}[0]{\mtx K_{yy}^{-1}}
\newcommand{\kss}[0]{\mtx K_{**}}
\newcommand{\ksf}[0]{\mtx K_{*f}}
\newcommand{\kfs}[0]{\mtx K_{f*}}

\title{Bayesian task embedding for few-shot Bayesian optimization}

\author{
	Steven Atkinson\footnote{Research Engineer, Mechanical Systems, GE Research, Niskayuna, New York.},
	Sayan Ghosh\footnote{Lead Engineer, Mechanical Systems, GE Research, Niskayuna, New York.},
	Natarajan Chennimalai-Kumar\footnote{Senior Engineer, Mechanical Systems, GE Research, Niskayuna, New York.},
	Genghis Khan\footnote{Senior Principal Engineer, Mechanical Systems, GE Research, Niskayuna, New York.},
	and
	Liping Wang\footnote{Technology Manager, Mechanical Systems, GE Research, Niskayuna, New York.}
}
\affil{GE Research, 1 Research Circle, Niskayuna, NY, 12309, USA}

\begin{document}

\maketitle

\begin{abstract}
We describe a method for Bayesian optimization by which one may incorporate data from multiple systems whose quantitative interrelationships are unknown \textit{a priori}.
All general (non-real-valued) features of the systems are associated with continuous latent variables that enter as inputs into a single metamodel that simultaneously learns the response surfaces of all of the systems.
Bayesian inference is used to determine appropriate beliefs regarding the latent variables.
We explain how the resulting probabilistic metamodel may be used for Bayesian optimization tasks and demonstrate its implementation on a variety of synthetic and real-world examples, comparing its performance under zero-, one-, and few-shot settings against traditional Bayesian optimization, which usually requires substantially more data from the system of interest.
\end{abstract}

\section{Introduction}
\label{sec:introduction}
Many engineering design problems can be cast as optimization problems of the form
\begin{equation}
	\vc x^* = \argmin_{\vc x \in \mc X} \eta(\vc x),
	\label{eqn:optimization}
\end{equation}
where $\vc x \in \mc X$ is a $d_x$-dimensional vector of design and operation parameters, and $\eta$ is some quantity of interest, such as the strength of a material or the efficiency of an engine.
Solving optimization problems that fit under the scope of Eq.\ (\ref{eqn:optimization}) can be challenging if the dimension $d_x$ of the search space is high and/or the function $\eta$ is nonconvex and highly rugged, resulting in an intractably large number of potential solutions that may be considered.
Furthermore, if it is expensive to evaluate $\eta$ (corresponding to performing a laboratory experiment or running a complex simulation code), one may be limited by their experimental budget and be forced to try and determine a good solution to Eq.\ (\ref{eqn:optimization}) with few opportunities to gain information about the true response surface for the system of interest.

The most elementary approaches for solving Eq.\ (\ref{eqn:optimization}) involve specifying some space-filling design and querying $\eta$ at all of the design points.
While rudimentary designs such as factorial designs suffer from the curse of dimensionality and are intractable for all but the simplest problems, more advanced space-filling designs such as Latin hypercube designs still may require far too many evaluations than is feasible.
Improving on these by orders of magnitude, in Bayesian optimization (BO), one defines a Bayesian metamodel $y(\vc x)$ [such as a Gaussian process (GP)] to approximate $\eta$ and uses it to adaptively select promising designs until the optimum is found or an experimental budget is exhausted.
Still, such an approach commonly takes tens to hundreds of evaluations to converge to a satisfactory degree in practice.
In this work, we are interested in improving on this by another order of magnitude, finding plausible solutions to Eq.\ (\ref{eqn:optimization}) with a handful--or \textit{no}---evaluations of $\eta$.

Humans intuitively solve optimization problems in their daily lives in novel settings by leveraging knowledge about previous related experiences.
By contrast, it is common to start quantitative engineering design problems from scratch since it is unclear how to re-use data from legacy systems with different response surfaces.
There is a growing literature around methods that seek to devise suitable models to bridge the gap between these disparate sources of information.
Through this, we hope to leverage so-called ``legacy'' data about previously seen tasks in a manner that endows our current effort with strong, but reasonable inductive biases that guide it towards effective solutions.

The main technical challenge is to devise a means of reasoning in a quantitative way about features of data that are not numerical in nature and therefore not suitable for standard modeling approaches\footnote{Some literature refer to these as ``qualitative'' features, though this seems somewhat of a misnomer since certain types of attributes in question can be numerical in nature, such as zip codes, yet are clearly unsuitable to treat as numbers in a model; others may not be quantitative, but are nonetheless precise (e.g.\ the name of an operator), yet ``qualitative'' does not convey this preciseness.}
These ``general'' features are typically found when describing the difference between tasks \cite{argyriou2007multi} and could include, for example, the serial number of a machine, the identity of an operator, or a chemical compound involved in some process of interest.

The key to our approach is to learn a probabilistic embedding associated with the general features associated with a system such that notions of similarity can be quantified and utilized by a downstream data-driven model.
Minding our ultimate goal of solving optimization problems, we focus on Gaussian process metamodels and call the composition of our probabilistic embedding with the Gaussian process metamodel ``Bayesian embedding GP'' (BEGP).
The contributions of this work are the following:
\begin{enumerate}
	\item We define the structure of the BEGP metamodel designed to fuse information from systems with differing general features.  
	We use a variational inference scheme to learn to infer reasonable probabilistic embeddings of the general features that capture uncertainty due to limited data while showing that the compositional model can be recognized as a deep Gaussian process \cite{damianou2013deep} with a particular choice of kernel function.
	\item We explain and demonstrate the application of this model to the task of Bayesian optimization, showing how the BEGP can be used to satisfy the usual requirements of the algorithm.
	\item We conduct a series of computational experiments on a variety of synthetic and real-world systems to illustrate the usage of our approach and compare its performance to existing methods and evaluating the contribution of various components of the metamodel.
\end{enumerate}

The scope of our work here is optimization problems of the form in Eq.\ (\ref{eqn:optimization}).
However, because our approach can be used as a drop-in replacement for other Bayesian metamodels, it is straightforward to extend our work to cases including multi-objective optimization and problems involving complex or unknown constraints.
We also consider regression tasks as a stepping stone to our ultimate application of interest since satisfactory predictive power is a desired preliminary skill.

The remainder of this paper is organized as follows:
In section \ref{sec:methodology}, we explain the formulation of our metamodel and its usage within Bayesian optimization
In section \ref{sec:related_work}, we review related approaches and results from the literature.
In section \ref{sec:examples}, we demonstrate our approach on a variety of synthetic and real-world examples.
In section \ref{sec:conclusion}, we offer concluding remarks.

\section{Methodology}
\label{sec:methodology}
In this section, we describe the methodology including our model definition, training objective, predictive distribution, and application to Bayesian optimization.

\subsection{Model definition}
We begin the discussion of our methodology by defining our model, explaining how inference may be done to determine its posterior, and how predictions are computed.

\subsubsection{Gaussian process regression}
We begin by reviewing Gaussian processes for regression; the interested reader may consult \cite{rasmussen2006gaussian} for more details.
A Gaussian process (GP) $\mc{GP}(\mu(\cdot), k(\cdot, \cdot))$ with mean function $\mu$ and kernel $k$ is a distribution over functions such that the marginal distribution of the random function over a finite index set is a multivariate Gaussian.
Concretely, let $f: \mc X \rightarrow \reals$ be described by a GP where $\mc X$ is an input space that indexes the random process; traditionally, this will be $d_x$-dimensional Euclidean space $\reals^{d_x}$ or some subset therein.
However, any (infinite) set of inputs might be used, and we are mindful of this potential generality in the review in this section.
Given $n$ inputs $\mtx X \in \scriptX^n$ with corresponding outputs $\vc f \in \reals^n$, we write the joint probability density of $\vc f$ as
\begin{equation}
	p(\vc f | \mtx X) = \G{\vc f}{\vc \mu}{\kff},
	\label{eqn:pf}
\end{equation}
where $\mu_i = \mu(\vc x_i)$, and $(\kff)_{ij} = k(\vc x_i, \vc x_j)$.
The quantity $\vc f$ is recognized as the latent output of the GP model.
Next, we define a Gaussian likelihood 
\begin{equation}
	p(\vc y|\vc f) = \G{\vc y}{\vc f}{\sigma_y^2 \mtx I_{n \times n}},
	\label{eqn:likelihood}
\end{equation}
where $\vc y \in \reals^n$ denotes the observed output values.
Integrating out $\vc f$ results in the familiar marginal likelihood
\begin{equation}
	p(\vc y | \mtx X) = \G{\vc y}{\vc \mu}{\kyy},
	\label{eqn:gp:marginal_likelihood}
\end{equation}
where $\kyy = \kff + \sigma_y^2 \mtx I_{n \times n}$.
The negative logarithm of this quantity is conventionally used to determine appropriate parameters for the mean and kernel functions as well as the likelihood through gradient-based minimization. 
Alternatively, approximate Bayesian inference of the model parameters may be done as well using Markov chain Monte Carlo or variational inference once suitable priors are defined \cite{bilionis2013multi}.

Given a training set $\scriptD = \{\mtx X, \vc y\} \in \scriptX^n \times \reals^n$, predictions are made by using Bayes' rule to condition the GP on the training set.
The resulting predictive distribution over the latent outputs $\vc f^* \in \reals^{n^*}$ at some test inputs $\mtx X^* \in \scriptX^{n^*}$ is
\begin{equation}
	p(\vc f^* | \mtx X^*, \scriptD) = \G{\vc f^*}{\ksf \kyyinv (\vc y - \vc \mu) + \vc \mu^*}{\kss - \ksf \kyyinv \kfs},
	\label{eqn:qf}
\end{equation}
where
\begin{align}
	(\kfs)_{i, j} &= k(\vc x_i, \vc x_j^*),
	\\
	(\kss)_{i, j} &= k(\vc x_i^*, \vc x_j^*),	
\end{align}
$\ksf = \kfs^\transpose$, and $\mu_i^* = \mu(\vc x_i^*)$.
Applying the likelihood and marginalizing out $\vc f^*$ gives the posterior in output space
\begin{equation}
	p(\vc y^* | \mtx X^*, \scriptD) = \G{\vc y^*}{\ksf \kyyinv (\vc y - \vc \mu) + \vc \mu^*}{\kss - \ksf \kyyinv \kfs + \sigma_y^2 \mtx I_{n^* \times n^*}}.
	\label{eqn:qy}
\end{equation}

\textbf{Remark:}
Our use of a likelihood term in our model definition generalizes the concept of a ``nugget'' or ``jitter'' that is sometimes used within the GP metamodeling community \cite{bilionis2012multi, bilionis2013multi}, usually with the stated purpose of either improving the conditioning of the kernel matrix or representing the noisiness in the observations being modeled.

\subsubsection{Incorporating general input features as inputs to Gaussian process regression models}
The main source of the rich behavior in GP models is their kernel function, which encodes strong inductive biases about the statistics of the functions being modeled.
It is most common to consider parametric kernels of the form 
$k(\cdot, \cdot; \vc \theta_k) : \reals^{d_x} \times \reals^{d_x} \rightarrow \reals$.
For example, the popular exponentiated quadratic kernel is defined as
\begin{equation}
	k(\vc x, \vc x'; \vc \theta_k) = \sigma_k^2 \exp \left[ - \sum_{i=1}^{d_x} \left(\frac{x_i - x_i'}{l_i} \right)^2 \right],
	\label{eqn:exponentiated-quadratic}
\end{equation}
and can be evaluated easily for any pair of $d_x$-dimensional vectors $\vc x$ and $\vc x'$.

While this is suitable for real-valued inputs commonly found in engineering problems, it is not suitable for more general inputs such as those mentioned above.
One workaround is to embed elements from general input spaces as $d_z$-dimensional latent variables; one can then form the complete input vector by concatenating these latents $\vc z$ with the real-valued inputs $\vc x^{(r)}$ to form a $d_{x^{(r)}} + d_z$-dimensional input vector amenable to computation with traditional parametric kernels such as Eq.\ (\ref{eqn:exponentiated-quadratic}).

More general nonparametric positive-definite kernels exist that are valid for Gaussian processes.
For example, the white noise kernel function,
\begin{equation}
	k_{white}(\vc x, \vc x'; \vc \theta) = 
	\begin{cases} 
		\sigma_{white}^2 & \textrm{if} ~ \vc x = \vc x' \\
		0 & \textrm{otherwise}
	\end{cases},
\end{equation}
does not require Euclidean inputs; instead, it merely requires that we are able to distinguish between elements in its index set.

\begin{figure}[hbt]
	\centering
	\includegraphics[width=0.6\textwidth]{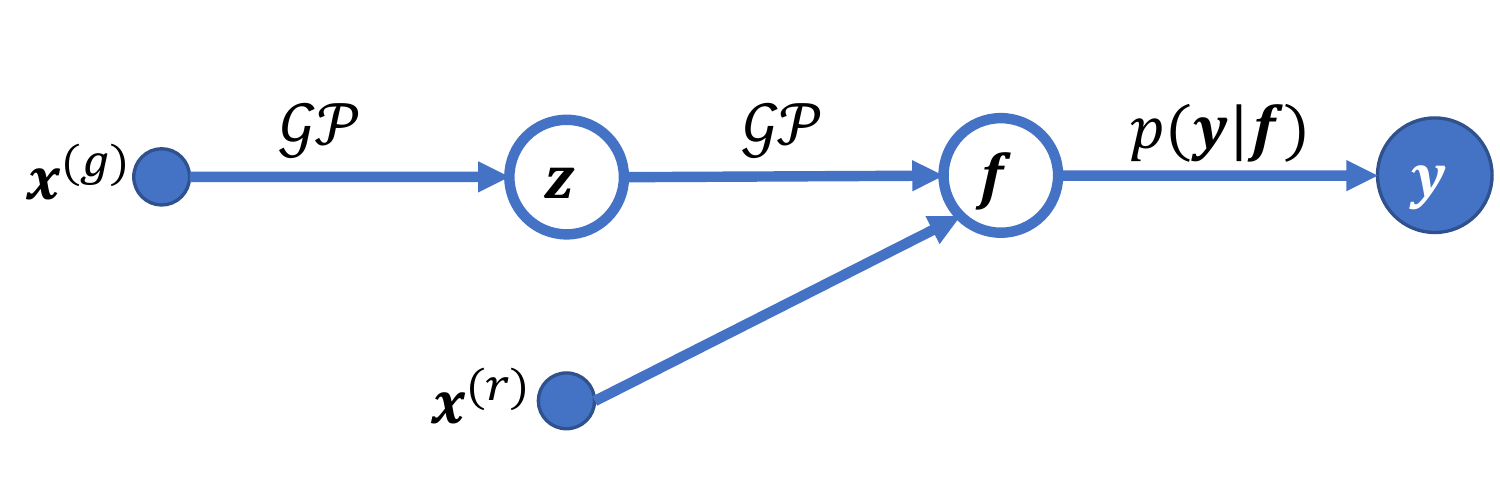}
	\caption{
		Probabilistic graphical model of the Bayesian embedding Gaussian process (BEGP) model.
		General inputs $\vc x^{(g)}$ are first mapped through a GP with white noise kernel to latent variables $\vc z$.
		These latents are concatenated with real-valued inputs $\vc x^{(r)}$ to form the full inputs $\vc x$ to the second GP model with a traditional parametric mean function and kernel.
		White nodes denote variables that are unobserved, and shaded nodes are observed.
		Point nodes are deterministic, and large nodes are associated with probability distributions.
		Model parameters associated with mean functions, kernels, and the likelihood are hidden for clarity.
	}
	\label{fig:egp}
\end{figure}

Therefore, to accomplish the embedding mentioned above, we map our general inputs $\vc x^{(g)} \in \scriptX_g$ through a $d_z$-variate Gaussian process with white noise kernel:
\begin{equation}
	\vc z_i \sim \mc{GP}\left(0; k_{white}(\cdot, \cdot)\right) : \scriptX_g \rightarrow \reals, ~i=1, \dots, d_z.	
\end{equation}
Sampling this GP at some general input returns potential embeddings of the input as a $d_z$-dimensional vector in Euclidean space.
These latent variables may then be concatenated with the numerical feature space and fed through a second Gaussian process that serves as the familiar regression model employed in Bayesian metamodeling.
The Bayesian embedding GP (BEGP) model that combines the embedding Gaussian process with the GP regression model is shown in Fig.\ \ref{fig:egp}.
This model is composed of two GPs, where the outputs of the first (embedding) GP are fed as inputs to the second (regression) GP, making it a deep Gaussian process with serial topology similar to those originally discussed in \cite{damianou2013deep}.

Our modeling approach may be thought of as a type of \textit{multi-task learning} in that one doing regression traditionally segregates datasets along differences in general features, building a separate model for each dataset using the remaining real-valued features; by contrast, here we build in correlations across general features within a single model, simultaneously improving the predictive power across all of the input space.

\textbf{Remark:} for systems with multi-dimensional general inputs (e.g.\ operator IDs \textit{and} machine serial numbers), it is straightforward to extend our approach to embed each general input dimension separately to its own latent space; these latents are then all concatenated with each other to define the total latent variable which is then concatenated with $\vc x^{(r)}$ as before.

\subsubsection{Inference for the Bayesian embedding GP model}
Having defined our model, we now discuss how to do inference on it.
Inference for the Bayesian EGP model presents two challenges.
First, inferring the posterior of the embedding layer is challenging due to its nonparameteric nature stemming from the white noise kernel combined with the fact that its posterior will generally be non-Gaussian due to the nonlinearity of the downstream model that operates on it. 
Second, unlike a traditional Gaussian process regression model, it is analytically intractable to marginalize out the latent variables $\vc z$.
In this section, we discuss simple strategies for overcoming these two challenges so that we may devise an objective function for training our model.

First, we consider the challenges associated with the embedding layer.
Let $n_g$ denote the number of distinct general inputs that we observe from $\scriptX_g$ (i.e.\ the number of legacy datasets that we would like to incorporate in our model).
Due to the kernel, the joint prior density over latents $\mtx Z \in \reals^{n_g \times d_z}$ associated with general inputs $\mtx X^{(g)} \in \scriptX_g^{n_g}$ is
\begin{equation}
	p \left( \mtx Z | \mtx X^{(g)} \right) = \prod_{i=1}^n \prod_{j=1}^{d_z} \G{z_{ij}}{0}{\sigma_w^2},
	\label{eqn:pz}
\end{equation}
where $z_{ij}$ is the $j$-th dimension associated with the latent for input $i$.
For reasons analogous to those mentioned in previous literature \cite{titsias2010bayesian, damianou2013deep, salimbeni2017doubly}, the posterior over these inputs is generally non-Gaussian.
We define a mean-field variational posterior $q(\mtx Z)$ to approximate the true posterior $p(\mtx Z | \scriptD)$ with the form
\begin{equation}
	q(\mtx Z) = \prod_{i=1}^{n_g} \prod_{j=1}^{d_z}  \G{z_{ij}}{m_{ij}}{s_{ij}},
	\label{eqn:qz}
\end{equation}
where $\mtx M, \mtx S \in \reals^{n_g \times d_z}$ are variational parameters.
Fortunately, one is usually only interested in a relatively small number of inputs in $\scriptX_g$, so it is not a challenge to track these variational parameters.
In fact, computational challenges associated with data scalability of Gaussian process models are likely to become problematic well before challenges associated with working with a high number of tasks, even though recent advances have make significant progress in lift traditional barriers to Gaussian process modeling in large-data regimes 
\cite{snelson2006sparse, titsias2009variational, hensman2013gaussian, wilson2015kernel, salimbeni2017doubly, wang2019exact}.

Lastly, note that if no data have been observed for some held-out general input $\vc x^{(g),*}$, then, by the nature of the white noise kernel, the posterior over the associated latent is equal to the prior.
this observation is critical for enabling our model to make credible predictions in a zero-shot setting, where no data about some task of interest is available at the outset of optimization.
Indeed, we will see that this probabilistic approach enables our model to make surprisingly good inferences to guide the first iteration of Bayesian optimization in such settings.

Having resolved the challenge of representing the posterior of the embedding layer, we now turn our attention to the second challenge regarding the intractability of the model's marginal log-likelihood (or evidence), which usually serves as the objective function for training probabilistic models.
Again, following a variational strategy, we derive a tractable evidence lower bound (ELBO) that can be used in place of the exact marginal log-likelihood.
While this approach was first demonstrated for Gaussian processes using a sparse approach based on inducing variables, it turns out that later advances in variational inference \cite{ranganath2014black} allow us to avoid relying on a sparse GP formulation.
That said, we recognize that such an approach may realistically be helpful in the plausible cases where an abundance of legacy data takes the problem into a large-data setting.
This is in contrast to the usual assumptions of data-scarcity in engineering metamodeling, which are usually myopically focused only on solving a single task in isolation.
For the sake of completeness, we now provide the derivation of the ELBO for the BEGP model.
Based on the probabilistic graphical model shown in Fig.\ \ref{fig:egp}, the joint probability of our model with $n$ observations is
\begin{equation}
	\mc P = p(\mtx Z, \vc f, \vc y | \mtx X^{(r)}, \mtx X^{(g)}) = p(\mtx Z | \mtx X^{(g)}) p(\vc f | \mtx Z, \mtx X^{(r)}) p(\vc y | \vc f),
\end{equation}
where p$(\mtx Z | \mtx X^{(r)})$ is given by Eq.\ (\ref{eqn:pz}), $p(\vc f | \mtx Z, \mtx X^{(r)})$ is analogous to Eq.\ (\ref{eqn:pf}), and $p(\vc y | \vc f)$ is the likelihood of Eq.\ (\ref{eqn:likelihood}).
and the marginal log-likelihood is obtained by simply integrating over the latent variables and taking the logarithm of the result:
\begin{equation}
	\log p(\vc y | \mtx X^{(r)}, \mtx X^{(g)}) = \log \int \mc P d \vc f d \mtx Z.
\end{equation}
From this point onwards, we will omit the conditioning on the inputs for brevity.
While we can integrate out $\vc f$ due to the conjugacy of the likelihood, $\mtx Z$ cannot be integrated out analytically.
We multiply and divide the integrand by $q(\mtx Z)$ to obtain
\begin{equation}
	\log p(\vc y) = \log \int q(\mtx Z) \frac{p(\vc y, \mtx Z)}{q(\mtx Z)} d \mtx Z.
\end{equation}
Applying Jensen's equality, we move the logarithm inside the integrand to get
\begin{equation}
\log p(\vc y) \ge ELBO = \int q(\mtx Z) \log \frac{p(\vc y, \mtx Z)}{q(\mtx Z)} d \mtx Z.
\end{equation}
Rearranging and using $p(\vc y, \mtx Z) = p(\vc y | \mtx Z) p(\mtx Z)$ gives
\begin{equation}
	ELBO = \int q(\mtx Z) \left( \log p(\vc y | \mtx Z) + \log p(\mtx Z) - q(\mtx Z) \right) d \mtx Z.
	\label{eqn:elbo}
\end{equation}
Additional progress may be made at this point by restricting our choice of kernel function for the second GP layer to either an exponentiated quadratic or a linear kernel since the kernel expectations that form the crux of the challenge may then be evaluated analytically.
Furthermore, under the special case where the same real-valued input points are sampled for each general input, Atkinson and Zabaras \cite{atkinson2018structured, atkinson2019structured} showed that the kernel expectations can be further broken down into a Kronecker product between a deterministic kernel matrix and a kernel expectation enabling extensions to modeling millions of data.
In this work, we instead opt to estimate Eq.\ (\ref{eqn:elbo}) through sampling so as to maintain the ability to leverage arbitrary kernel functions.

The model is trained by maximizing Eq.\ (\ref{eqn:elbo}) with respect to the parameters of the variational posteriors as well as the kernel hyperparameters and likelihood variance $\sigma_y^2$ through gradient-based methods.
We find that it is typically sufficient to approximate the integral with a single sample from $q(\mtx Z)$ at each iteration of the optimization.

Note that there is a redundancy in defining the scale of the latent space in that our model possesses both a parameter $\sigma_w^2$ for the prior variance of the embedding layer as well as length scales for the parametric kernel of the GP regression layer.
Given this freedom, one can make the ELBO arbitrarily high by bringing $\sigma_w^2$ towards zero while scaling the variational parameters of $q(\mtx Z)$ and the length scales of the regression kernel correspondingly without changing the predictions of the model.
to avoid this pathology and establish the scale of the latent space, we fix $\sigma_w^2 = 1$.

We choose to learn point estimates of the other model parameters since they are informed by all of the tasks' data.
Therefore, the uncertainty contributed to the predictions due to uncertainty in these parameters will generally be small.
This is again only because we are incorporating a large amount of data into our model by leveraging legacy data; in the typical single-task engineering setting, we might expect to have far fewer data, and Bayesian inference over the parameters may provide value to the model.

\paragraph{A remark on variational inference for EGP}
A reasonable alternative to our variational inference approach is to instead perform inference using Monte Carlo.
We favor VI because we need to update the model posteriors repeatedly during Bayesian optimization as new data are added to the model.
This is simple to do with VI since we can continue our gradient-based optimization from the previous variational posterior.
By contrast, Monte Carlo requires us to run the chain for almost as long as the initial chain every time that we need to update the model.
Thus, while there is not such a clear advantage for one method compared to the other when initially training the model on legacy data, VI quickly becomes more attractive in the following steps.
Furthermore, we note that updating the model parameters is rather important since we are focusing on extreme low-data regimes (fewer than 10 data from the task of interest), where each new datum will generally have considerable effects on the posterior.

Another reason for using MC inference is because highly-factorized variational posterior distributions are known to underestimate the entropy of the true posterior when there is mismatch between the variational form and the true posterior \cite{bishop2006pattern}.
One might reasonably worry that we are neglecting important uncertainty information by using a mean-field posterior for $q(\mtx z)$.
We find in practice that this approximation successfully captures much of the uncertainty and is sufficient to provide competitive or superior predictions compared to the baselines considered in our examples.
However, our method is not incompatible with more powerful VI methods such as multivariate Gaussians with full covariance matrices \cite{atkinson2019structured} or normalizing flow-based approaches \cite{kingma2016improved}.

\subsubsection{Predictions with the model}
Given a trained model, we are now interested in making predictions at held-out inputs in $\scriptX_r \times \scriptX_g$.
Consider a set of test inputs $\mtx X^* \in \scriptX_g^{n_g^*} \times \scriptX_r^{n_r^*}$ with $n_g^*$ distinct general inputs and $n_r^*$ distinct real-valued inputs, combined to form $n^*$ test inputs in all.
We first apply the embedding layer $\mtx X^{(g),*}$ to obtain a sample of the latents $\mtx Z^* \in \reals^{n_g^* \times d_z}$ as well as the training data's latents $\mtx Z$.
These latents are expanded to form the full training and test inputs for the regression model, 
$\{\mtx X, \mtx X^*\} = \left\{ [\mtx X^{(r)}, \mtx Z], [\mtx X^{(r),*}, \mtx Z^*] \right\} \in \reals^{n \times d_x} \times \reals^{n^* \times d_x}$,
where $d_x = d_{x^{(r)}} + d_z$.
Given these samples we can compute the conditional predictive distributions over the latent and observed outputs, given by Eq.\ (\ref{eqn:qf}) and (\ref{eqn:qy}), respectively.

Due to the nonlinearity of the GP model, marginalizing over latent variables' variational posterior $q(\mtx Z, \mtx Z^*)$ generally induces a non-Gaussian predictive distribution.
However, given a Gaussian variational posterior, the moments of the marginal predictive distribution admit analytic expressions; previous works \cite{girard2003gaussian, titsias2010bayesian} have approximated the predictive distribution as Gaussian with these statistics.
Alternatively, one may sample the predictive distribution in order to better resolve its details.
We utilize the former approach in our examples.

\subsection{Bayesian optimization}
Having discussed the formulation of our model, including training and predictions, we now turn our attention to using it in the context of optimization.
The main algorithm for Bayesian optimization is given in Algorithm \ref{alg:BO}.
We restrict ourselves to the task of optimizing subject to a fixed general input $\vc x_g^*$, though our method permits searching over multiple general inputs with trivial modification.

\begin{algorithm}
	\caption{Bayesian optimization}
	\label{alg:BO}
	\textbf{Require:} 
	Training data $\scriptD$, 
	general input of interest $\vc x_g^*$, 
	acquisition function $a(\cdot)$, 
	computational budget $n^*$.
	
	\textbf{Ensure:} optimal design $\vc x^{(r),*}$
	
	\begin{algorithmic}[1]
			\State $\scriptM \rightarrow BEGP(\scriptD)$.
			\State $n \rightarrow 0$, $\scriptD^* = \emptyset$.
			\For{$i=1, \dots, n^*$}
				\State Train $\scriptM$ on $\scriptD$.
				\State $\vc x_{next}^{(r)} \rightarrow \argmin_{\vc x^{(r)} \in \mc X_r} a(\vc x^{(r)})$.
				\State $y_{next} \rightarrow \eta(\vc x_g^*, \vc x_{next}^{(r)})$, $n \rightarrow n + 1$.
				\State $\scriptD \rightarrow \scriptD \cup \{ (\vc x_g^*, \vc x_{next}^{(r)}, y_{next}) \}$
				\State $\scriptD^* \rightarrow \scriptD^* \cup \{ (\vc x_g^*, \vc x_{next}^{(r)}, y_{next}) \}$
			\EndFor
			\Return $\vc x^{(r),*} = \argmin_{x^{(r)} \in \scriptD^*} y(\vc x^{(r)})$.
	\end{algorithmic}
\end{algorithm}

\subsubsection{Acquisition functions}
One ingredient that must be specified for BO is the acquisition function $a$, which is used to steer the selection of points at which one conducts experiments to evaluate $\eta$.
In this work, for systems in which we can evaluate $\eta$ at any location in $\scriptX_r$ for the current task $\vc x_g^*$, we consider the expected improvement
\begin{equation}
EI(\vc x) = \expectation{p(y | \vc x, \scriptD)}{(y(\vc x) - y_{min})\Theta(y_{min} - y(\vc x))},
\label{eqn:expected_improvement}
\end{equation}
where $\Theta(\cdot)$ is the Heaviside theta function and $y_{min}$ is the value of the current best design.
When $p(y|\vc x, \scriptD)$ has a known form such as a Gaussian, evaluation of Eq.\ (\ref{eqn:expected_improvement}) can be done cheaply.
Notice as a matter of convention that we have defined $a$ in Eq.\ (\ref{eqn:expected_improvement}) such that lower values are better.

Finally, the traditional approach to maximizing the acquisition function has been to simple select a large random sampling of $\mc X_r$, evaluate $a$ on all of the samples (which is generally cheap), then select the best sample.
However, following the widespread adoption of computational frameworks supporting automatic differentiation such as TensorFlow \cite{abadi2016tensorflow} and PyTorch \cite{paszke2017automatic}, it has become customary to accelerate the inner loop optimization over $a$ using gradient-based methods by simply backpropagating the acquisition function back through the model to obtain its gradient with respect to the inputs $\vc x^{(r)}$ \cite{snoek2012practical}.
Other recent work has investigated this matter more broadly \cite{zhang2019finding}.
Here, we use gradient-based search with restarts to find the next point to select similarly to \cite{snoek2012practical}.
For the case where one must estimate $a$ through sampling (e.g.\ due to estimating the predictive posterior by sampling latents from $q(\mtx Z)$, stochastic backpropagation \cite{rezende2014stochastic} can be used to deal with the variance in the estimates of $a$ due to sampling.
We also note that this setup may be applied immediately to solve \textit{robust} optimization problems with no modification by additionally sampling from any uncontrolled inputs.

In our real-world examples, we do not have access to the real-world data-generating process and must instead work with a static, pre-computed dataset.
In this case, our design space is practically the finite set at which experiments have already been carried out.
In this case, we can instead directly estimate the negative\footnote{We take the negative probability to keep with the convention that lower values of $a$ are better.} probability that a given available point will be the best design:
\begin{equation}
	a(\vc x_i^{(r)}) = \prod_{j \not = i} P\left(y(\vc x_i^{(r)}, \vc x^{(g), *}) < y(\vc x_j^{(r)}, \vc x^{(g), *}) \right).
	\label{eqn:prob-best}
\end{equation}
We approximate Eq.\ (\ref{eqn:prob-best}) by repeatedly sampling the joint predictive distribution over all available design points and counting the frequency with which each design point is predicted to have the best output.
We find that this crude approximation is not overly computationally burdensome and results in good performance in practice as shown by our examples.

\section{Related work}
\label{sec:related_work}
McMillan et al.\ \cite{mcmillan1999analysis} study Gaussian process models based on ``qualitative'' and ``quantitative'' variables.
This approach would be analogous to a deterministic version of our Bayesian embedding layer.
As we show in our examples, the incorporation of Bayesian uncertainty estimates provides highly valuable information that is essential to making the predictive distribution of the model credible.

The Gaussian process latent variable model \cite{lawrence2004gaussian} is a seminal work for inferring latent variables by using the marginal log-likelihood of a Gaussian process as training signal.
The later Bayesian extension by Titsias and Lawrence \cite{titsias2010bayesian} was found to significantly improve the quality of the model and gives compelling evidence in favor of modeling latent variables in a Bayesian manner.

The task of inferring input-output relationships where the inputs are only partially-specified was first identified by Damianou and Lawrence as ``semi-described learning'' \cite{damianou2015semi}.
Our problem statement may be regarded as similar in that we choose to associate the general features in our data with latent variables that are unspecified \textit{a priori}.

Using related systems to improve knowledge about a related system of interest has a long history in multi-fidelity modeling \cite{kennedy2000predicting, forrester2007multi}, where one traditionally considers a sequence of systems that constitute successively more accurate (and expensive) representations of a true physical process.
However, it is unclear how to define a ``sequence'' of legacy datasets that are all ``on equal footing'' (such as different operators or machines performing the same task), particularly as the number of such instances becomes large.

Our model is similar to the multi-task Gaussian process introduced in \cite{swersky2013multi}.
Our approaches are similar in that we both learn a kernel over tasks; however, \cite{swersky2013multi} require that the kernel matrix decompose as a Kronecker product between a task-wise covariance matrix and the input kernel matrix. 
The Kronecker product kernel matrix can be equivalent to a kernel matrix evaluated on our augmented inputs in the special cases of if one uses either a linear or a Gaussian kernel, but more general kernels cannot be composed in this way.
By finding representations of tasks in a Euclidean input space, we lift this restriction, allowing us to use arbitrary kernels instead, wherein the Kronecker product decomposition of \cite{swersky2013multi} is a special case.
Additionally, by posing a prior over the latent input space, our method also allows us to do principled zero-shot predictions on unseen inputs; it is significantly more challenging to formulate a reasonable prior on the kernel matrix itself.
Finally, we can inject additional modeling bias into our model by selecting the dimensionality $d_z$ of our latent space.
While a low-rank approximation to the task covariance matrix might be derived to obtain similar savings, it again seems more natural in our opinion to exercise this ability in a latent space.

The task of utilizing ``legacy'' data in order to improve the predictive capability on a related system of interest was explored in \cite{ghosh2018bayesian}.
However, the approach used in that work can only utilize legacy models through linear combinations, making extrapolation on the system of interest difficult, particularly when few (or no) data are available from the task of interest.
Additionally, their method requires a cross-validation scheme for model selection; by contrast, the probabilistic representation of tasks in the current method enables the use of Bayesian inference to guide model selection in the sense of identifying relevance between tasks.

A recent work by Zhang et al.\ \cite{zhang2019bayesian} also studies Bayesian optimization in the context of qualitative and quantitative inputs.
However, they learn point embeddings via maximum likelihood estimation rather than inferring a posterior through Bayesian inference; this makes the overall model prone to overfitting in practice and generally unsuitable for providing credible uncertainty estimates.
This also makes their model difficult to apply in few-shot settings, and impossible to apply in a principled way for zero-shot learning.
A related work by Iyer et al.\ \cite{iyer2019data} identifies a pathology associated with the point estimate inference strategy and proposes an ad hoc workaround; a natural outcome of our Bayesian embedding approach is that this challenge vanishes.

The BEGP model can be thought of as a latent variable model \cite{kingma2013auto, lawrence2004gaussian, titsias2010bayesian} and has similarities with previous works with latent variable GPs \cite{dai2017efficient, atkinson2018structured, atkinson2019structured}, though those works additionally impose structure over latent variables as well as (potentially) the input training data.
However, neither works were applied within the context of Bayesian optimization, nor did they identify that multiple general feature sets could be decomposed as we have chosen to do.

\section{Examples}
\label{sec:examples}
In this section, we demonstrate our method on a variety of synthetic and real-world examples.
The embedding GP model is implemented using \texttt{gptorch}\footnote{\url{https://github.com/cics-nd/gptorch}}, a Gaussian process framework built on PyTorch \cite{paszke2017automatic}.
Source code for implementing the model and synthetic experiments described below can be found on GitHub\footnote{\url{https://github.com/sdatkinson/BEBO}}.
For each system, we consider both a regression and optimization problem using our embedding GP model with both probabilistic and deterministic embedding layers; the latter is achieved by replacing the posterior of Eq.\ (\ref{eqn:qz}) with a delta function at its mode.
We use an RBF kernel as in Eq.\ (\ref{eqn:exponentiated-quadratic}) and constant mean function whose value is determined via an MLE point estimate.

We compare our results against a vanilla Gaussian process metamodel that is unable to directly leverage the legacy data as a baseline.
Since GP models typically fare very poorly the the extreme low-data regime that we are interested in, we conduct full Bayesian inference over the mean and kernel function parameters, using as prior a factorized Gaussian over the parameters\footnote{Or the logarithm of the parameters that are constrained to be positive.} whose mean and variance are estimated by fitting GPs to each legacy task and using the statistics of the point estimates of the model parameters found therein.
Inference for the Bayesian GP is carried out using Hamiltonian Monte Carlo \cite{duane1987hybrid} with the NUTS sampler \cite{hoffman2014no} as implemented in Pyro \cite{bingham2018pyro}.

\subsection{Systems}
We begin by describing the systems that we consider.
Each system contains a set of tasks that we will jointly learn.

\subsubsection{Toy system}
\label{sec:examples:systems:toy}
We first consider a system with one-dimensional input and $\Omega_x = [0, 1]$.
The set of response surfaces are given by
\begin{equation}
	\eta(x, \vc \theta) = 0.1 * z^4 - z^2 + (2.0 + \theta_2) \sin(2z),
\end{equation}
where we define $z = \theta_1 + 4x - 4$.
Different response surfaces are selected by sampling $\vc \theta \sim \mc U[0, 1]^2$.
We consider a problem setting where $5$ legacy tasks are available, each with $5$ data with inputs sampled uniformly from $\Omega_x$.
In the regression setting, we randomly sample a number of data from one response surface as the current task and split it into a training and test set.
We quantify our prediction accuracy in terms of mean negative log probability (MNLP), mean absolute error (MAE), and root mean square error (RMSE).
In the optimization setting, we perform Bayesian optimization over the current task and track the best design point as a function of the number of evaluations on the current task.
We repeat both experiments with $10$ different random seeds to quantify the performance statistics of each metamodeling approach.

\subsubsection{Forrester function}
\label{sec:examples:systems:forrester}
We also consider a generalization of the Forrester function \cite{forrester2007multi} to a many-task setting:
\begin{equation}
	\eta(x; \vc \theta) = \theta_1 * \eta_{forrester}(x) + \theta_2 * (x - 0.5) + \theta_3,
	\label{eqn:forrester-param}
\end{equation}
where 
\begin{equation}
	\eta_{forrester}(x) = (6x-2)^2 \sin(12x-4).
	\label{eqn:forester-hi}
\end{equation}
Under the parameterization of Eq.\ (\ref{eqn:forrester-param}), the original ``high-fidelity'' function considered in \cite{forrester2007multi} is obtained when $\vc \theta = (1, 0, 0)$, and the low-fidelity function corresponds to $\vc \theta = (0.5, 10, -5)$.
We always include the low-fidelity function in the legacy tasks along with other tasks generated by sampling 
$\vc \theta \sim \mc U[0, 1] \times \mc U[0, 10] \times \mc U[-5, 5]$.
Note that this places the high-fidelity task at an extremum of the system parameter space, meaning that multi-task models will not necessarily benefit by assuming \textit{a priori} that the held-out task lives at the center of latent space (assuming that their learned latent representation resembles that of the parameterization we have chosen).
We consider a design space $\Omega_x = [0, 1]$.
Each of the 5 legacy tasks includes 5 data where the input was sampled uniformly in $\Omega_x$.
We consider a regression and an optimization with identical setup to the toy system in Sec.\ \ref{sec:examples:systems:toy}.

\subsubsection{Pump data}
\label{sec:examples:systems:pump}
Our first real-world dataset comprises of data in which the performance of a pump is quantified in terms of seven design variables.
Our data includes data from 6 pump families, each with of which has between $11$ and $55$ data.
We consider each pump family in turn as the ``current'' task while using the other families as legacy task data.
We repeat our experiment 6 times, each time using a different pump family as the current task and the other 5 datasets as legacy task data.
In the regression setting, we split the current task data into a train and test set, training our metamodel on all of the legacy data as well as whatever data is available from the current task, then quantify our predictions on the held-out test set of the current task in terms of MNLP, MAE, and RMSE.
Each experiment is repeated 10 times with different random initializations and train-test splits.
In the optimization setting, we begin with no evaluations from the current task and carry out Bayesian optimization using the available data as a (finite) design set.
Data are selected at each iteration according the acquisition function of Eq.\ (\ref{eqn:prob-best}).
We track the best evaluation (relative to the per-task best design) as a function of the number of evaluations on the current task.

\subsubsection{Additive manufacturing data}
\label{sec:examples:systems:additive}
We consider a dataset in which the performance of an additive manufacturing process is quantified in terms of four design variables.
We have data from $35$ different chemistries, each of which has $24$ data.
We conduct experiments on each of the $35$ chemistries in an analogous manner to the experiments on the pump data.

\subsection{Regression results}
\begin{figure}[hbt]
	\centering
	\includegraphics[width=0.3\textwidth]{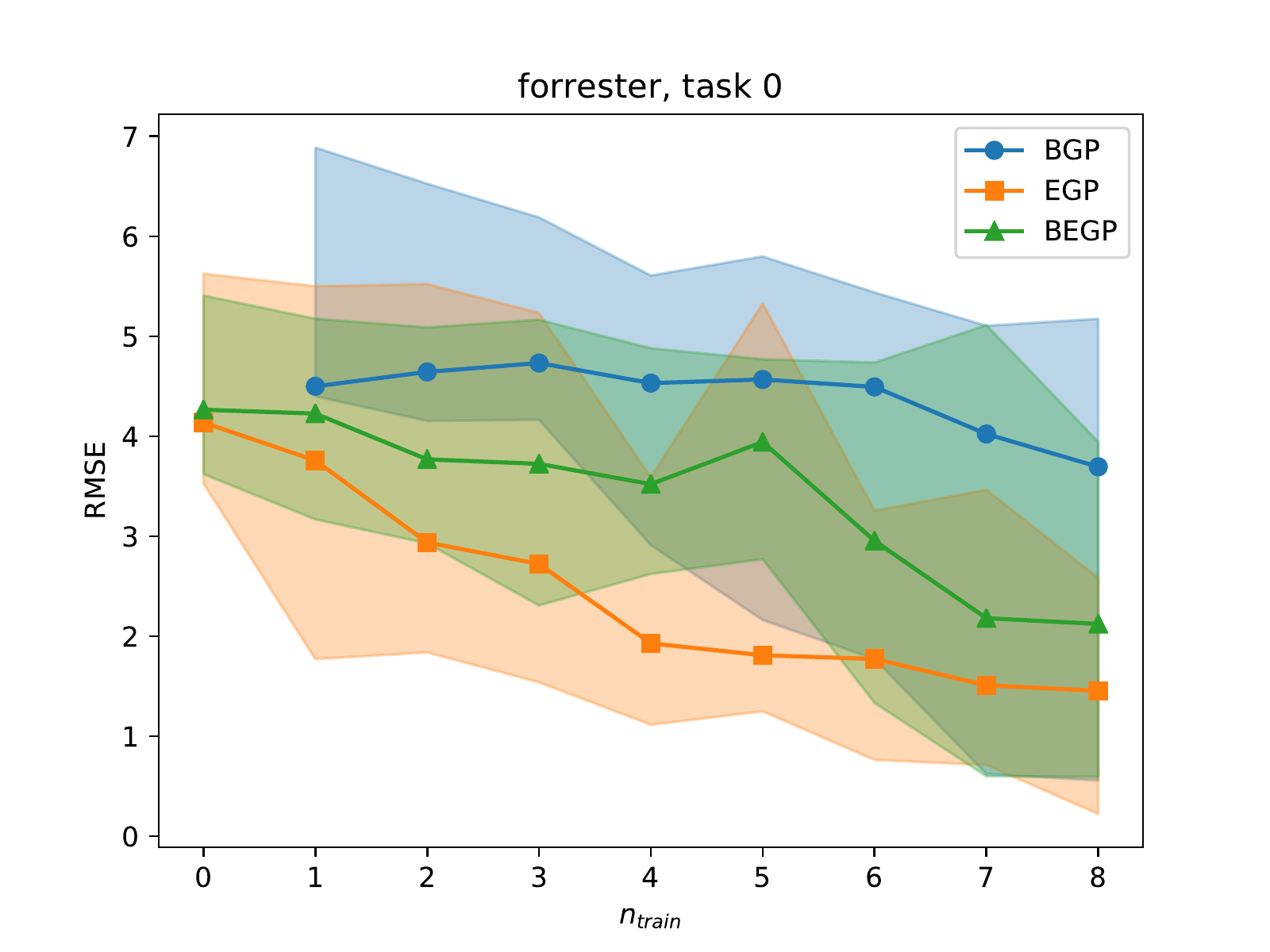}
	\includegraphics[width=0.3\textwidth]{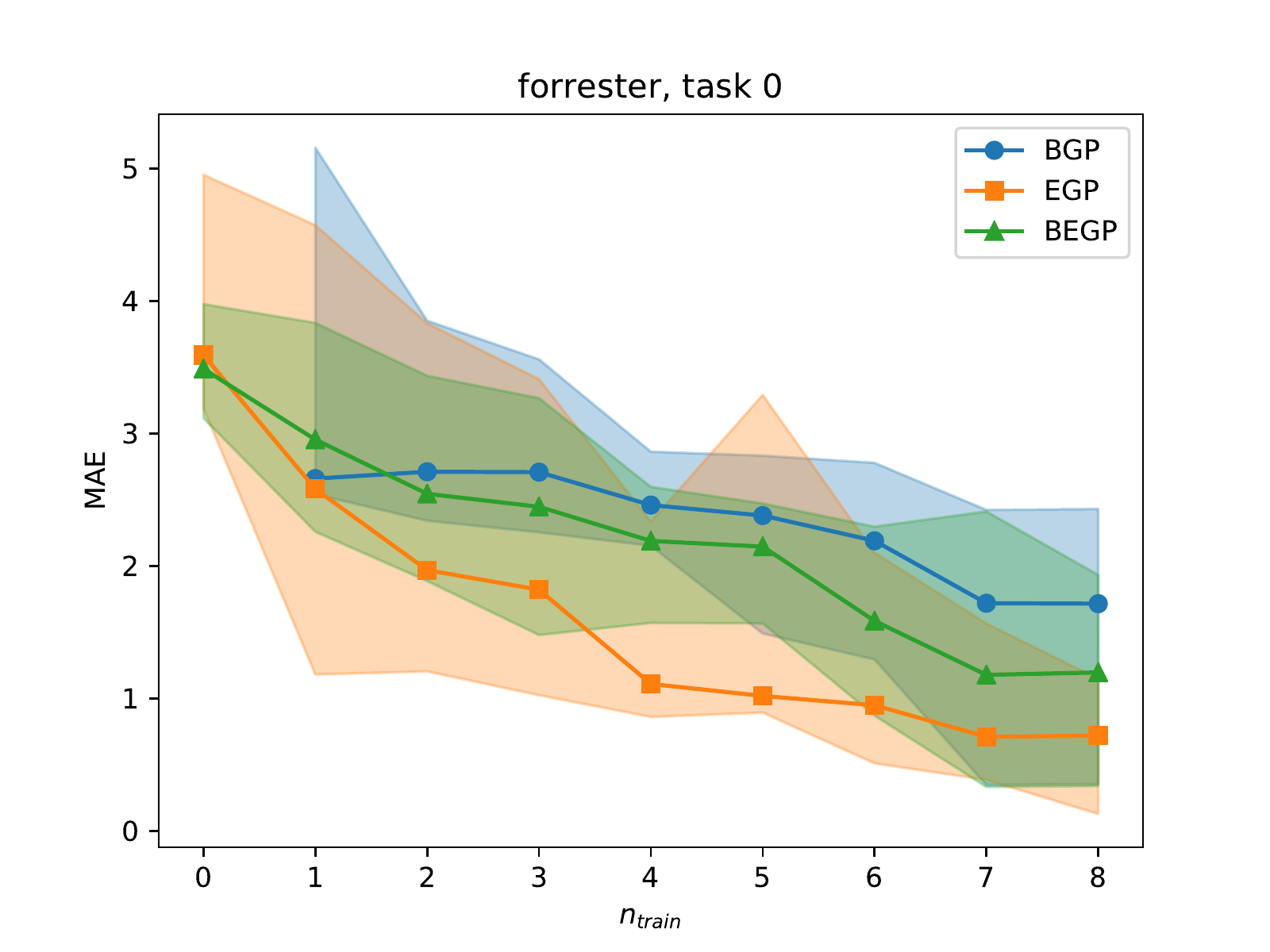}
	\includegraphics[width=0.3\textwidth]{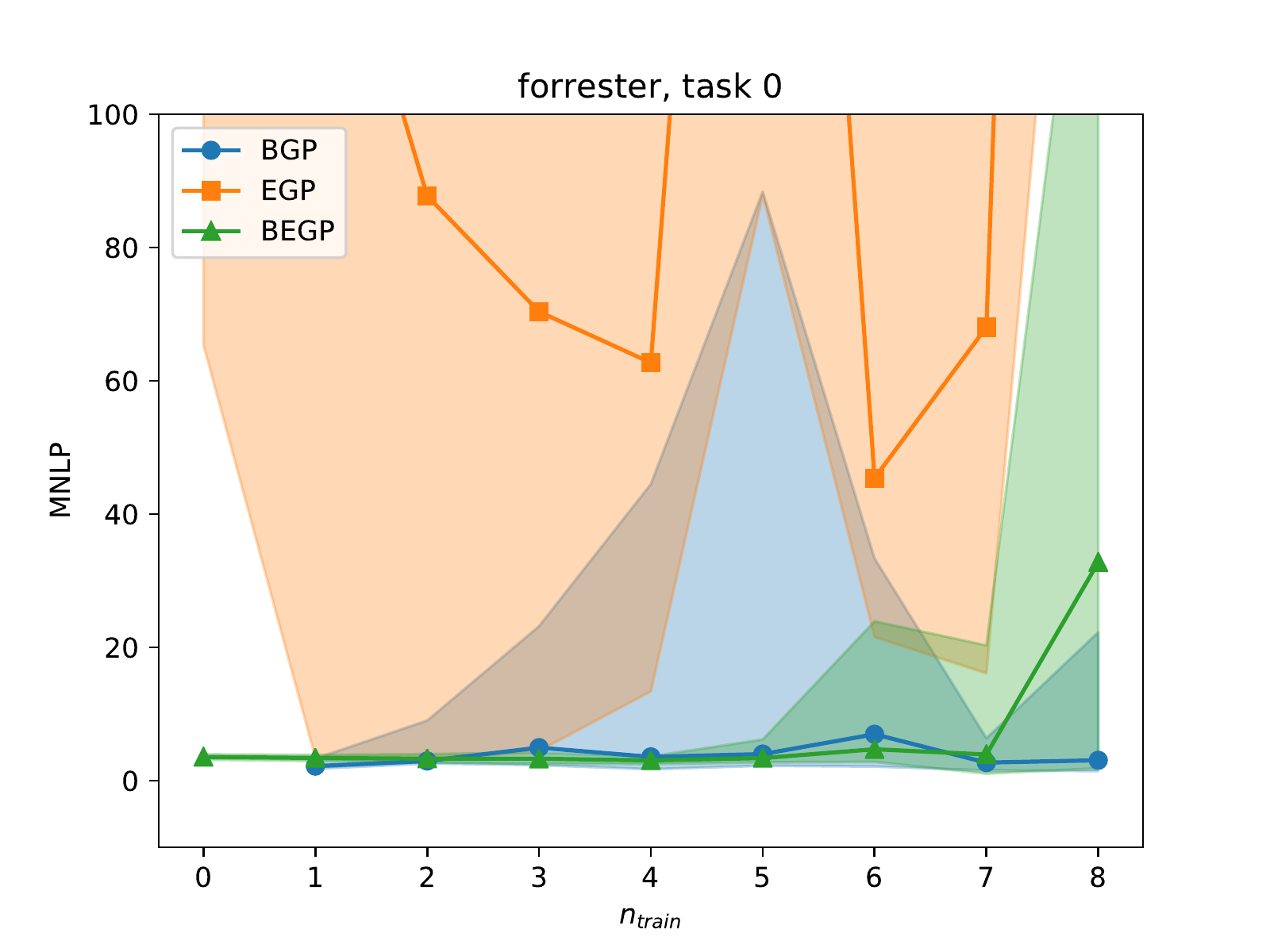}
	\\
	\includegraphics[width=0.3\textwidth]{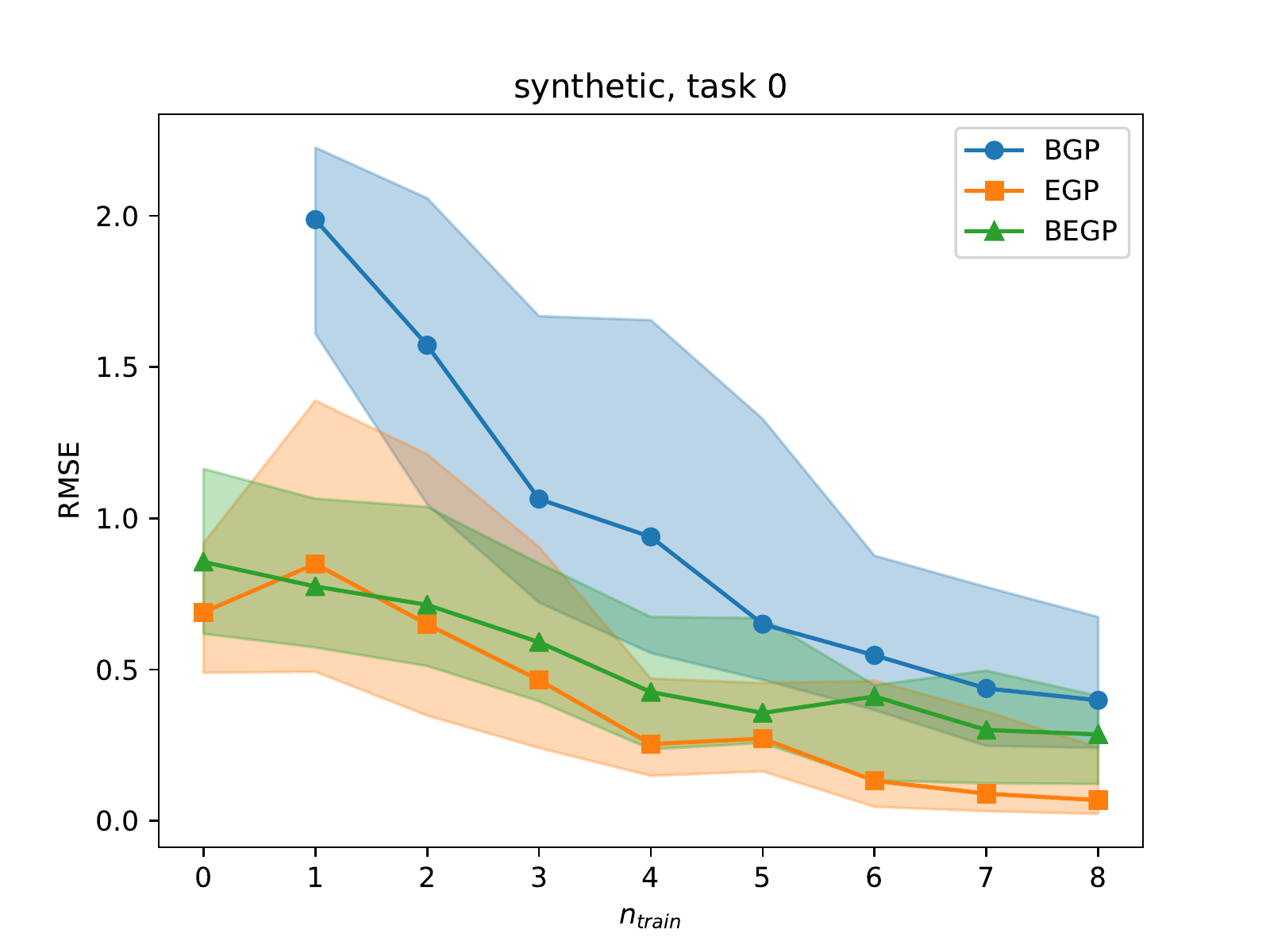}
	\includegraphics[width=0.3\textwidth]{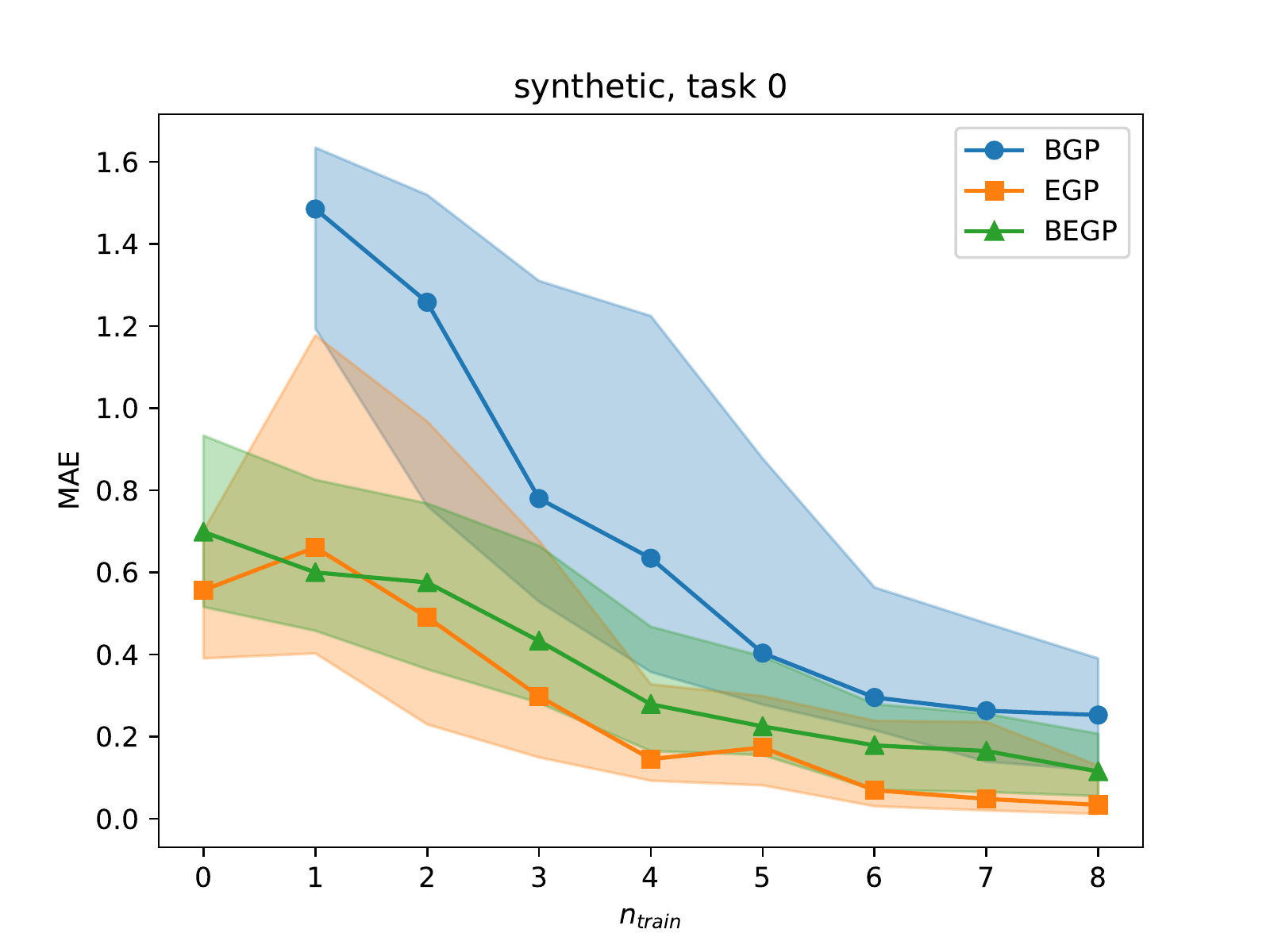}
	\includegraphics[width=0.3\textwidth]{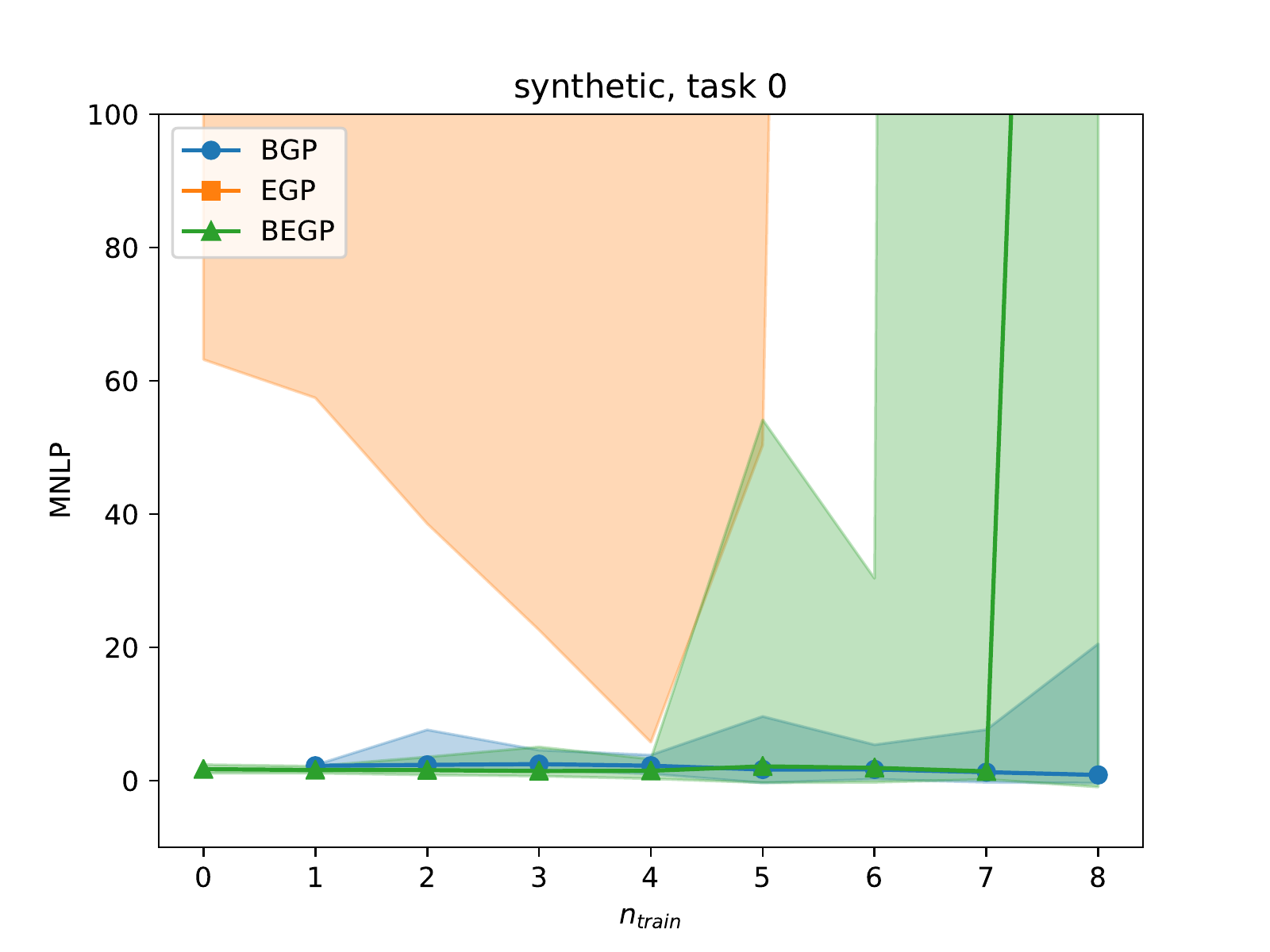}
	\caption{
		(Regression, synthetic systems)
		Performance on held-out tasks.
		From left to right, performance in terms of root mean squared error (RMSE), mean absolute error (MAE), and median negative log probability (MNLP).
		Top row: Forrester function.
		Bottom row: synthetic system of functions.
		Solid lines show the median performance over 10 random seeds and the shaded region shows the 80\% coverage interval.
	}
	\label{fig:examples:regression:synthetic}
\end{figure}

Figure \ref{fig:examples:regression:synthetic} shows regression results for the synthetic systems described in Sections \ref{sec:examples:systems:toy} and \ref{sec:examples:systems:forrester}.
We notice that the embedding GPs typically outperform the Bayesian GP by a margin, though there is overlap in the statistics of their performance over repeats.
Curiously, we notice that the deterministic embedding GP occasionally outperforms the Bayesian embedding in terms of RMSE and MAE, but fails catastrophically in terms of MNLP, where the overconfidence of its predictions is severely penalized.

\begin{figure}[hb!t]
	\centering
	\includegraphics[width=0.3\textwidth]{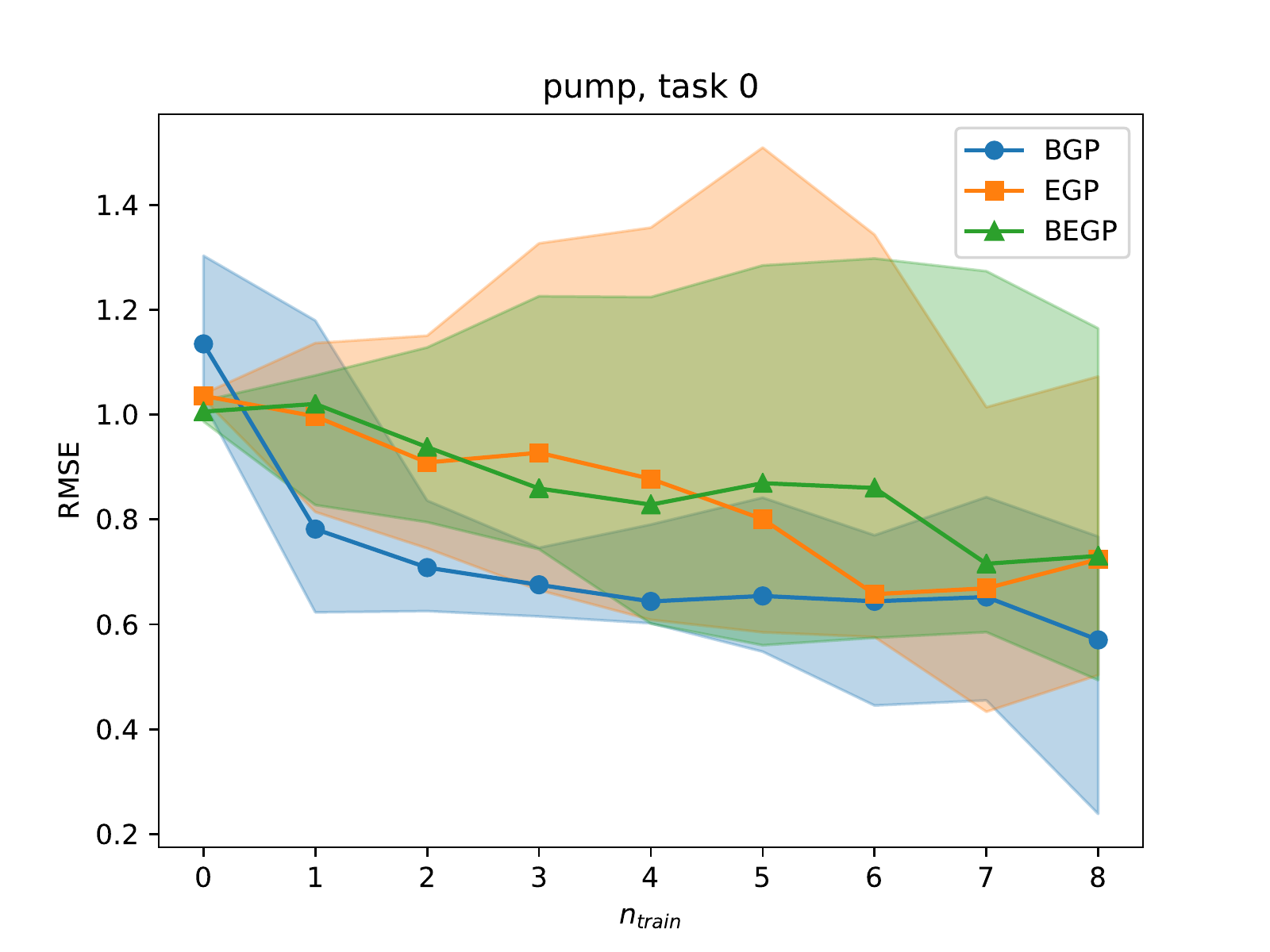}
	\includegraphics[width=0.3\textwidth]{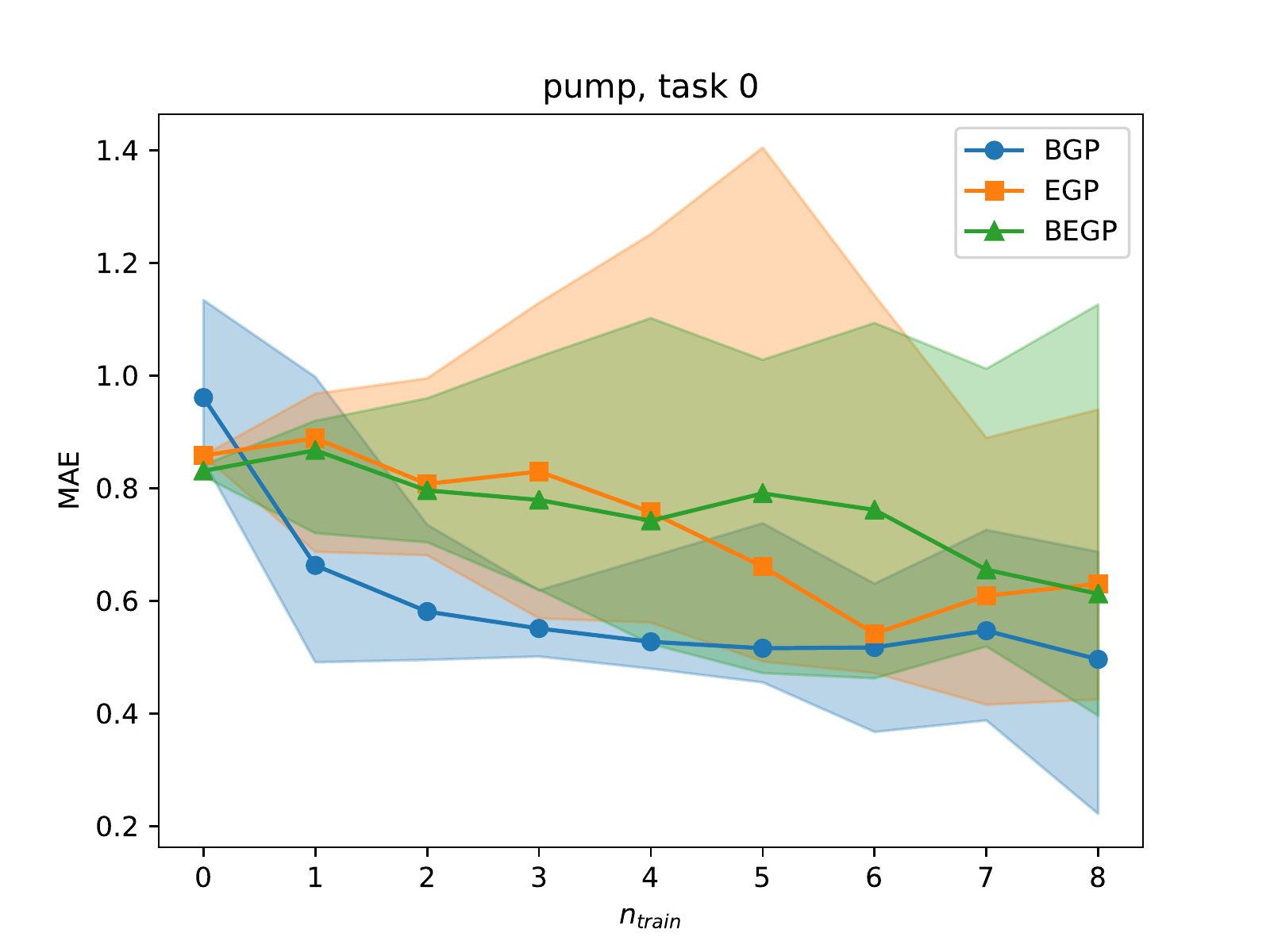}
	\includegraphics[width=0.3\textwidth]{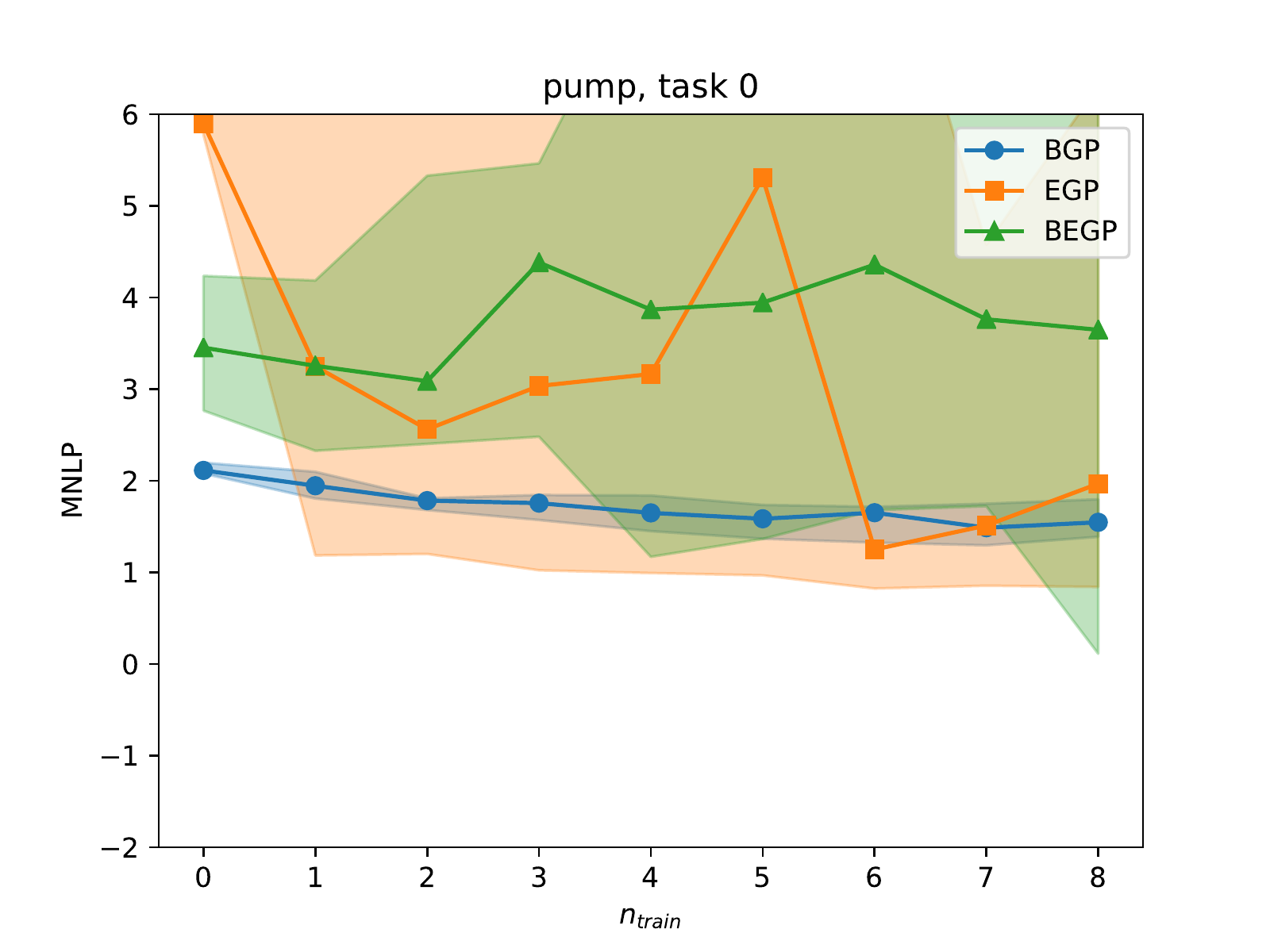}
	\\
	\includegraphics[width=0.3\textwidth]{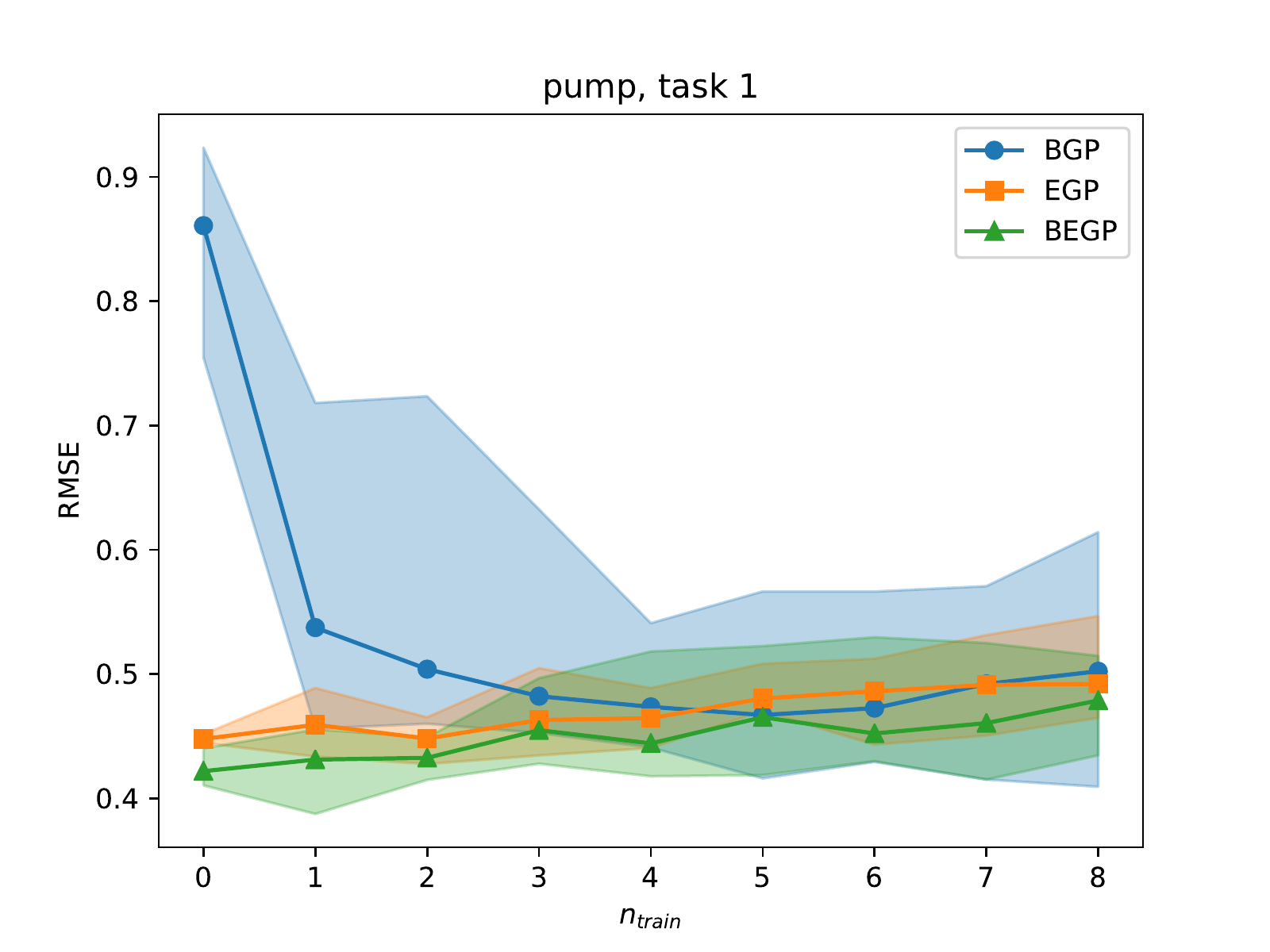}
	\includegraphics[width=0.3\textwidth]{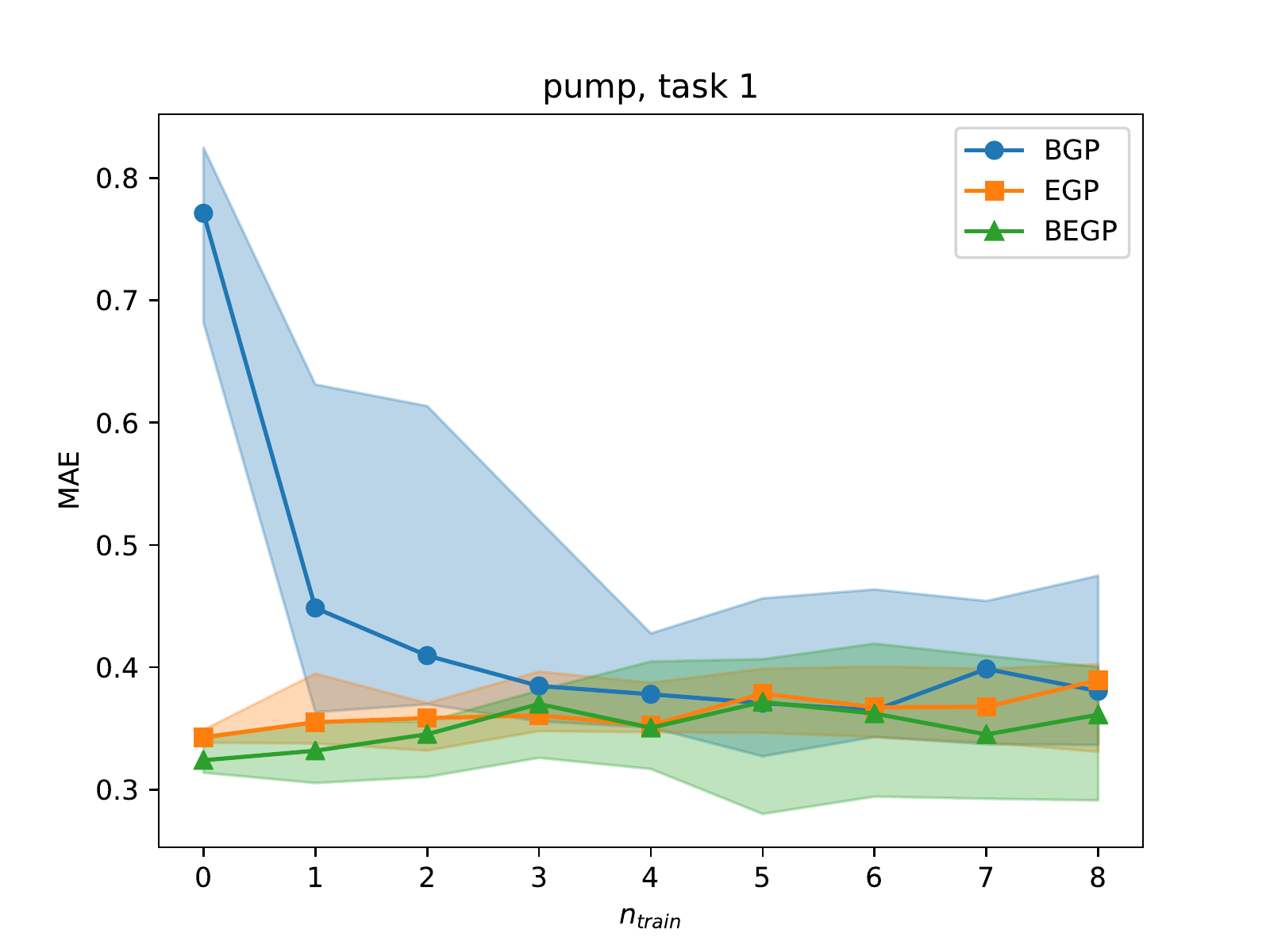}
	\includegraphics[width=0.3\textwidth]{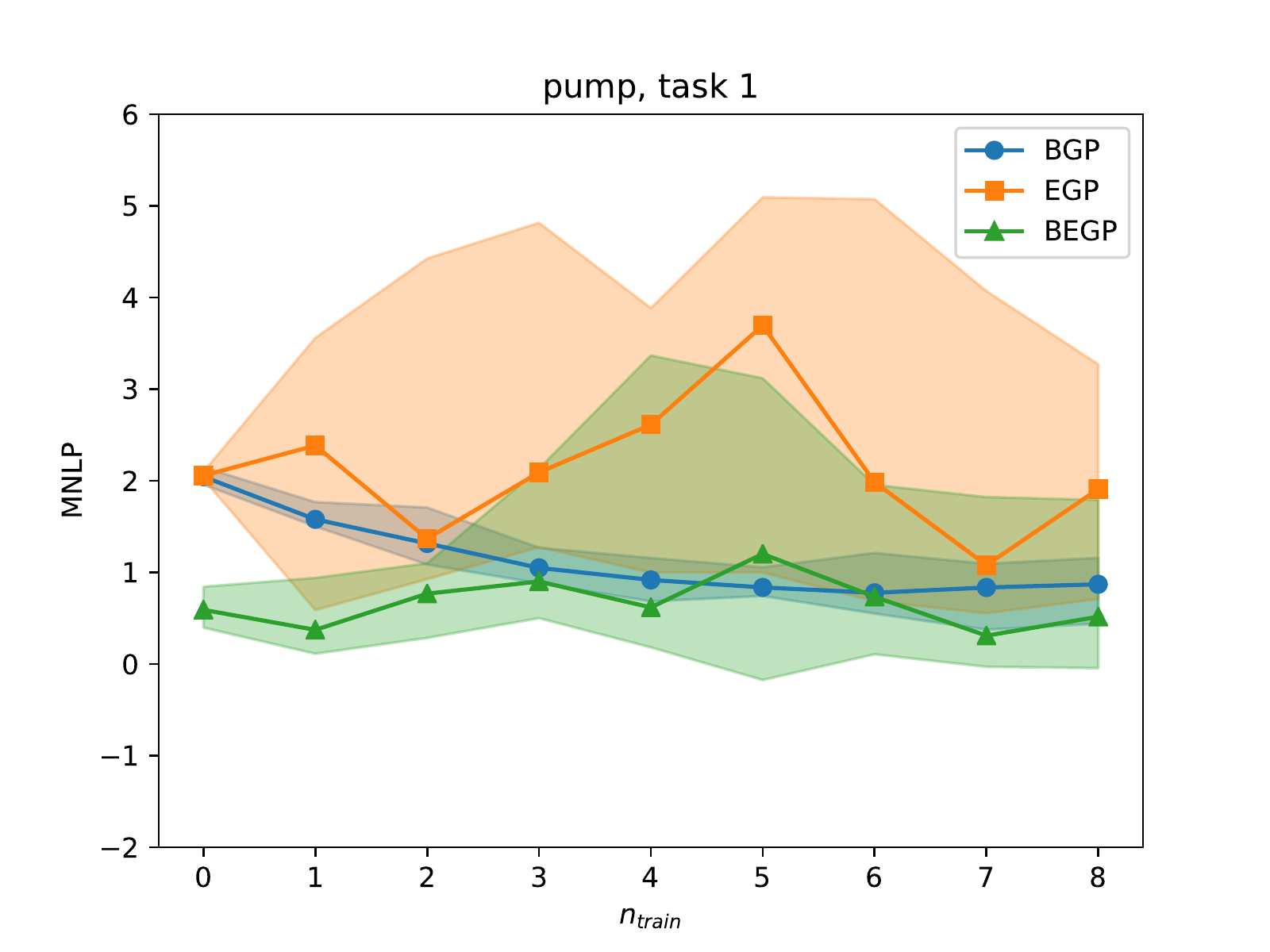}
	\\
	\includegraphics[width=0.3\textwidth]{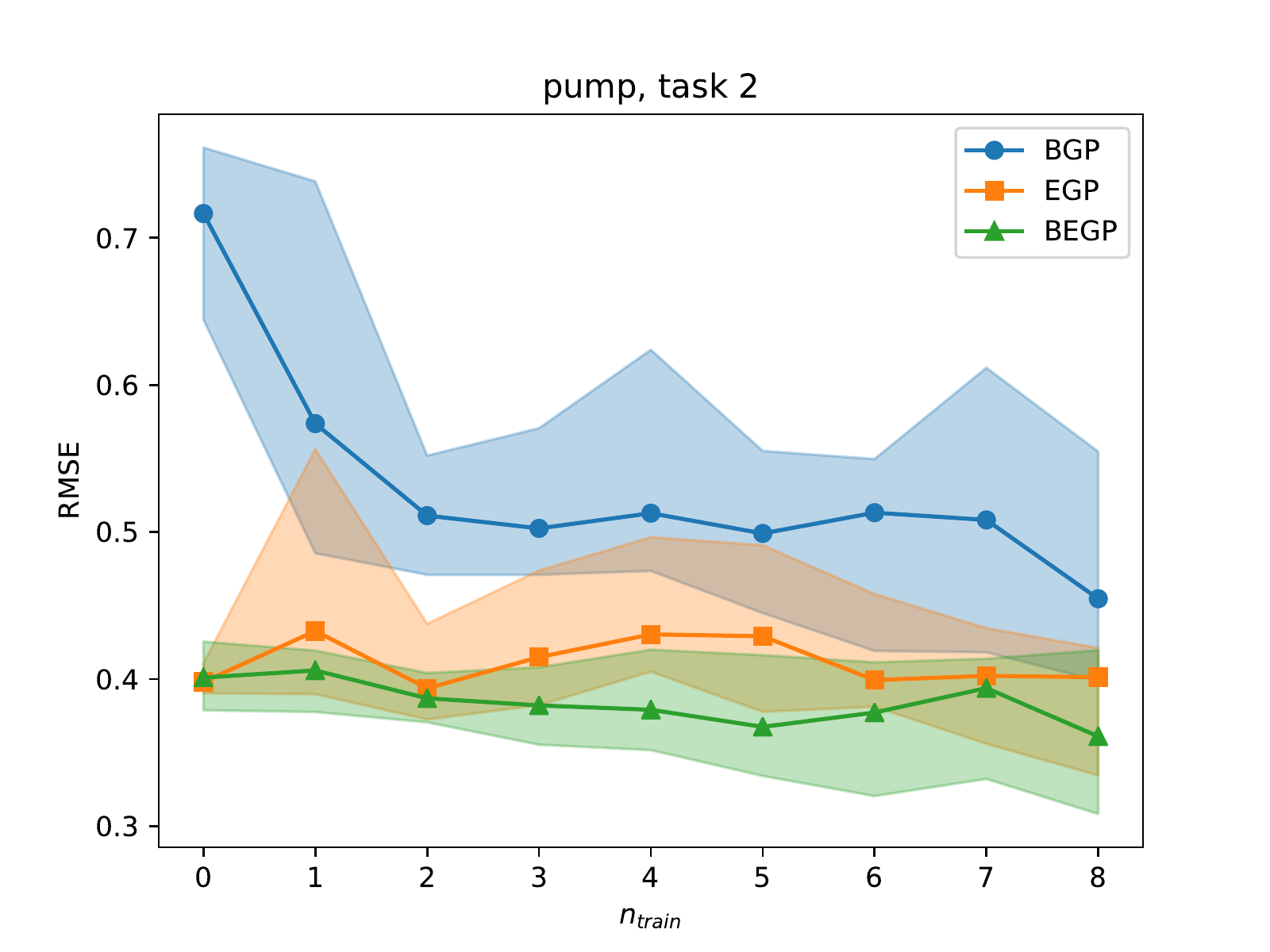}
	\includegraphics[width=0.3\textwidth]{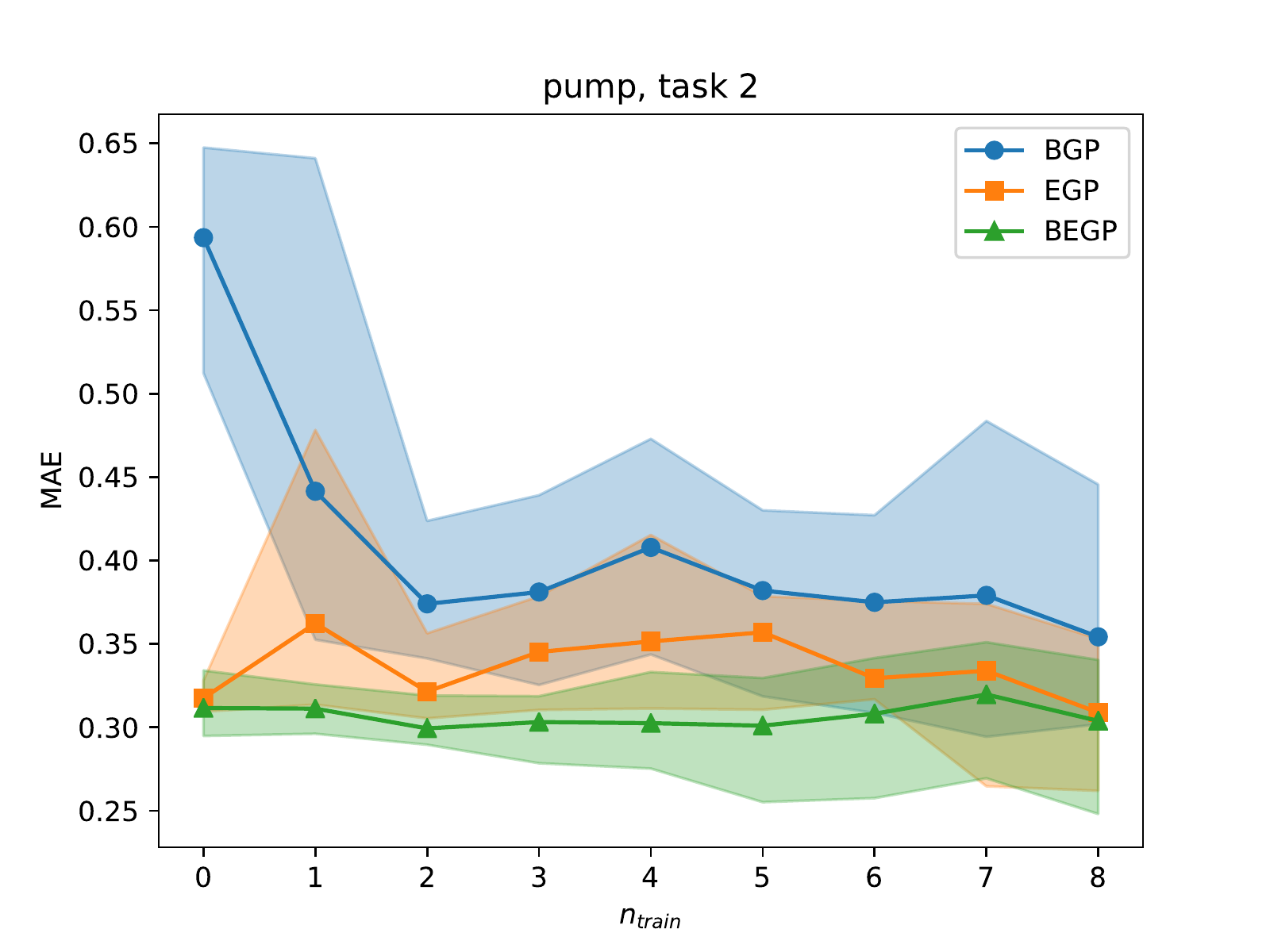}
	\includegraphics[width=0.3\textwidth]{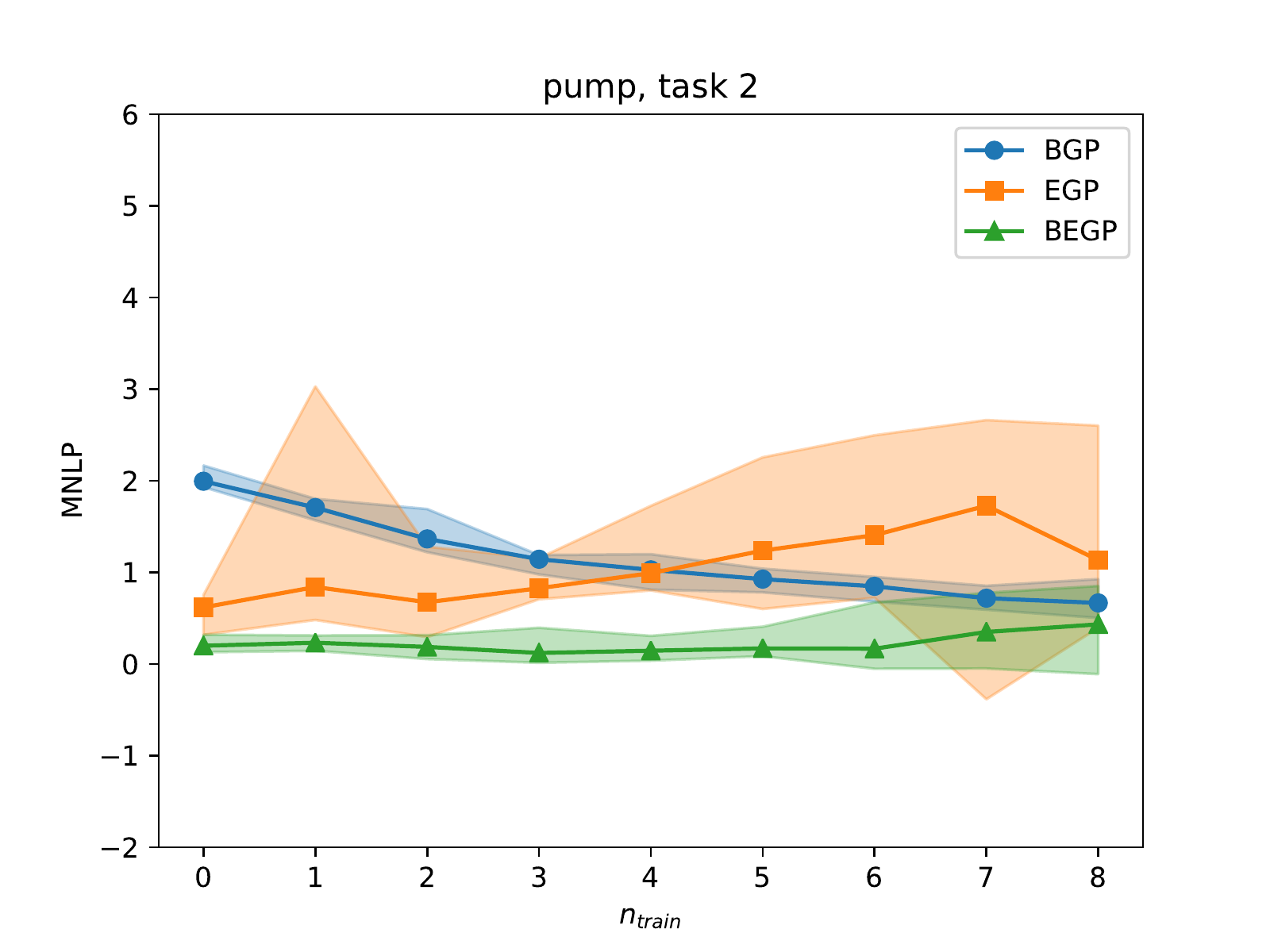}
	\\
	\includegraphics[width=0.3\textwidth]{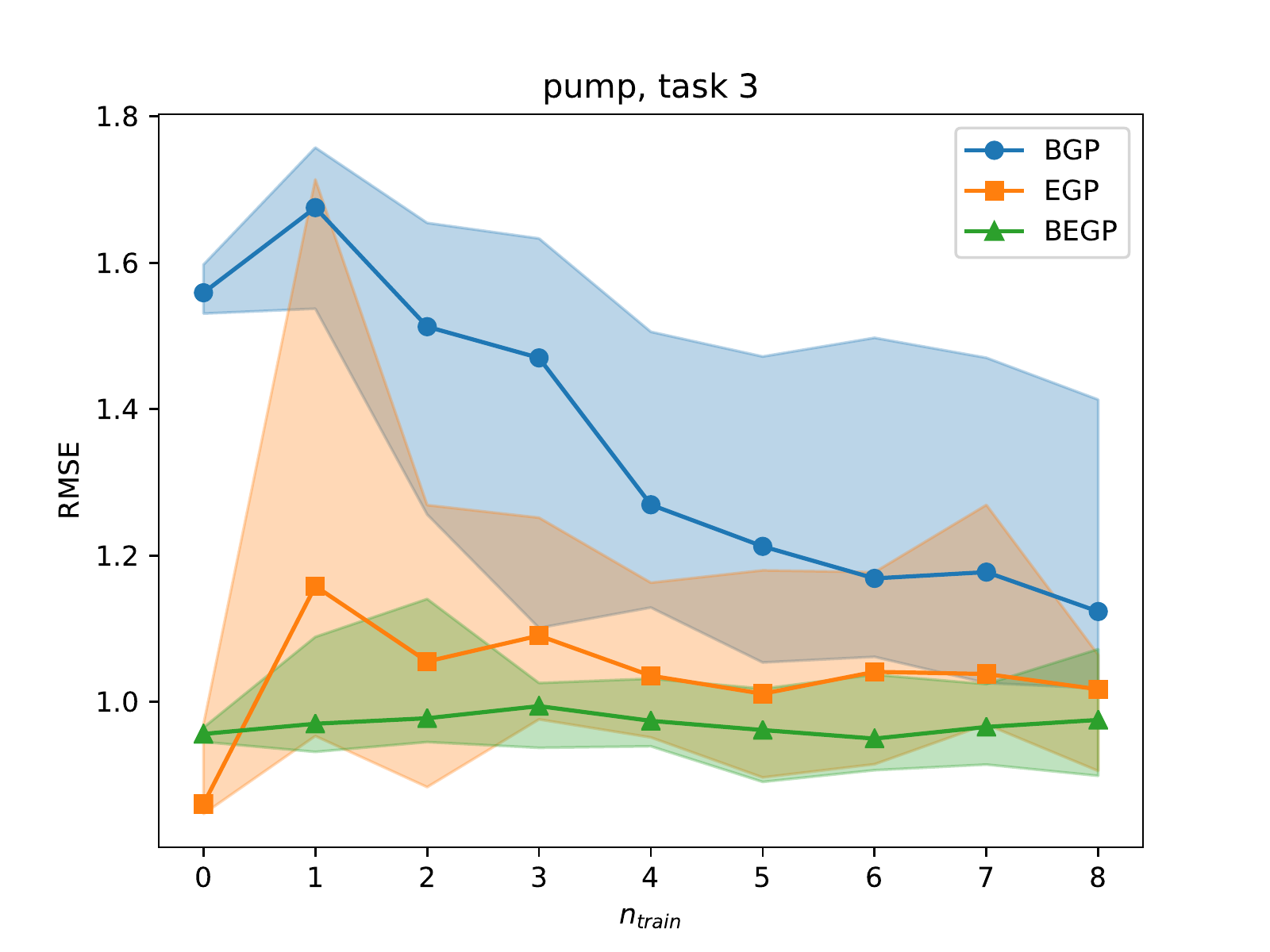}
	\includegraphics[width=0.3\textwidth]{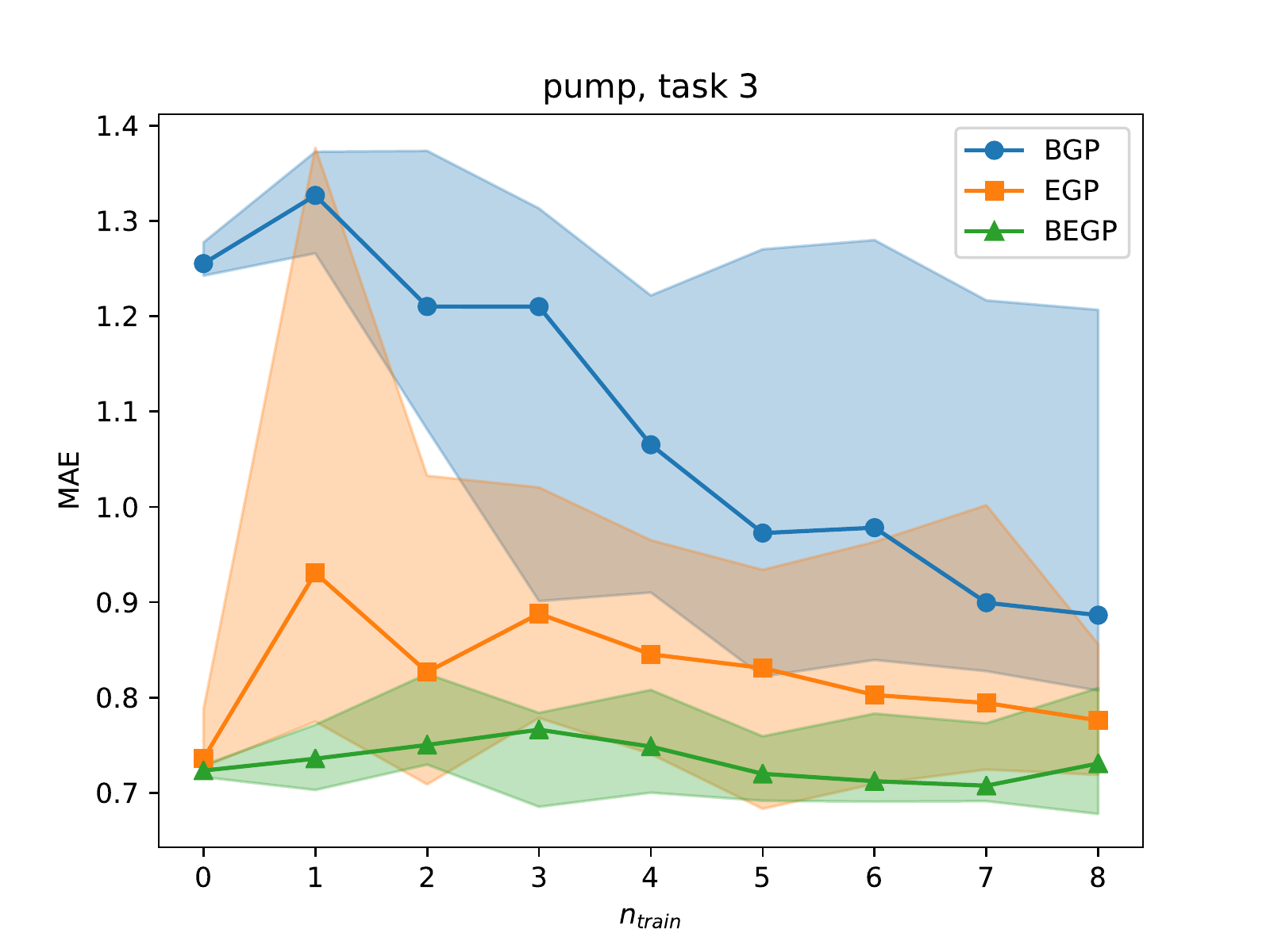}
	\includegraphics[width=0.3\textwidth]{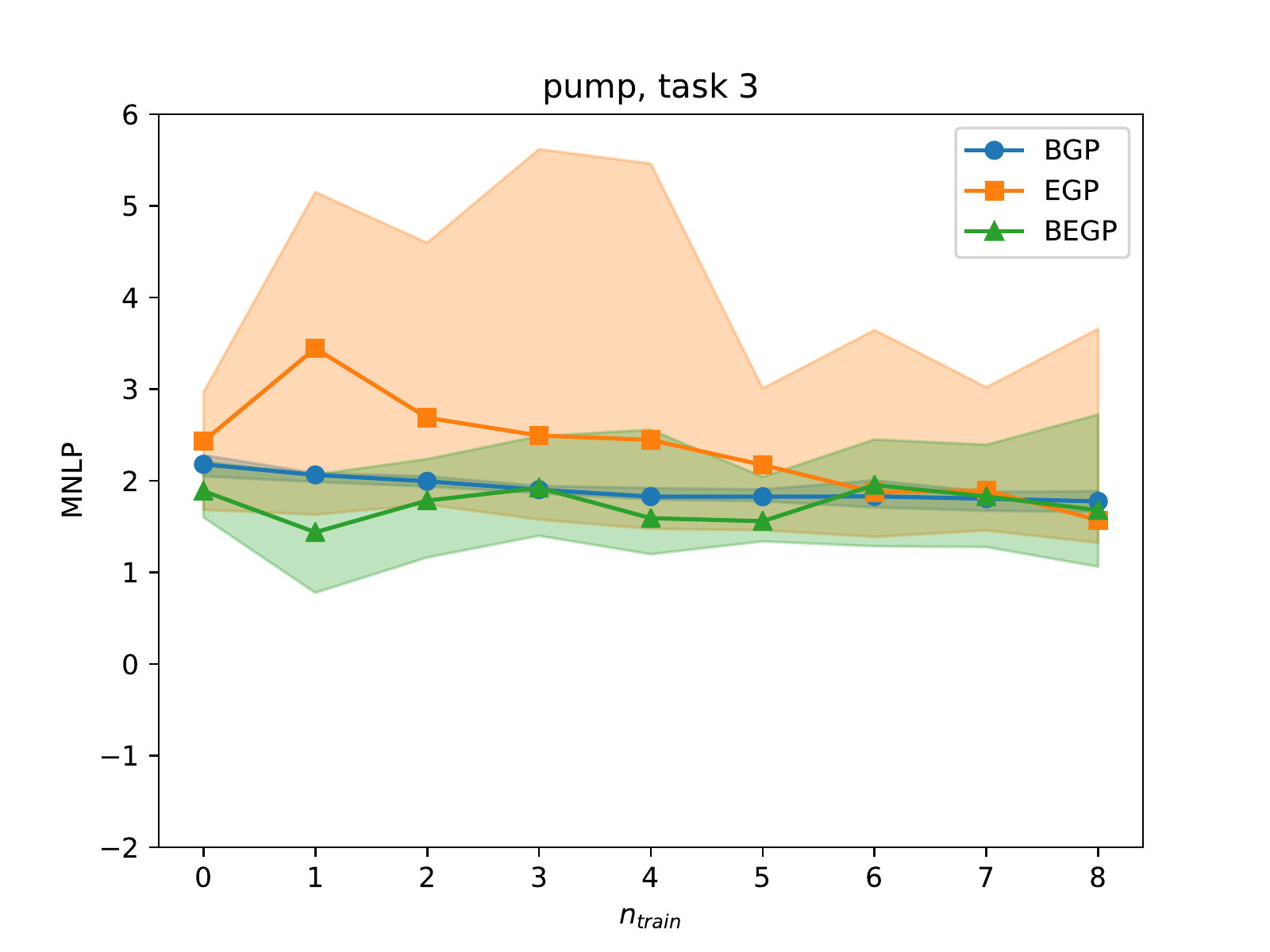}
	\\
	\includegraphics[width=0.3\textwidth]{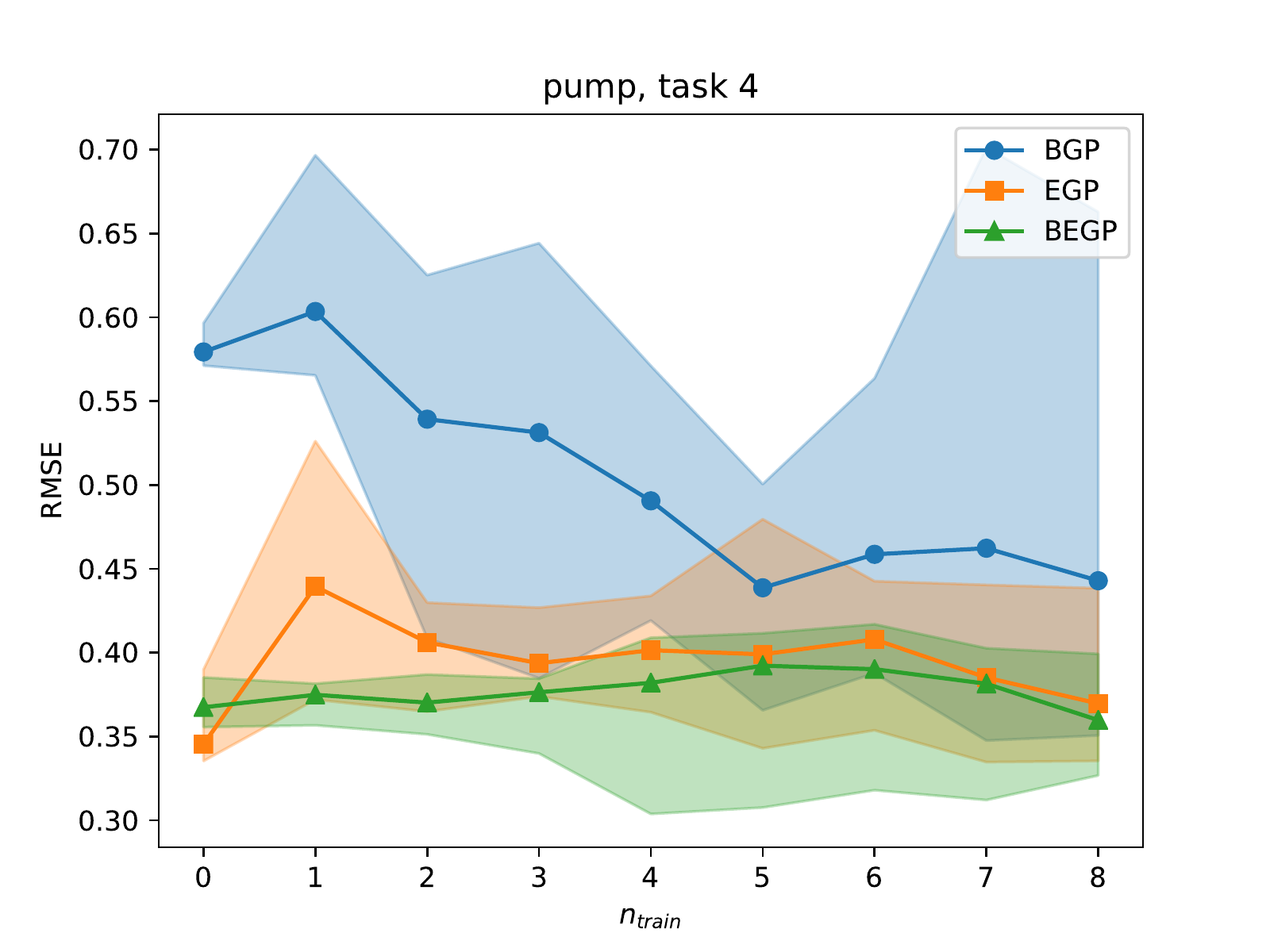}
	\includegraphics[width=0.3\textwidth]{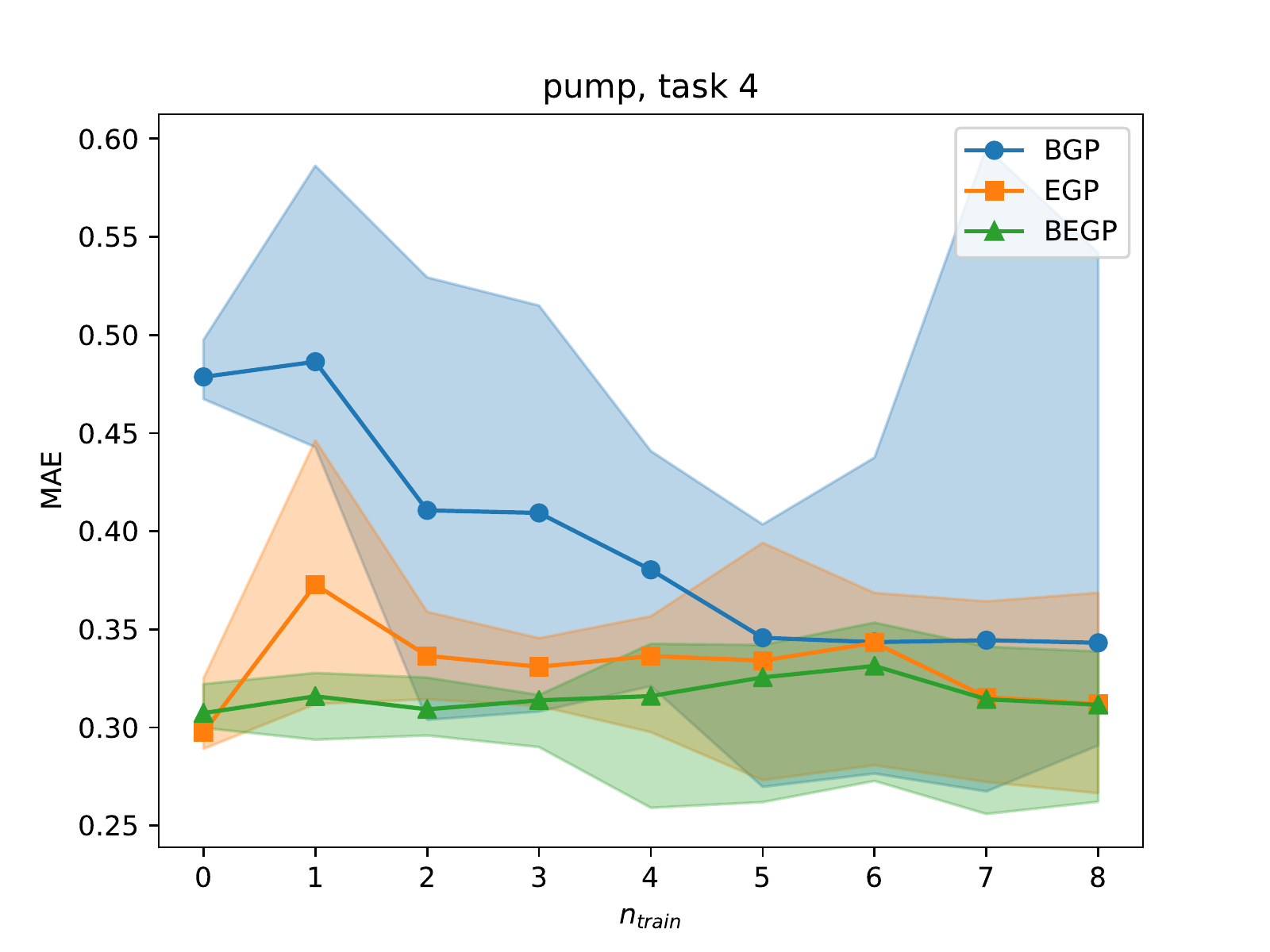}
	\includegraphics[width=0.3\textwidth]{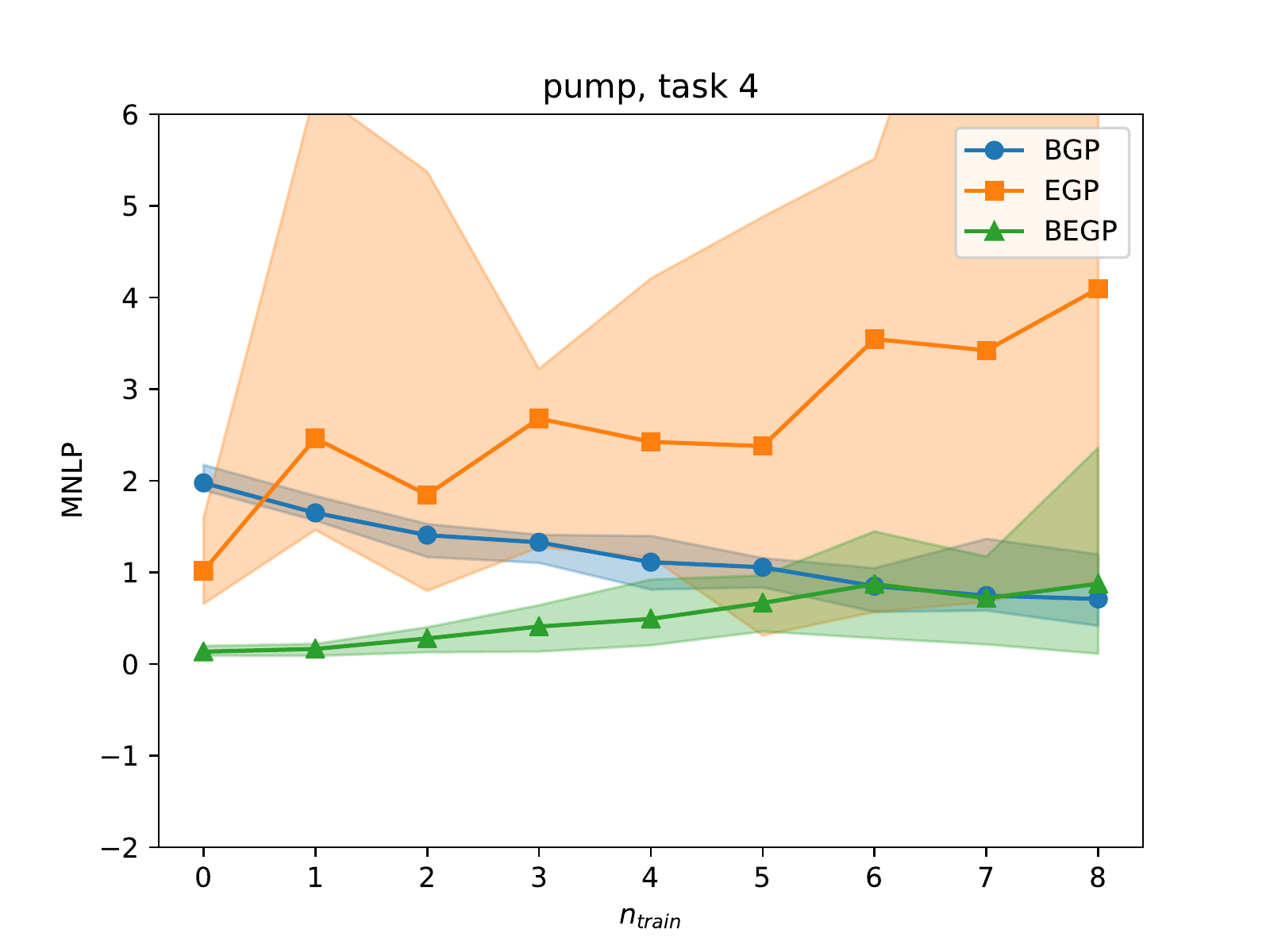}
	\caption{
		(Regression, pump dataset)
		Regression performance on held-out tasks.
		From left to right, performance in terms of root mean squared error (RMSE), mean absolute error (MAE), and median negative log probability (MNLP).
		Each row shows the performance for a different held-out task.
		Solid lines show the median performance over 10 random seeds and the shaded region shows the 80\% coverage interval.
	}
	\label{fig:examples:regression:pump}
\end{figure}

Figure \ref{fig:examples:regression:pump} shows our results for the pump dataset.
For this problem, we observe mixed results with the embedding GPs outperforming the baseline on some of the tasks, but underperforming on others. 
We hypothesize that this is because certain tasks are substantially different from the others, causing the prior information from legacy tasks to be unhelpful.
However, on certain tasks (3 and 4), we see that the Bayesian embedding GP outperforms the baseline even in a zero-shot setting.
We also notice that the deterministic embedding GP consistently gives rather poor uncertainty estimates as evidenced by its MNLP scores.
Thus, we see a clear benefit to the Bayesian embedding approach for improving the credibility of the predictions.

\begin{figure}[hb!t]
	\centering
	\includegraphics[width=0.3\textwidth]{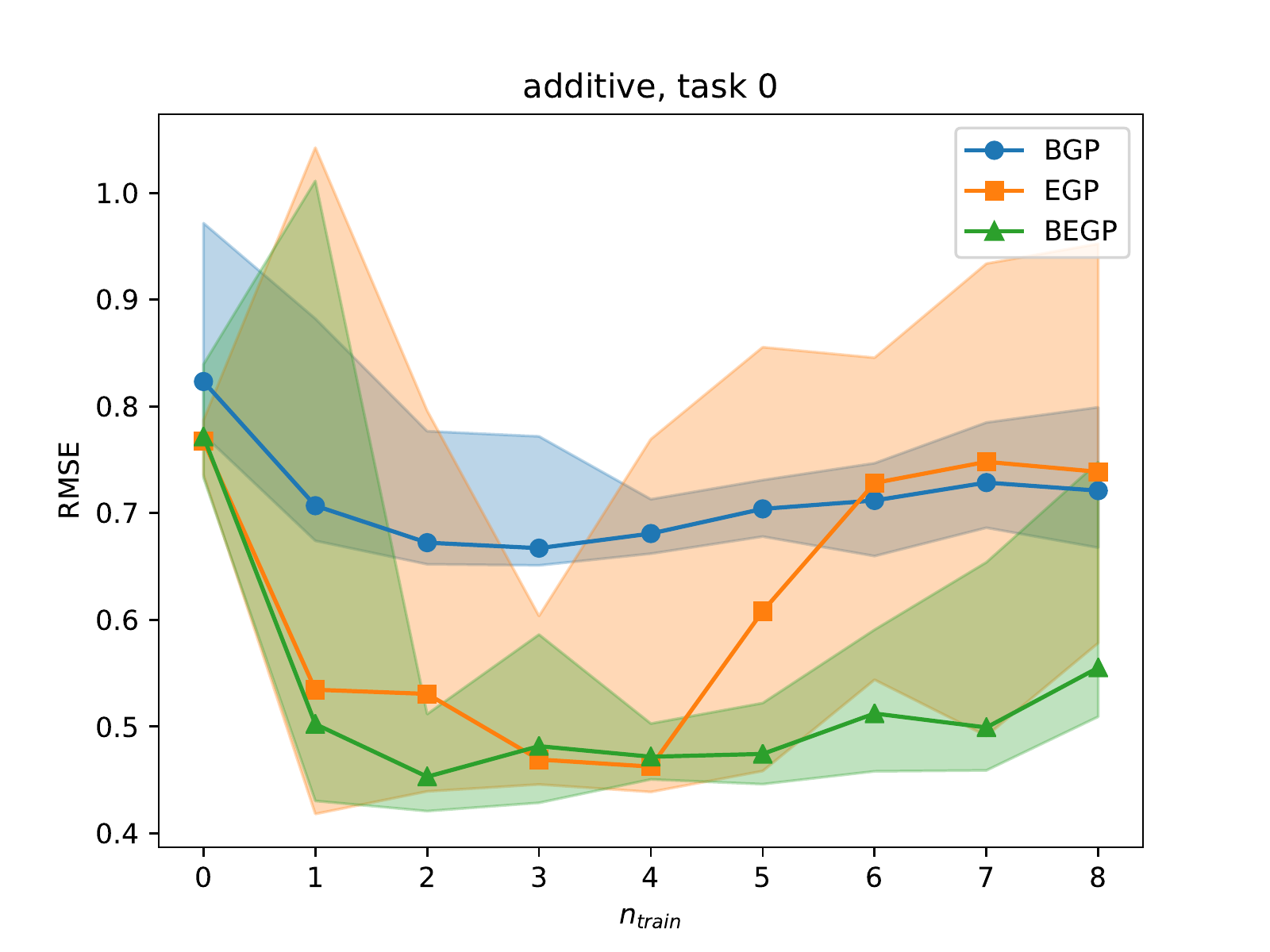}
	\includegraphics[width=0.3\textwidth]{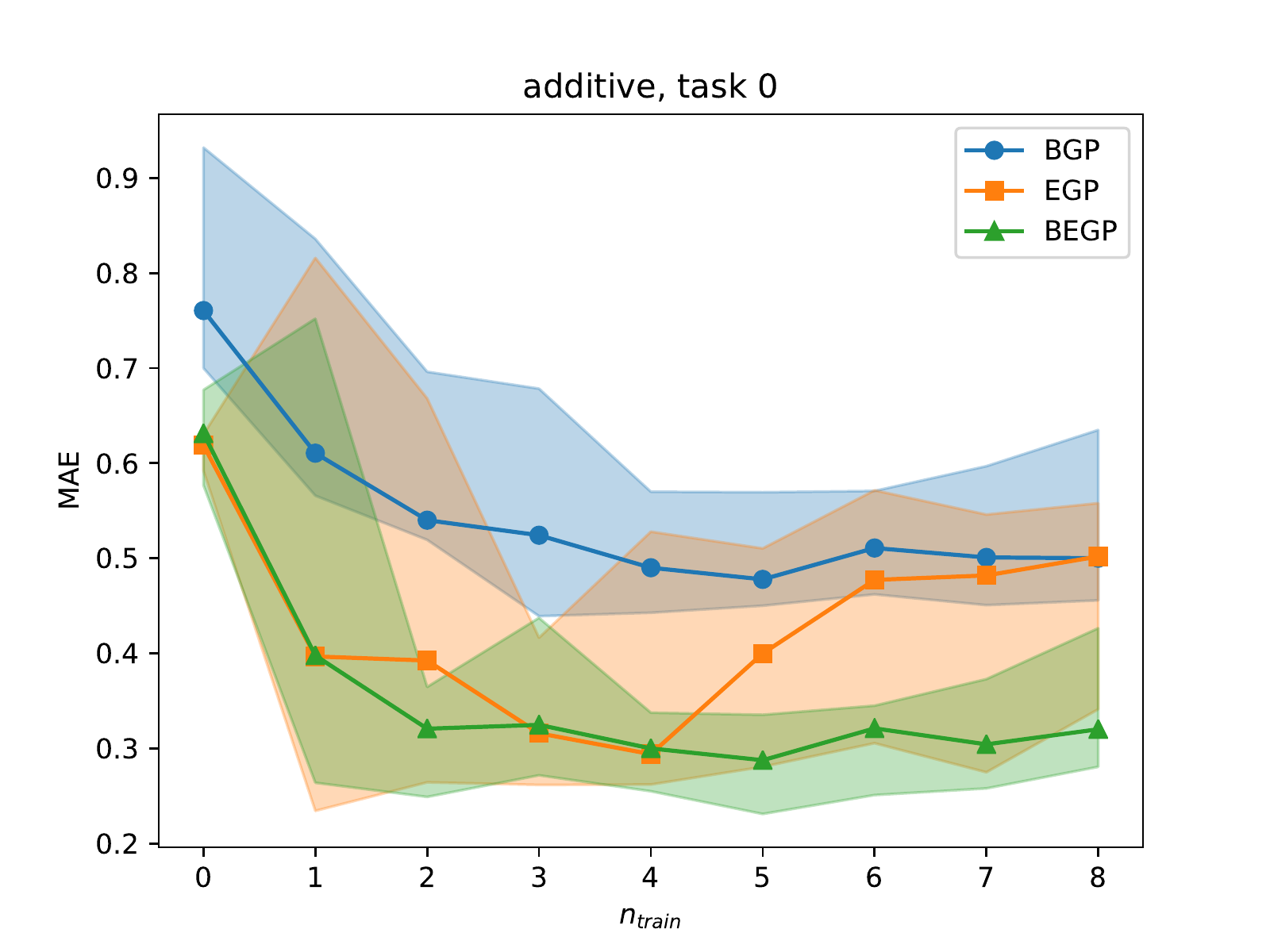}
	\includegraphics[width=0.3\textwidth]{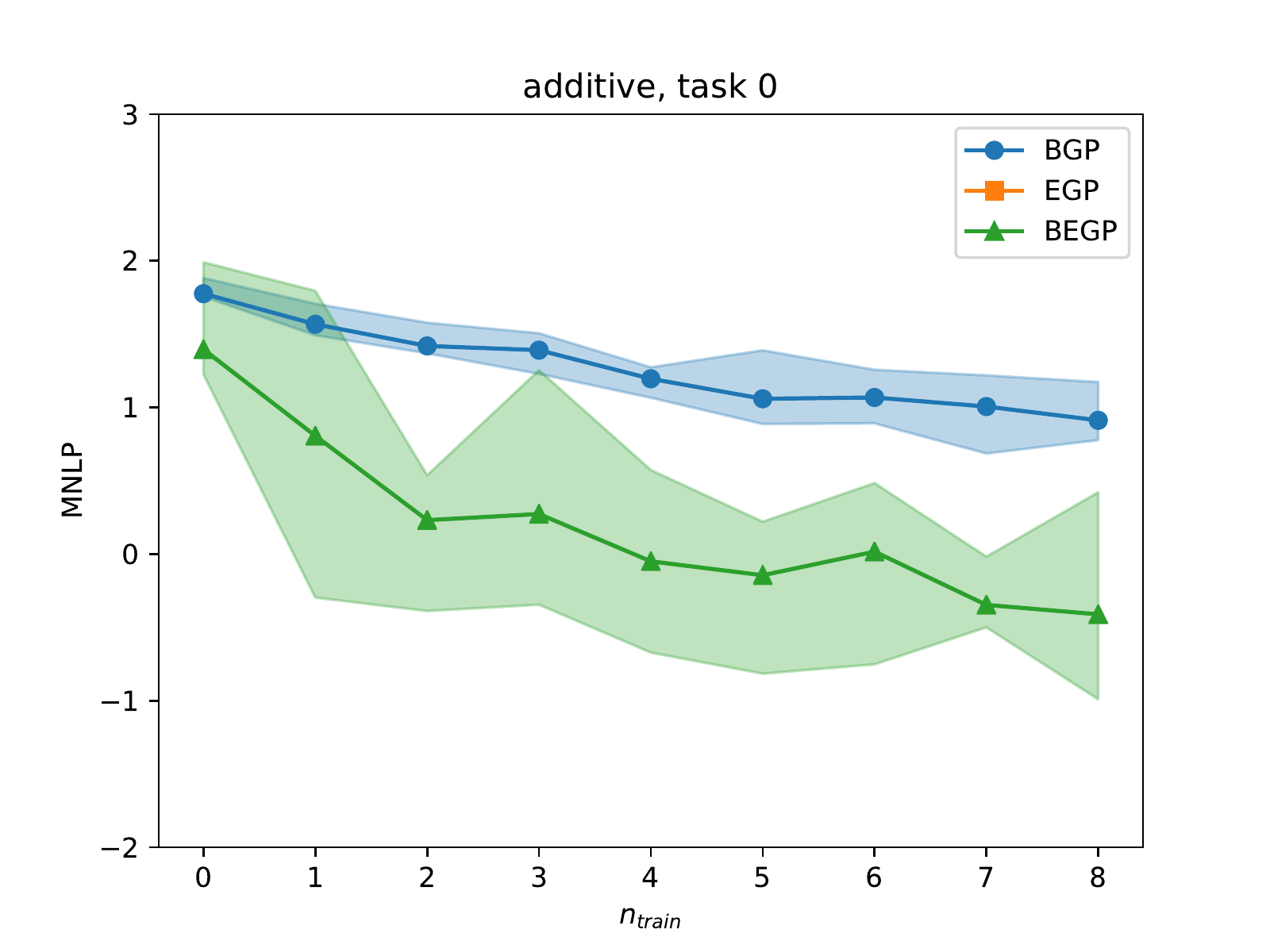}
	\\
	\includegraphics[width=0.3\textwidth]{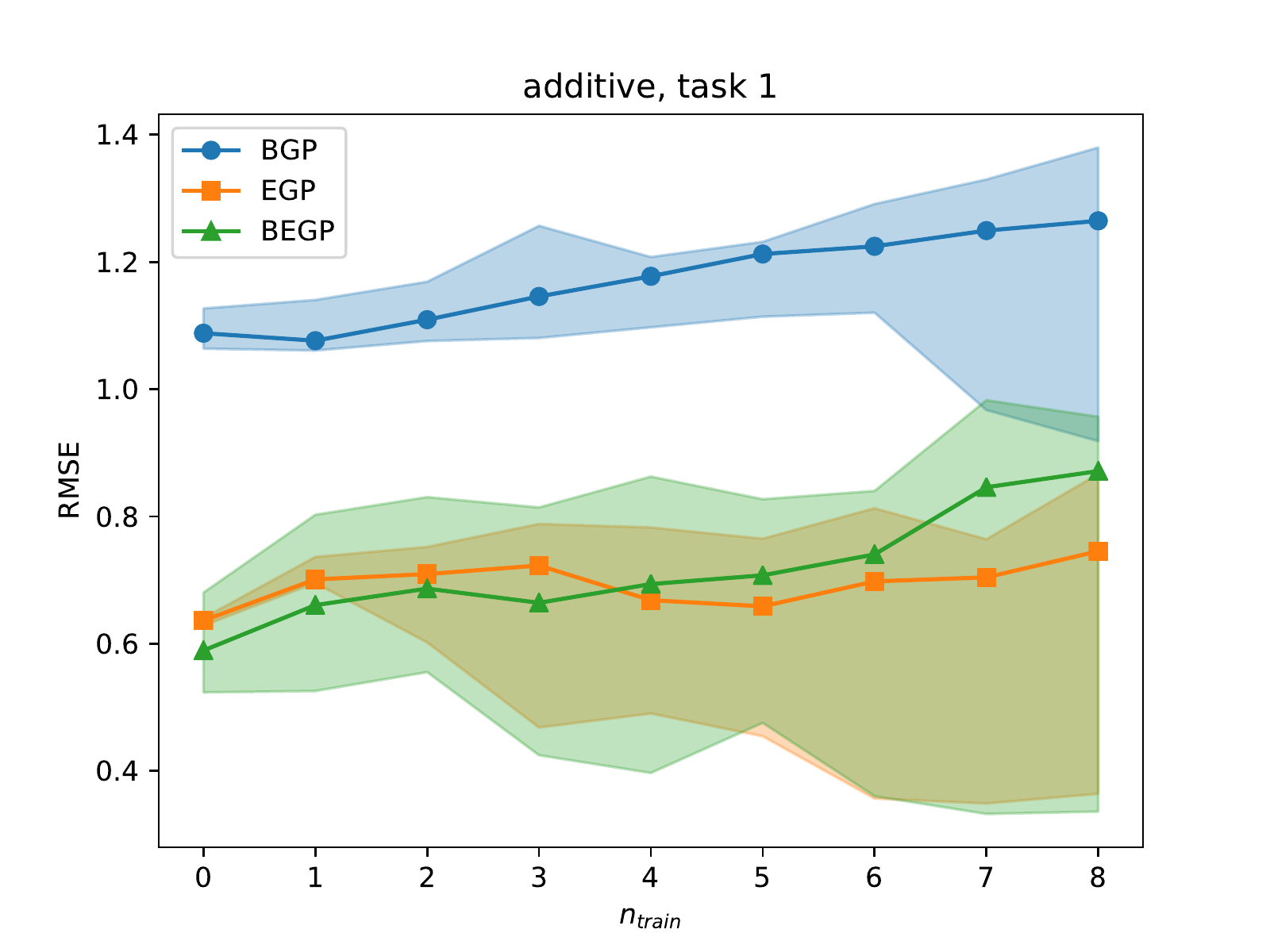}
	\includegraphics[width=0.3\textwidth]{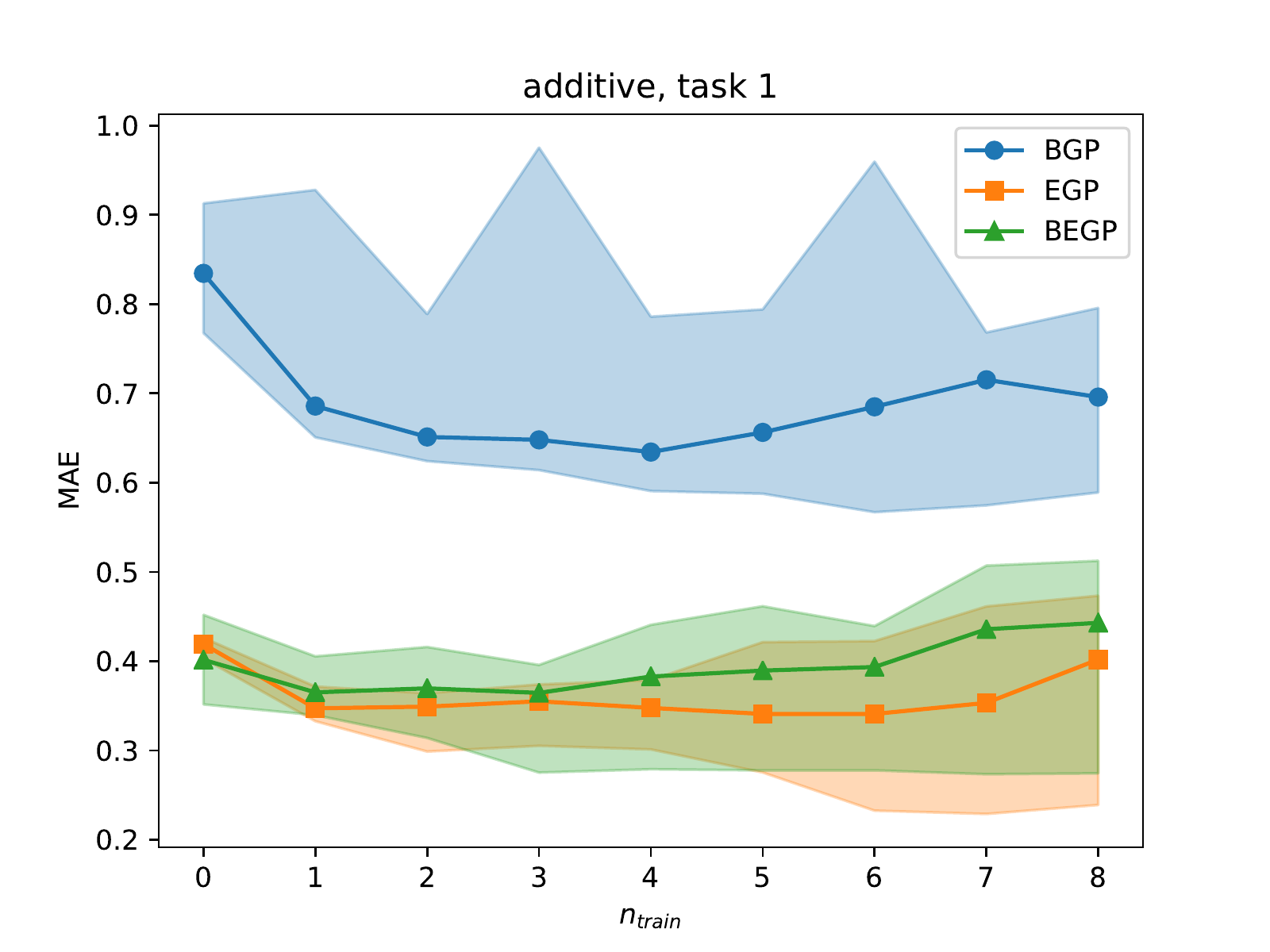}
	\includegraphics[width=0.3\textwidth]{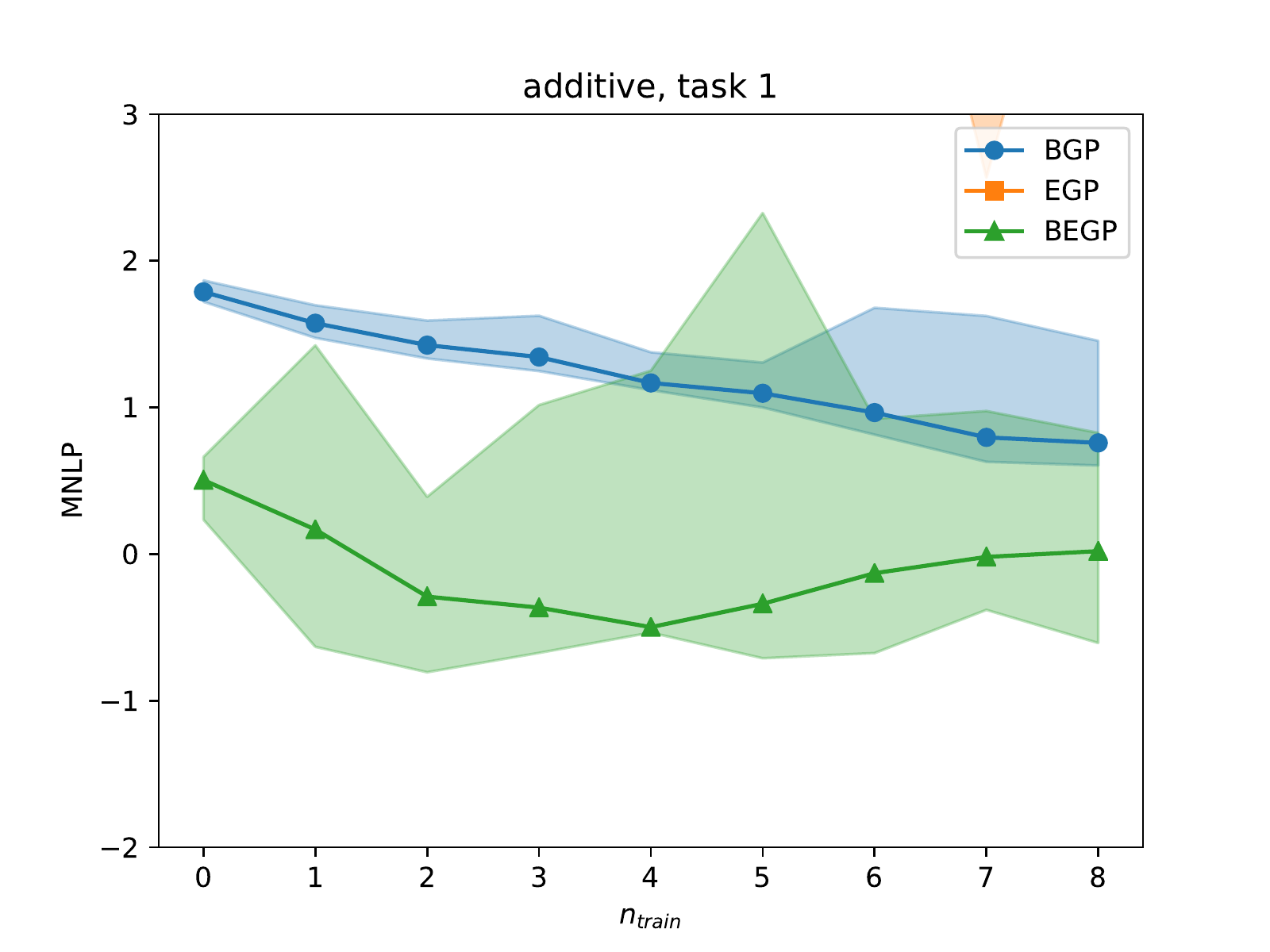}
	\\
	\includegraphics[width=0.3\textwidth]{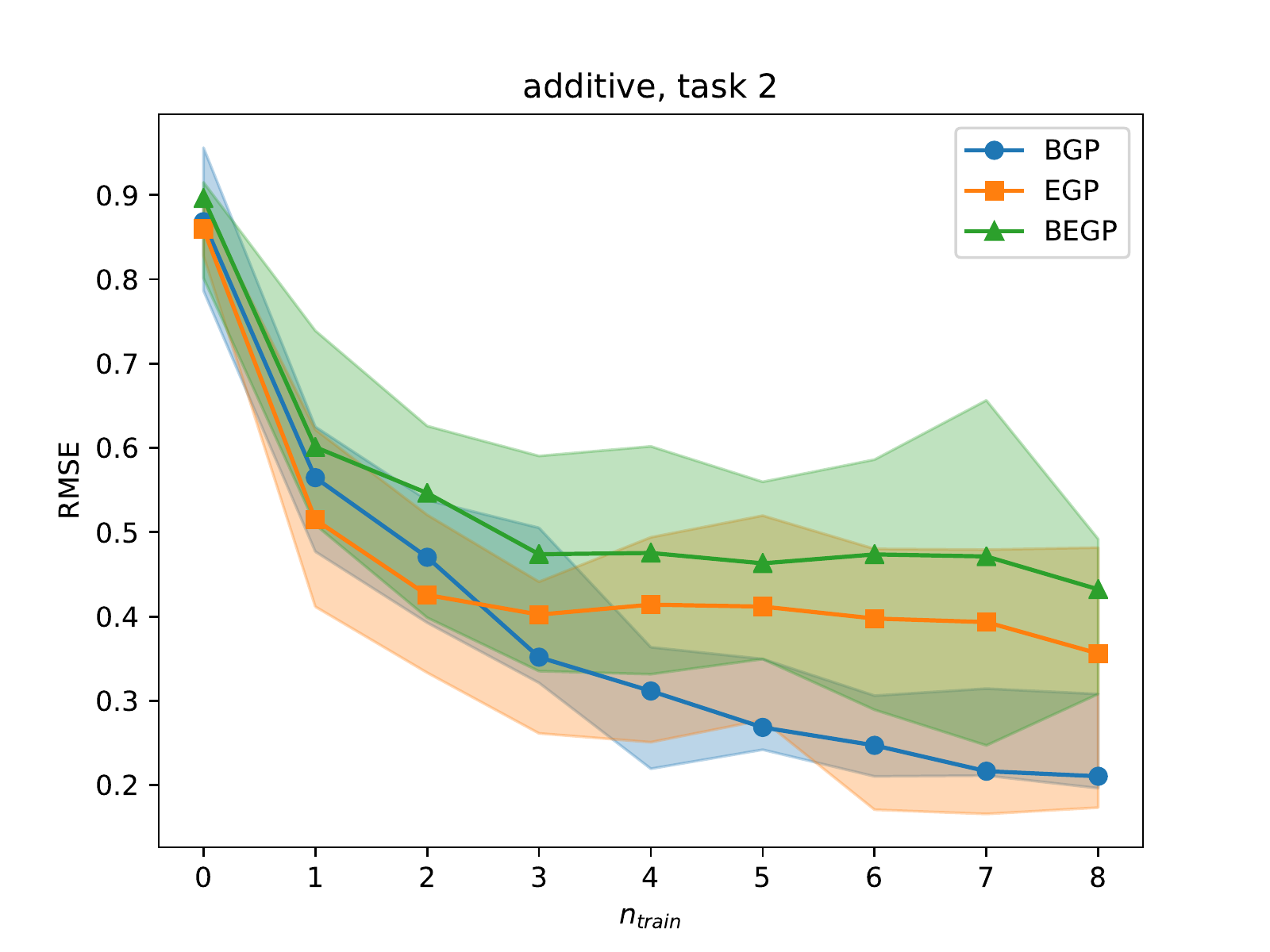}
	\includegraphics[width=0.3\textwidth]{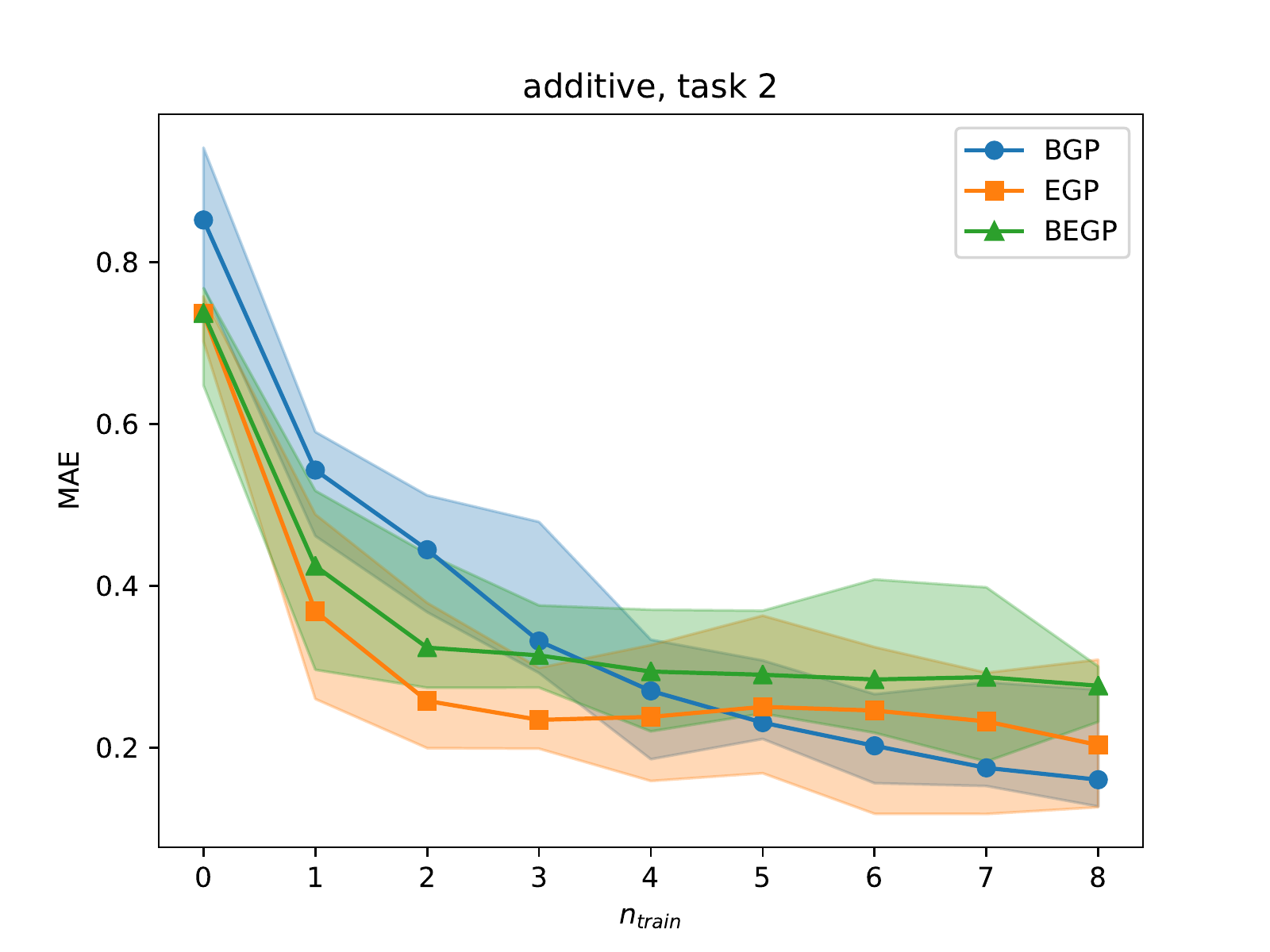}
	\includegraphics[width=0.3\textwidth]{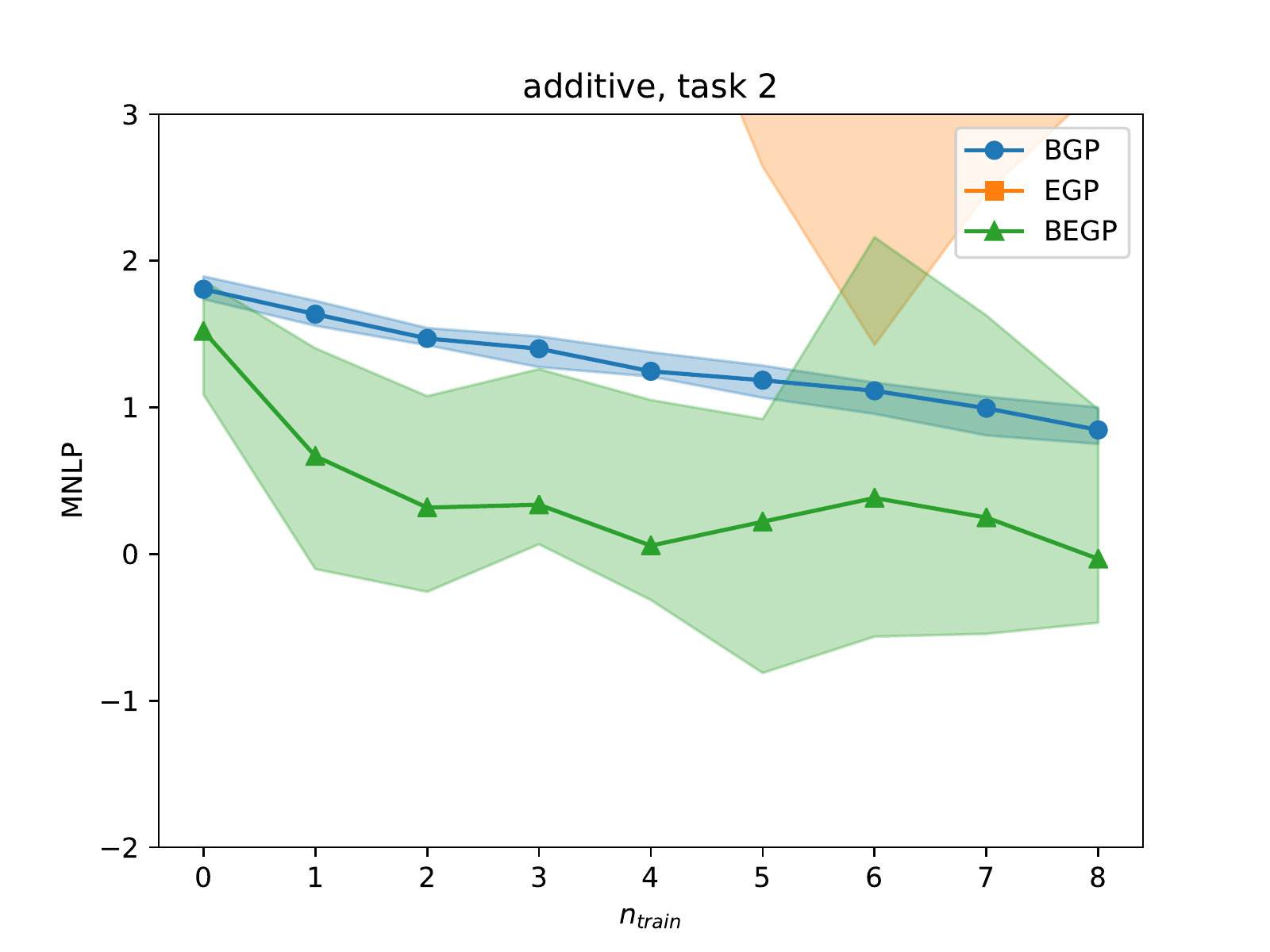}
	\\
	\includegraphics[width=0.3\textwidth]{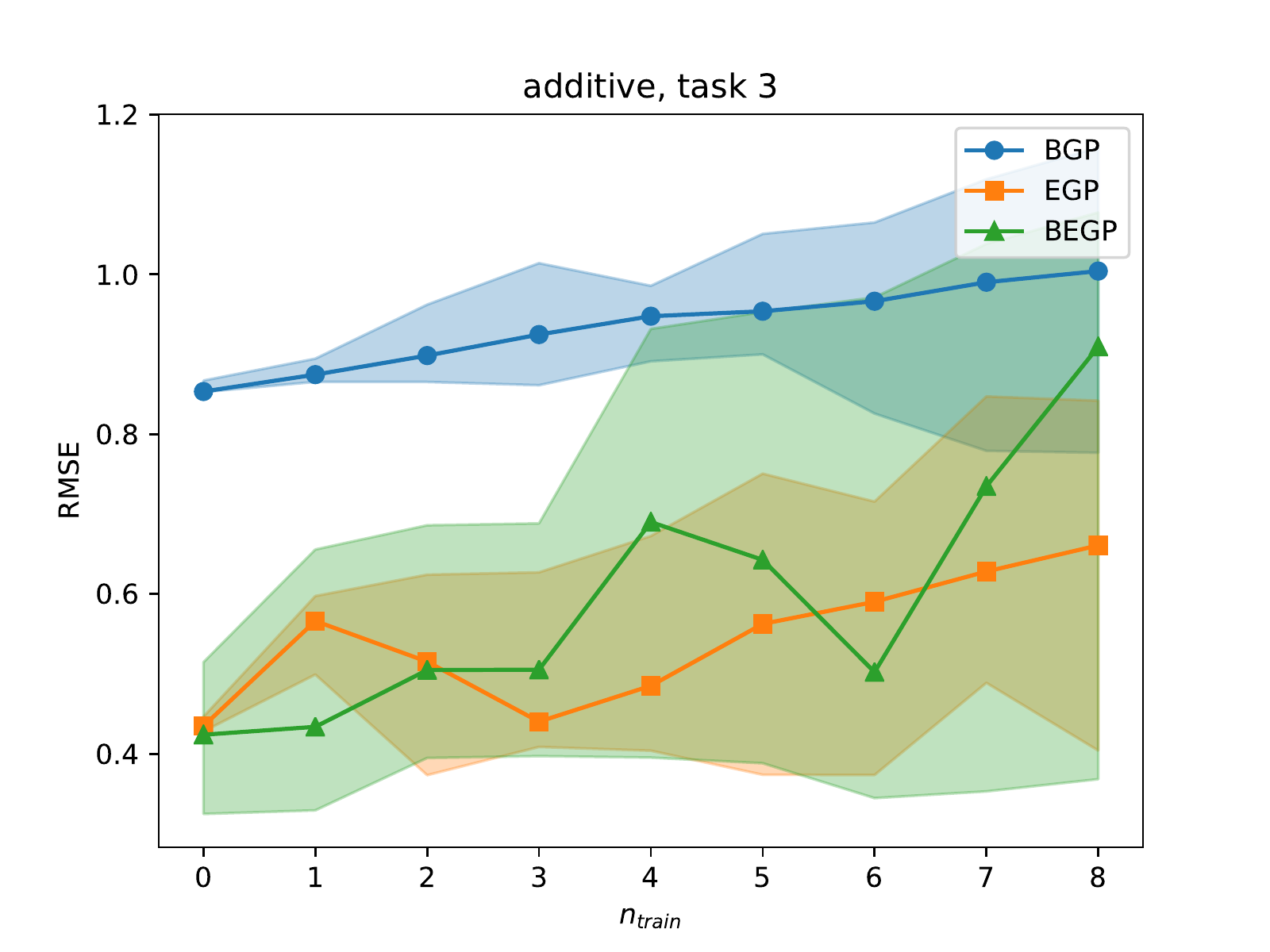}
	\includegraphics[width=0.3\textwidth]{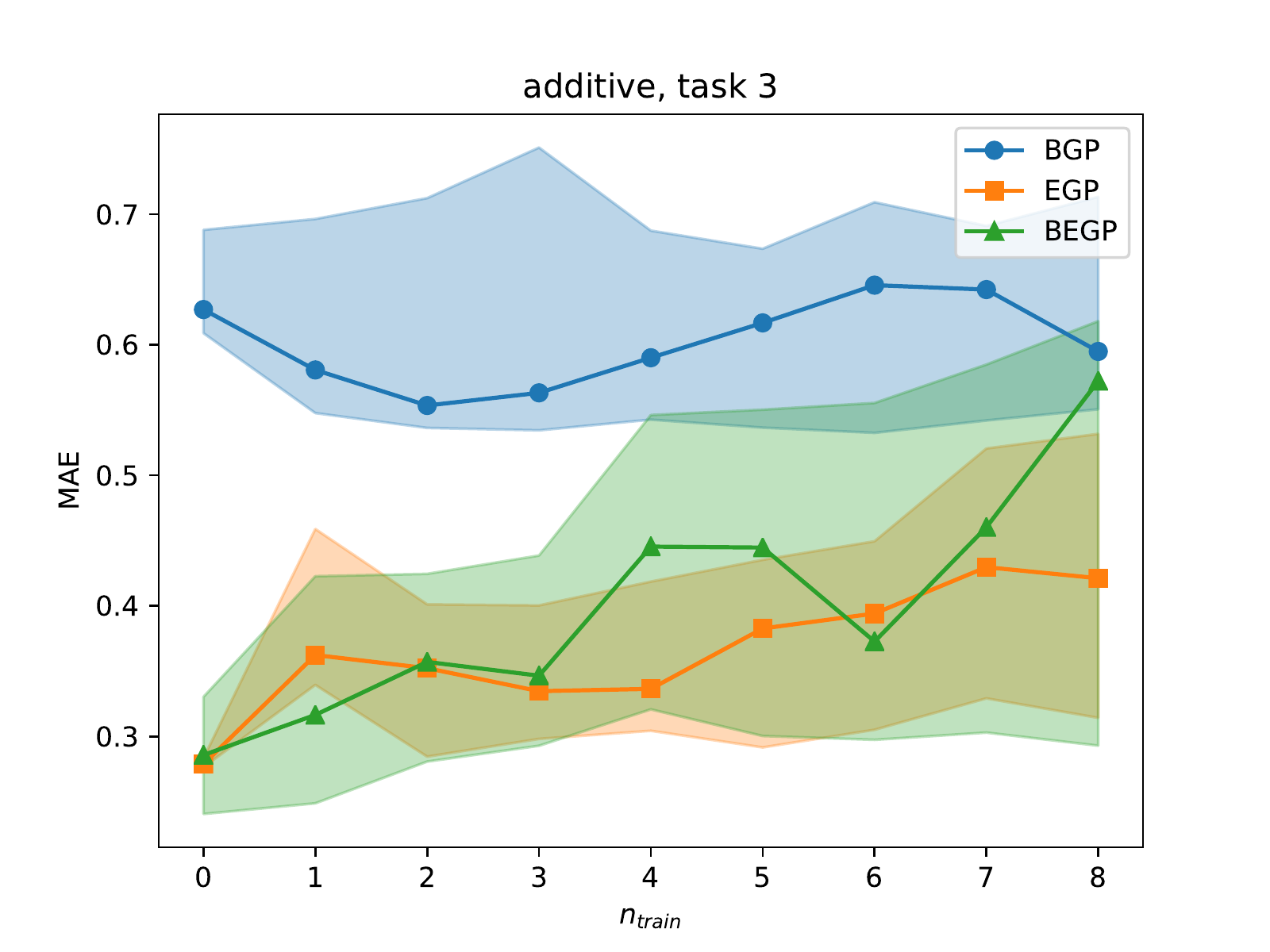}
	\includegraphics[width=0.3\textwidth]{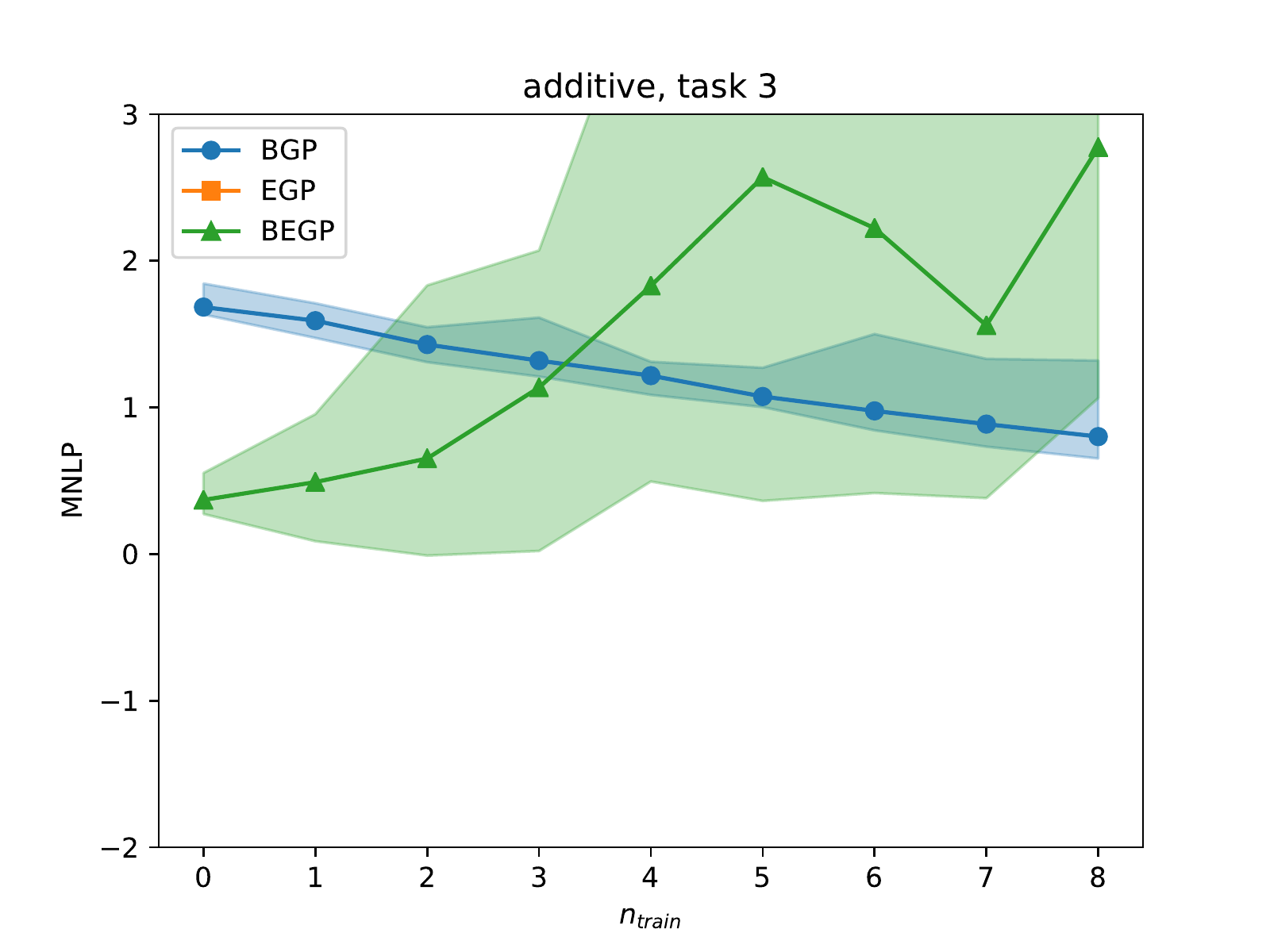}
	\\
	\includegraphics[width=0.3\textwidth]{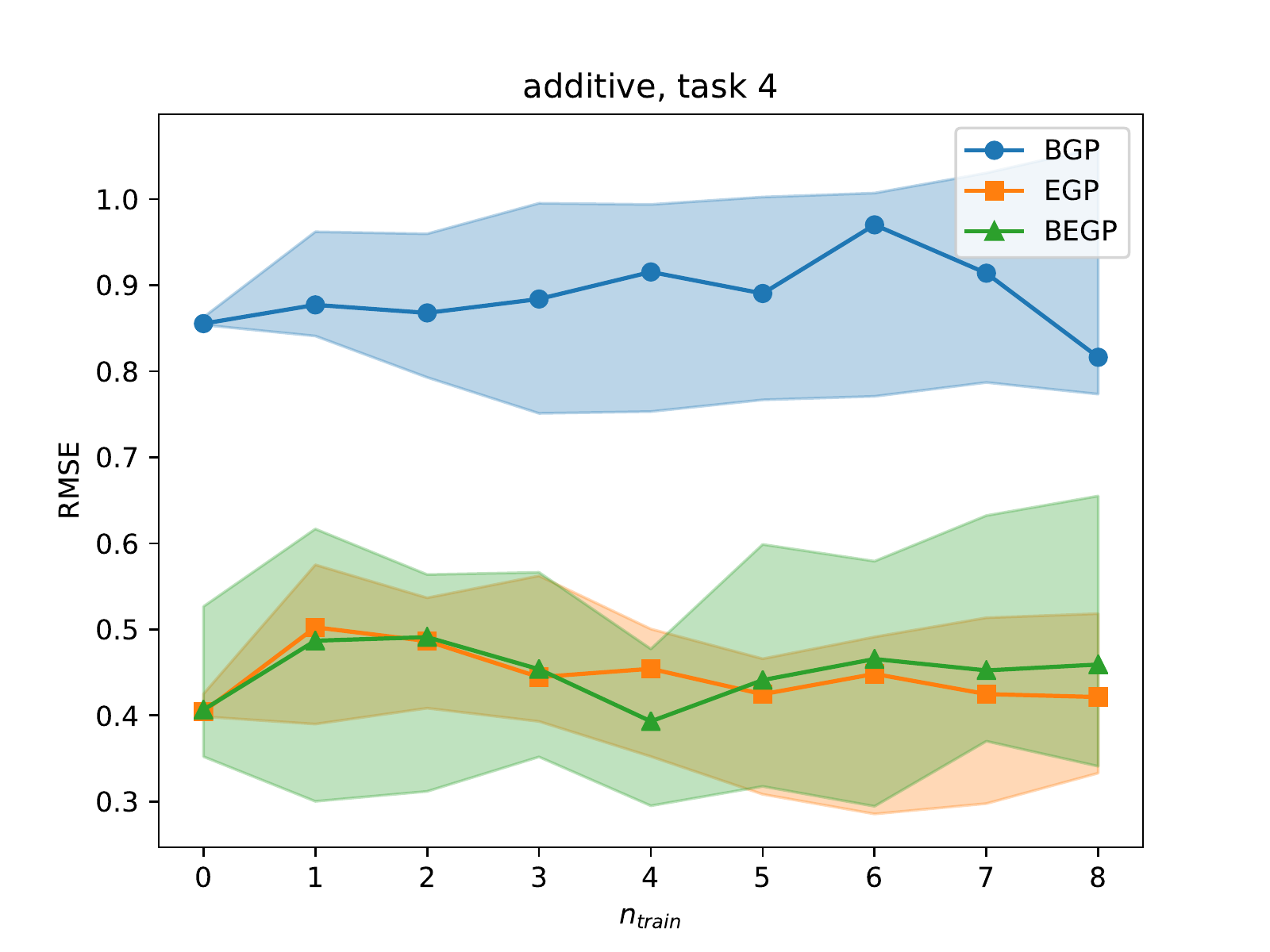}
	\includegraphics[width=0.3\textwidth]{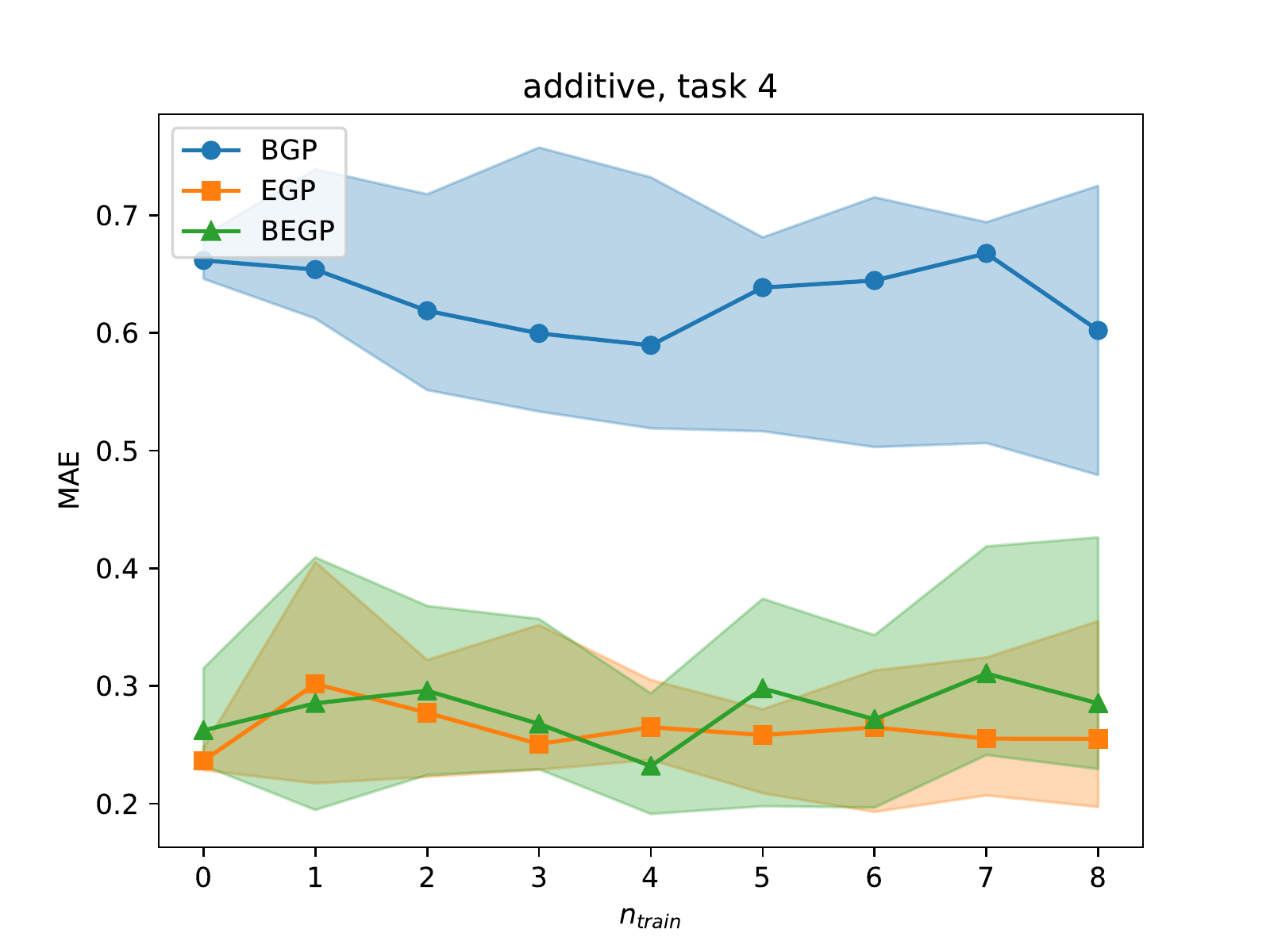}
	\includegraphics[width=0.3\textwidth]{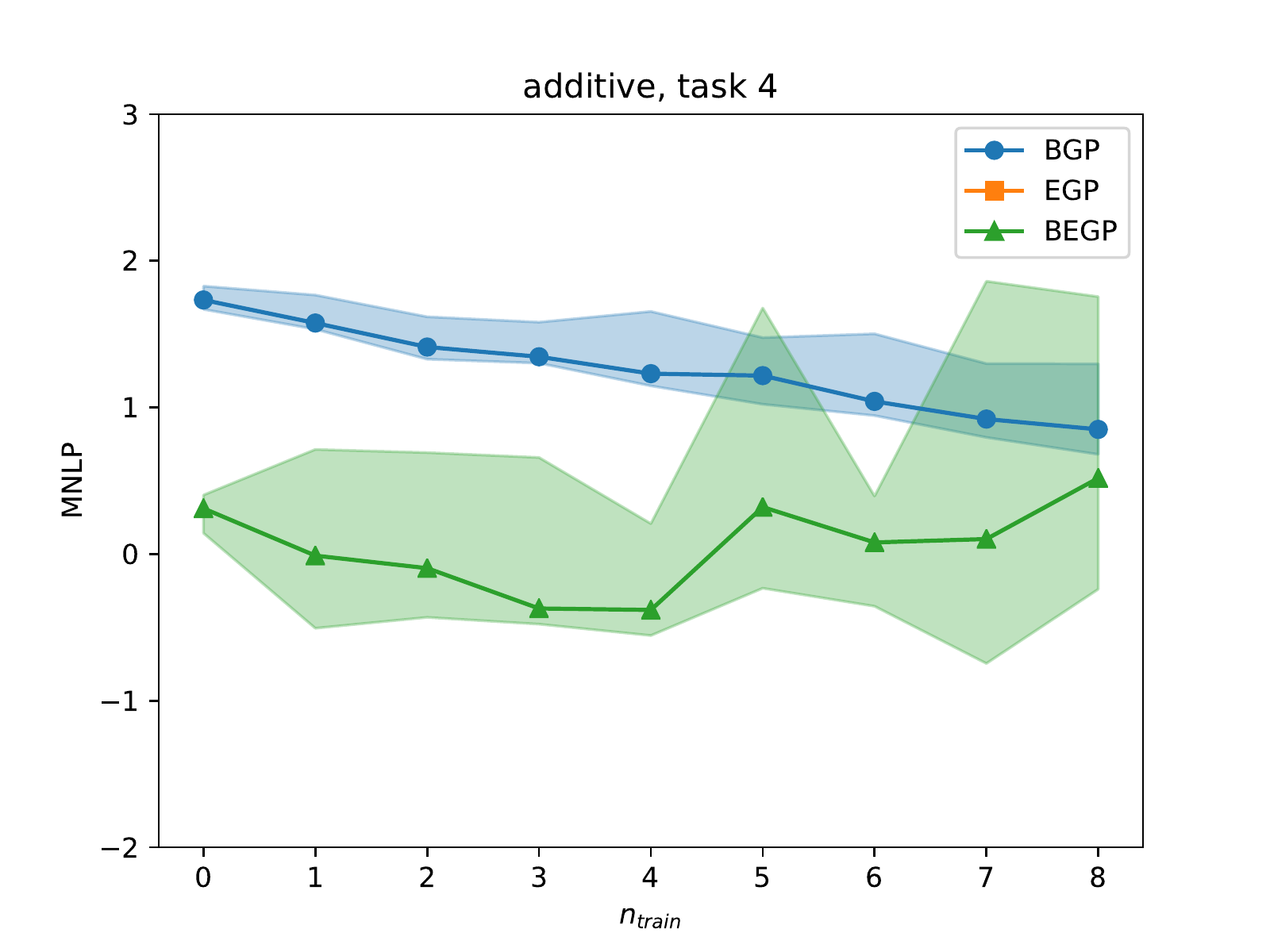}
	\caption{
		(Regression, additive dataset)
		Regression performance on held-out tasks.
		From left to right, performance in terms of root mean squared error (RMSE), mean absolute error (MAE), and median negative log probability (MNLP).
		Each row shows the performance for a different held-out task.
		Solid lines show the median performance over 10 random seeds and the shaded region shows the 80\% coverage interval.
	}
	\label{fig:examples:regression:additive}
\end{figure}

Figure \ref{fig:examples:regression:additive} shows our results for the additive dataset for the first few held-out tasks.
In contrast to the pump problem, we often see considerable improvements by using the embedding models.
For many of the held-out tasks, the \textit{zero-shot} predictions from the embedding models are, on average, superior to those of the Bayesian GP even after it has seen 10 examples from the current task.
On others (e.g.\ RMSE, MAE for Task 2 in Fig. \ref{fig:examples:regression:additive}), the Bayesian GP does somewhat outperform the embedding models, though the embedding models continue to improve.
This may be because the prior knowledge built up from the legacy tasks is unhelpful for the held-out task; the embedding model must learn to ``overturn'' this prior knowledge by finding a way to separate the embedding of the held-out task from the unrelated legacy tasks.
However, this requires observations to gradually overturn the belief via Bayesian updating.

\begin{figure}[hbt]
	\centering
	\includegraphics[width=0.5\textwidth]{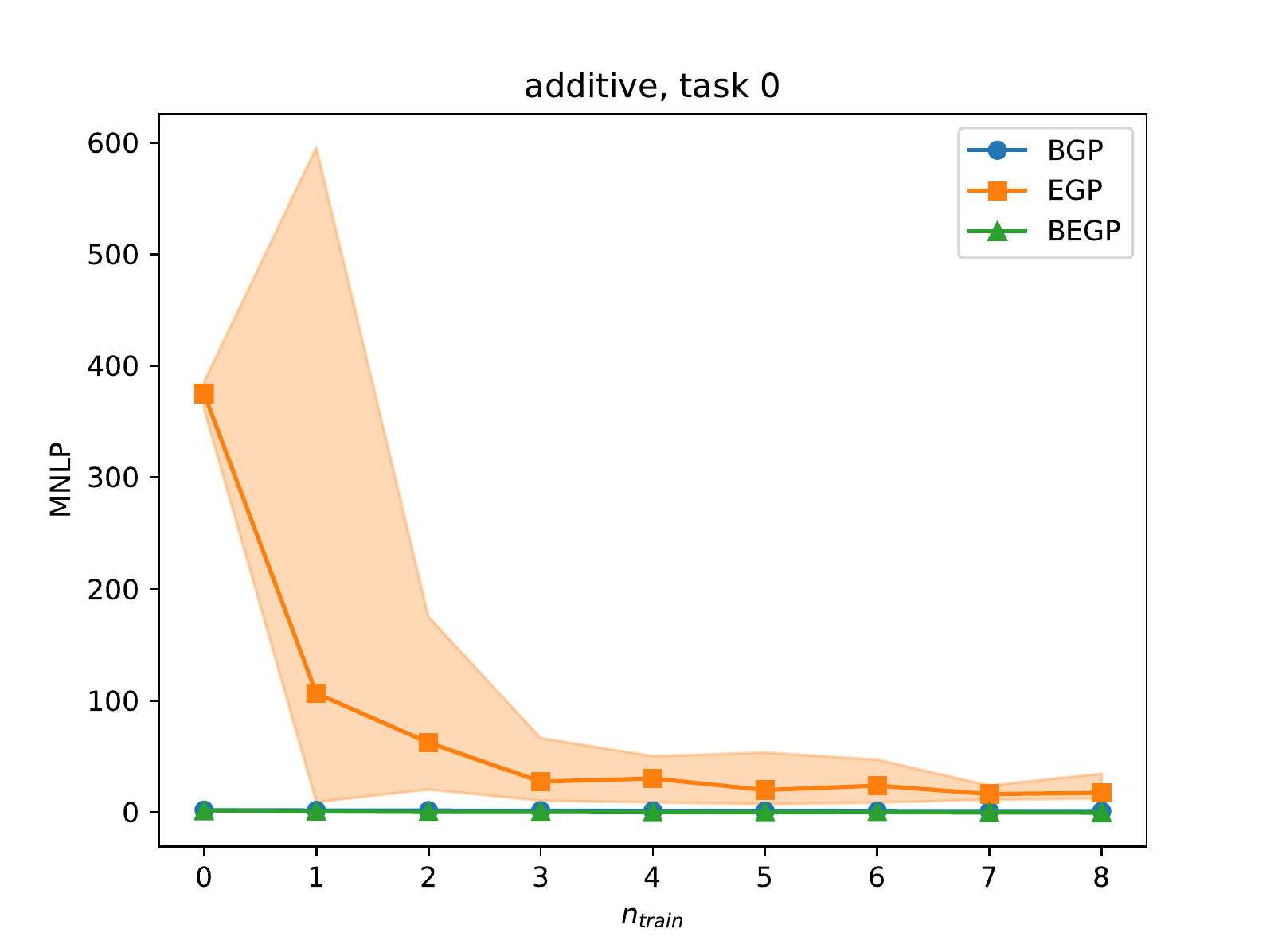}
	\caption{
		(Regression, additive dataset)
		Zoomed-out MNLP performance for held-out task 0.
		The deterministic embedding GP performs worse than the other methods by orders of magnitude.
	}
	\label{fig:examples:regression:additive:mnlp-no-zoom}
\end{figure}

Additionally, we see again that the deterministic embedding consistently results in very poor performance in terms of MNLP, frequently performing so much worse that it is impractical to show on the same axes as the Bayesian GP and Bayesian EGP models.
Figure \ref{fig:examples:regression:additive:mnlp-no-zoom} shows the results when zoomed out to view the performance of the deterministic EGP for one task.
We see a similar deficiency in all of the other held-out tasks, firmly underscoring that it is critical to account for uncertainty in the embedding in order to obtain credible predictions.
This makes sense, given the high number of tasks we must embed relative to the number of data available to elucidate their input-output characteristics.
This is not uncommon in engineering design problems, where one typically has a limited computational budget to explore any given design problem, but may potentially have access to a large archive of legacy tasks.

\subsection{Optimization results}
\begin{figure}[hbt]
	\centering
	\includegraphics[width=0.45\textwidth]{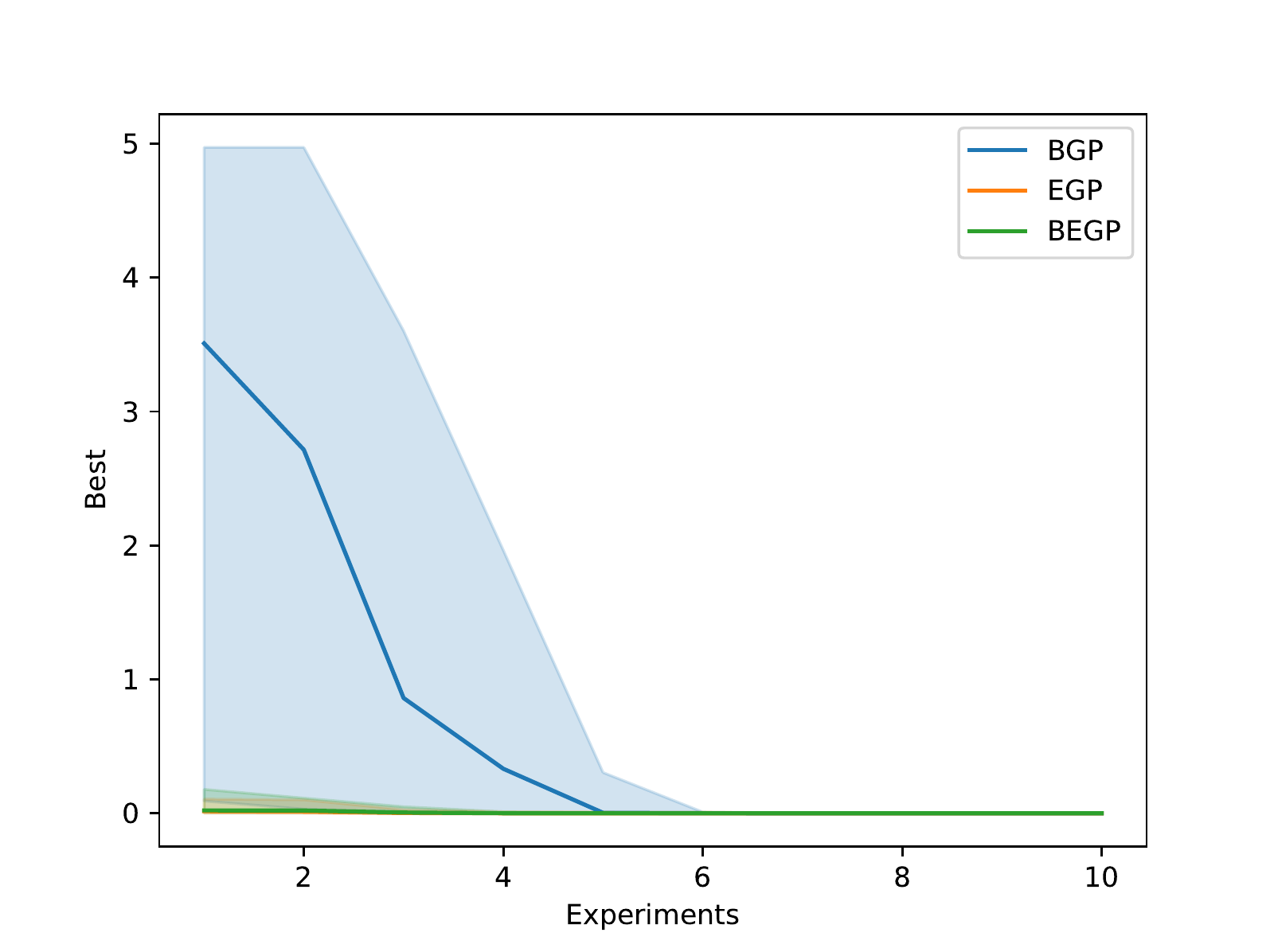}
	\includegraphics[width=0.45\textwidth]{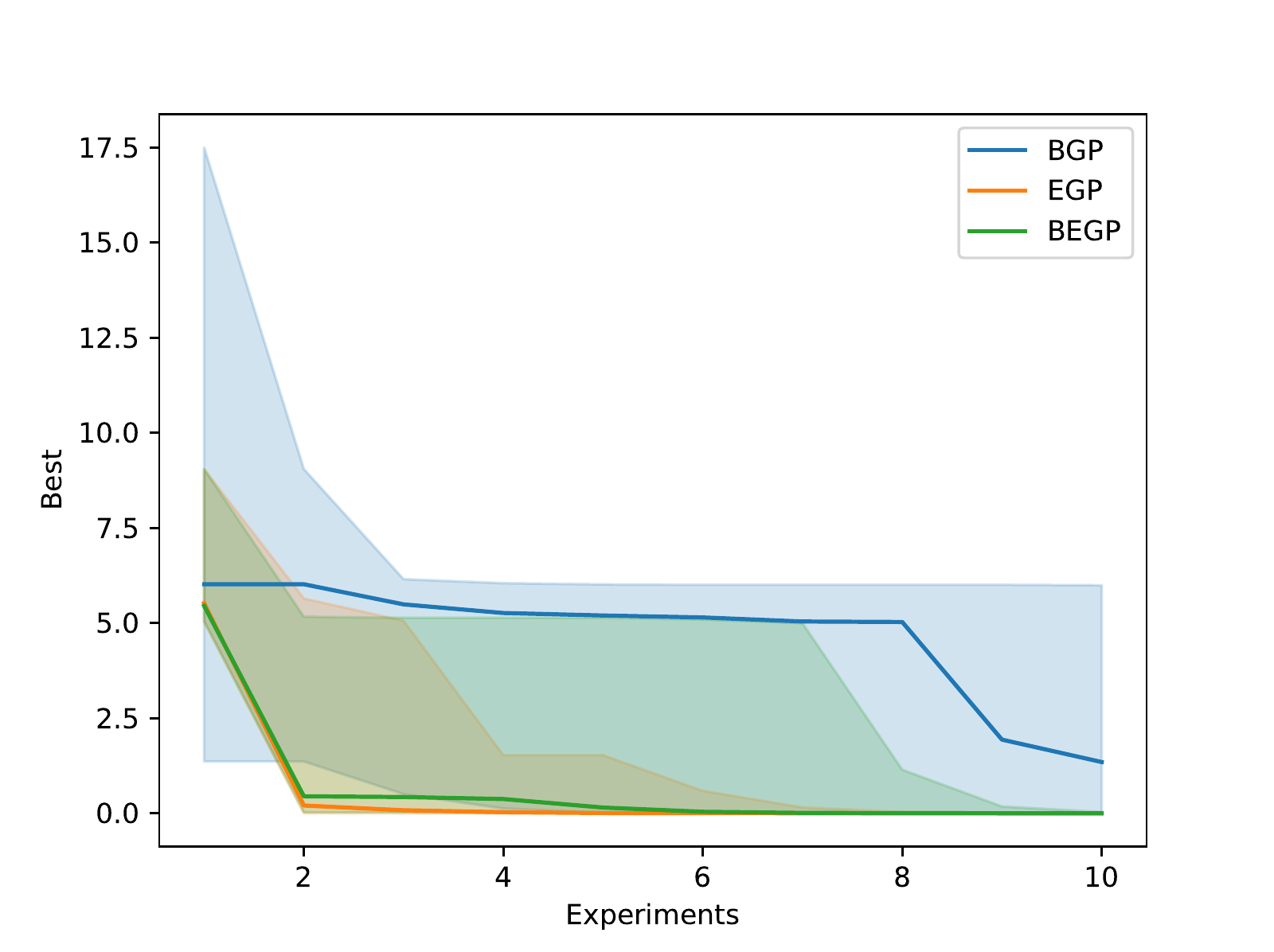}
	\caption{
		Bayesian optimization for the synthetic and Forrester systems.
		The solid curves show the median performance over $10$ repeats, and the shaded region shows the 80\% coverage interval over all repeats.
	}
	\label{fig:examples:optimization:synthetic:running_best}
\end{figure}

Figure \ref{fig:examples:optimization:synthetic:running_best} shows the best design discovered for the synthetic systems as a function of the number of evaluations on the current task.
We report the median performance over $10$ splits as well as the 80\% coverage interval ($10$-th to $90$-th percentile).
We see that the legacy task data enables the embedding GPs to consistently find near-optimal designs with a either a single evaluation (Forrester) or \textit{without any data from the current task} (synthetic).
By contrast, the GP usually requires a handful of evaluations before it begins to find good solutions.

\begin{figure}[hbt]
	\centering
	\includegraphics[width=0.3\textwidth]{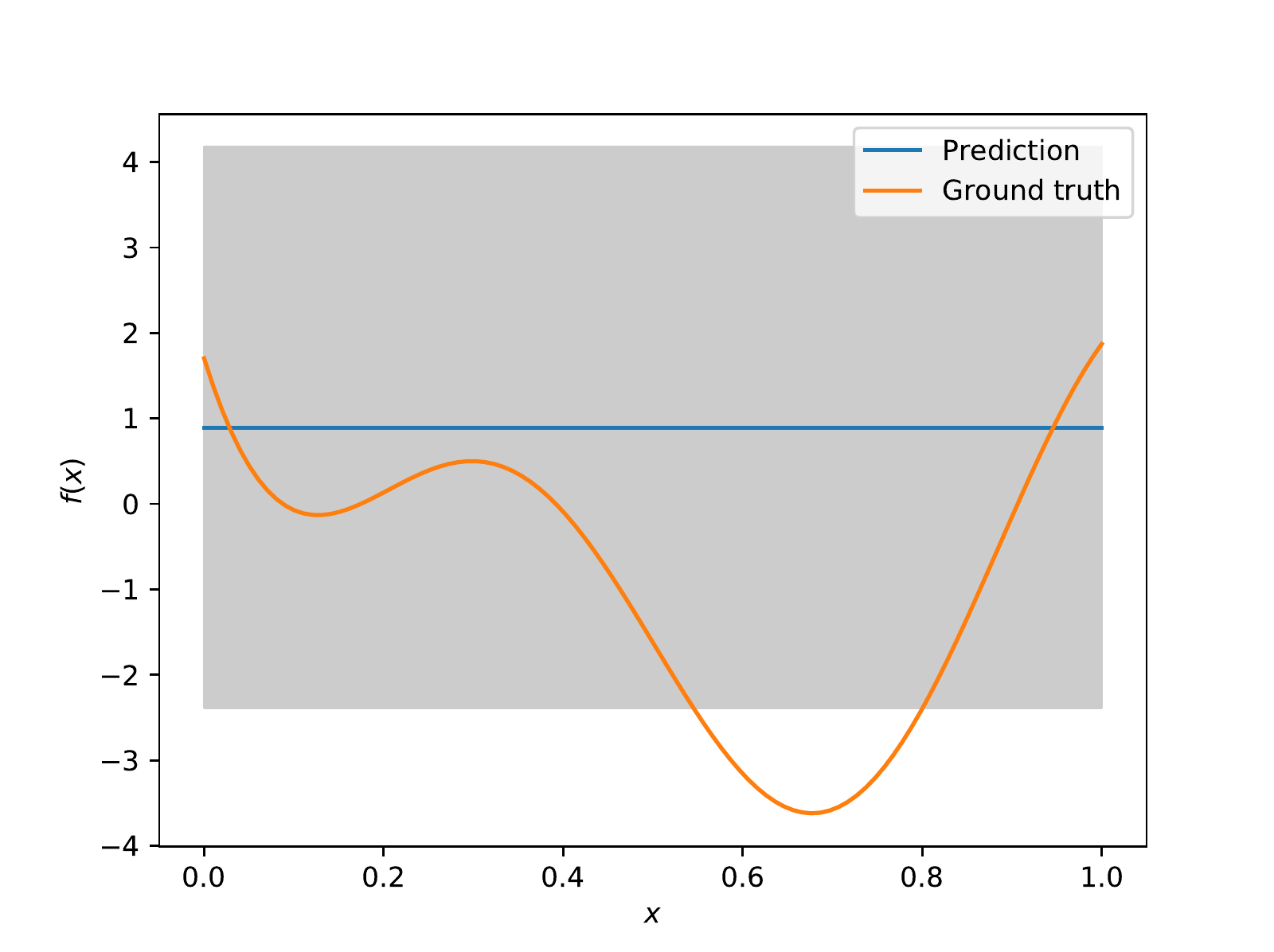}
	\includegraphics[width=0.3\textwidth]{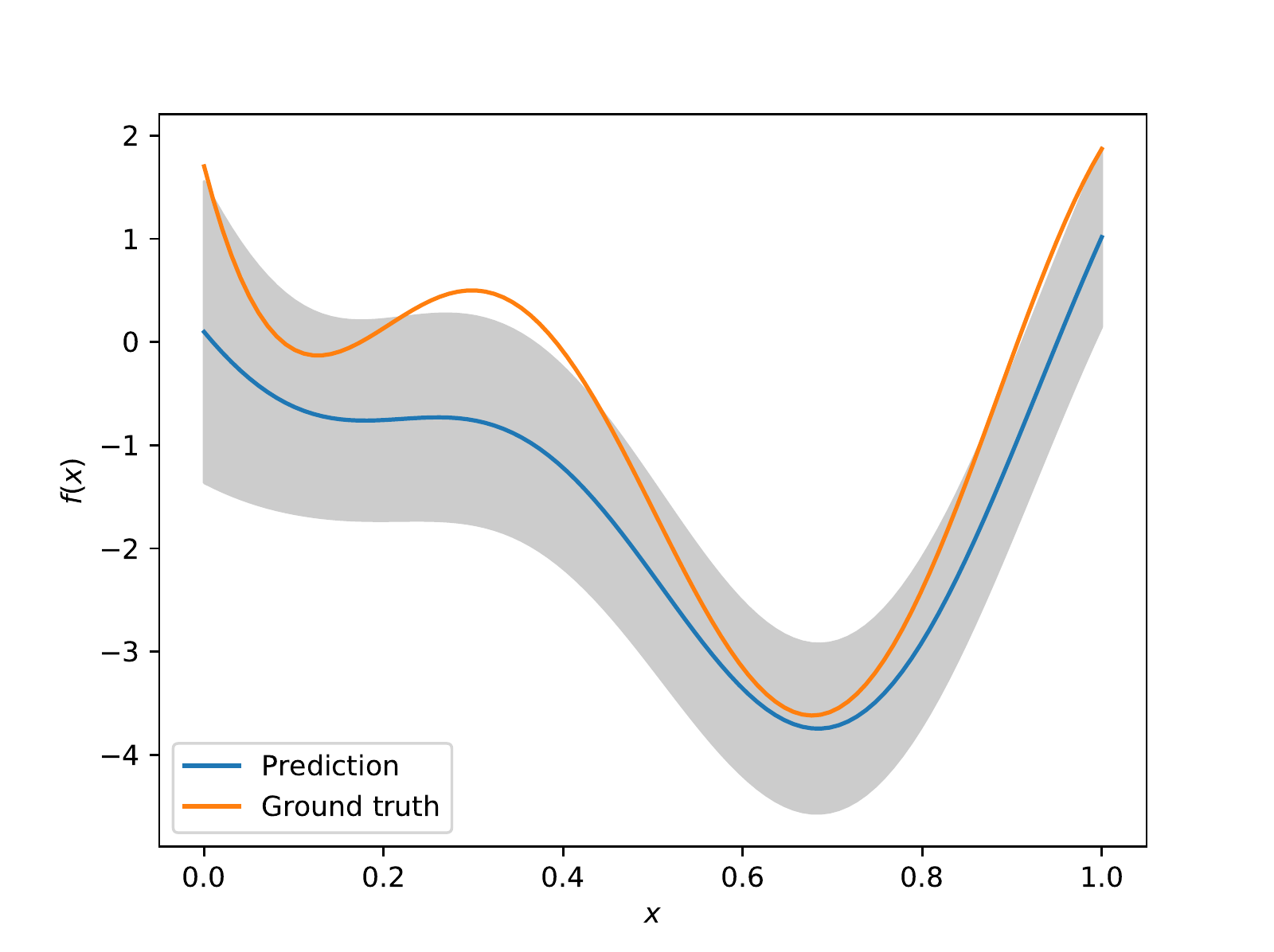}
	\includegraphics[width=0.3\textwidth]{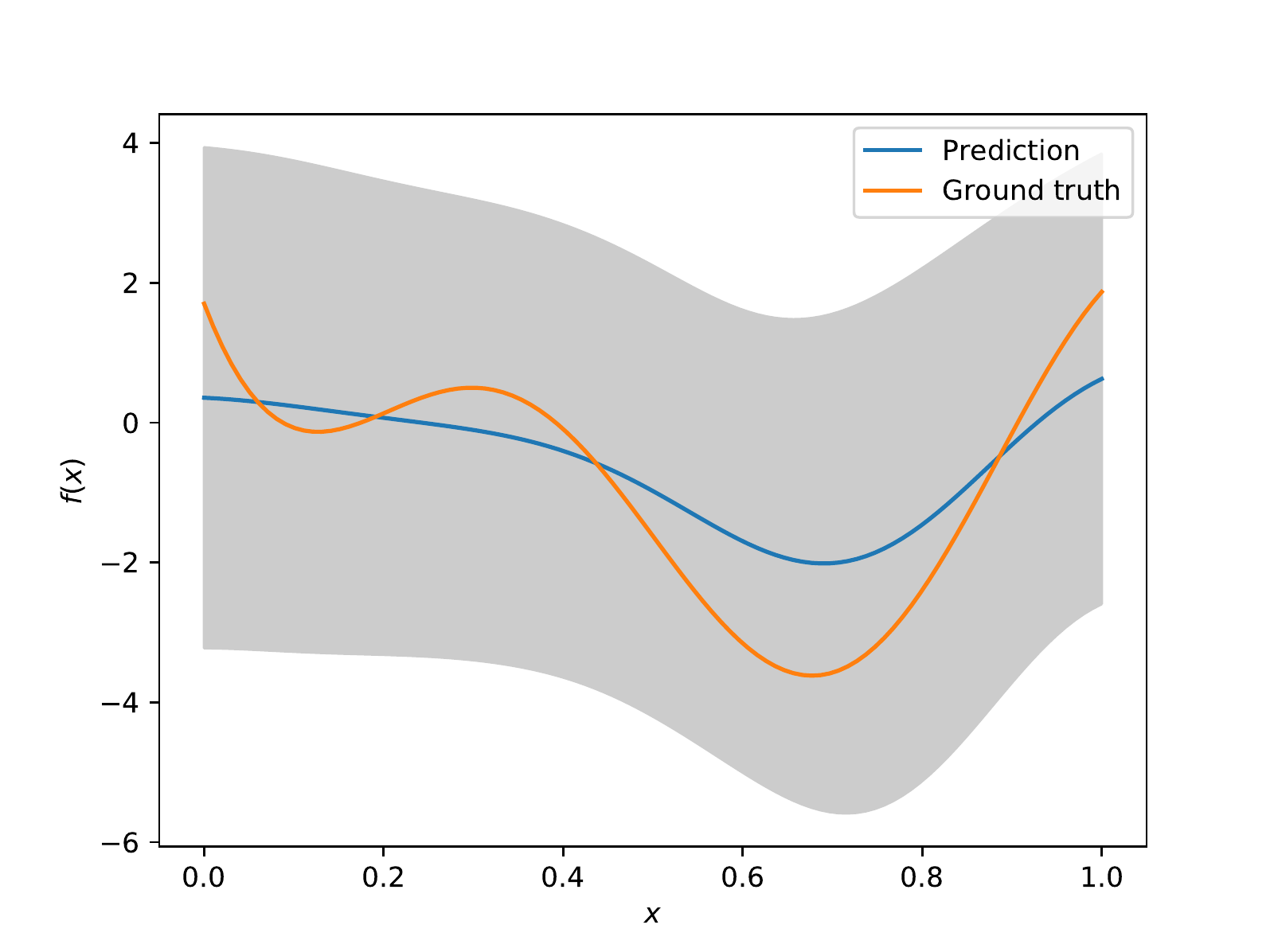}
	\\
	\includegraphics[width=0.3\textwidth]{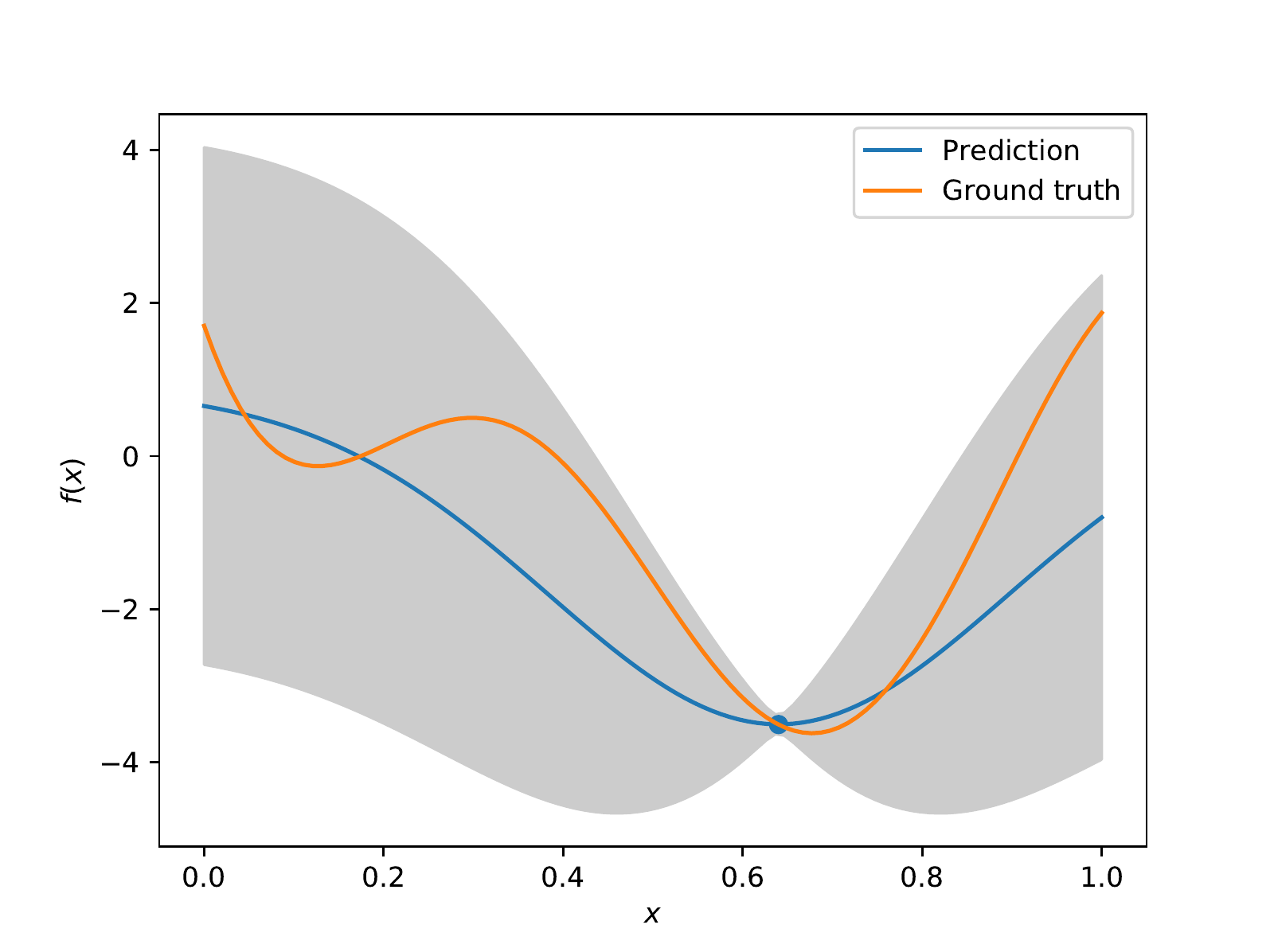}
	\includegraphics[width=0.3\textwidth]{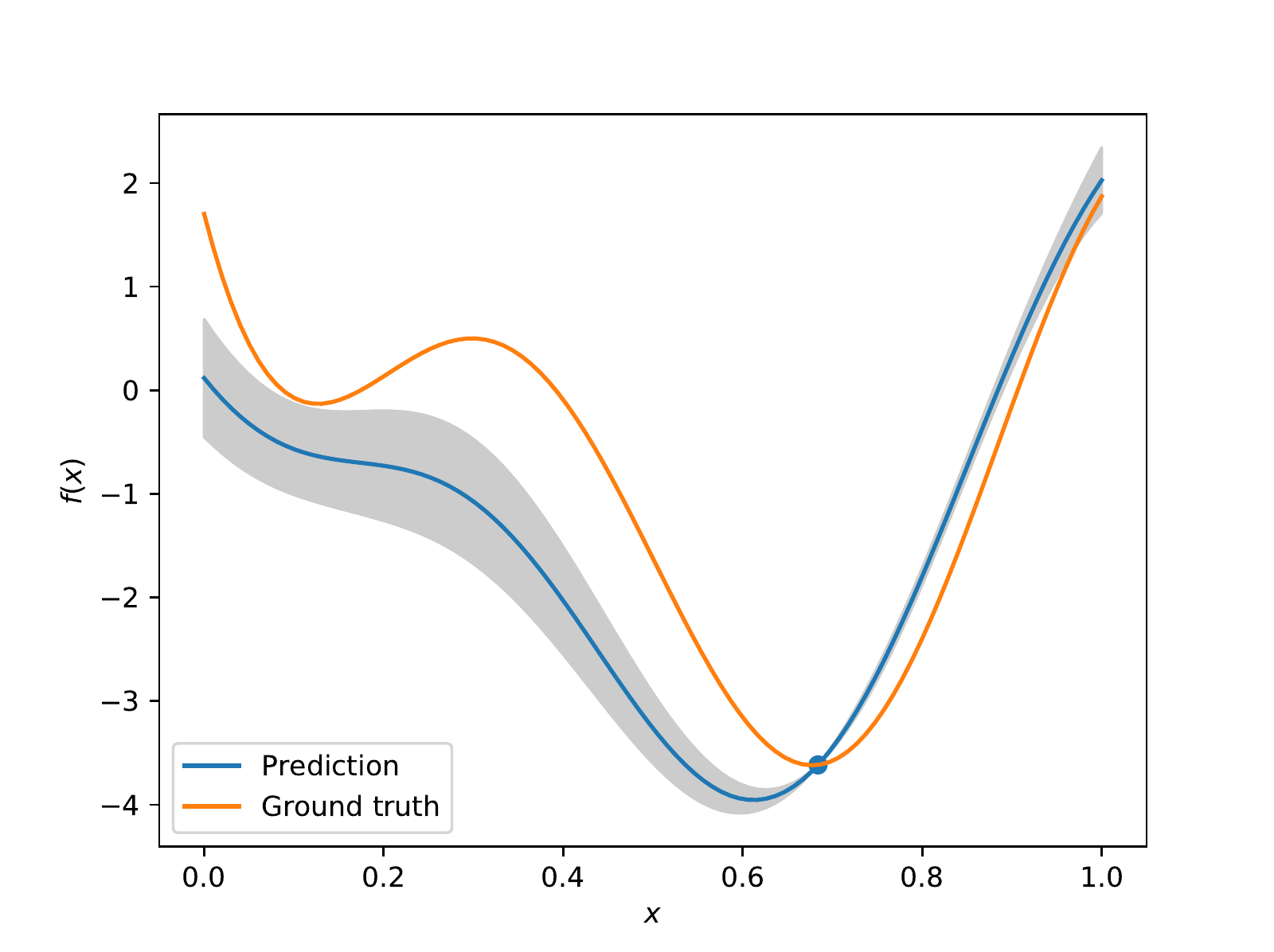}
	\includegraphics[width=0.3\textwidth]{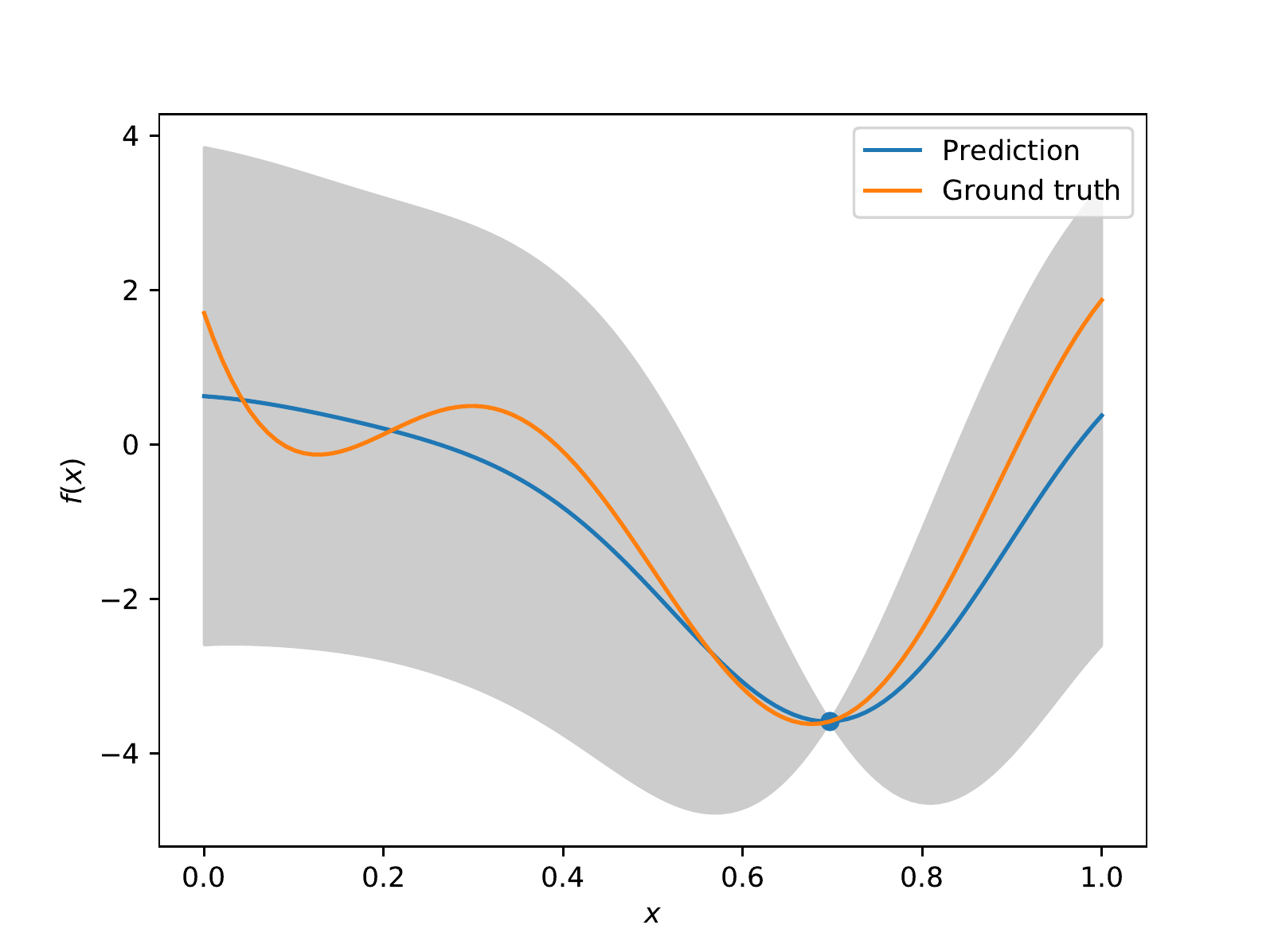}
	\caption{
		Predictive posteriors for the metamodels on the toy system.
		Top row: zero-shot predictions.
		Bottom row: one-shot predictions.
		From left to right: BGP, EGP, BEGP metamodels.
	}
	\label{fig:examples:optimization:synthetic:posterior}
\end{figure}

\begin{figure}[hbt]
	\centering
	\includegraphics[width=0.3\textwidth]{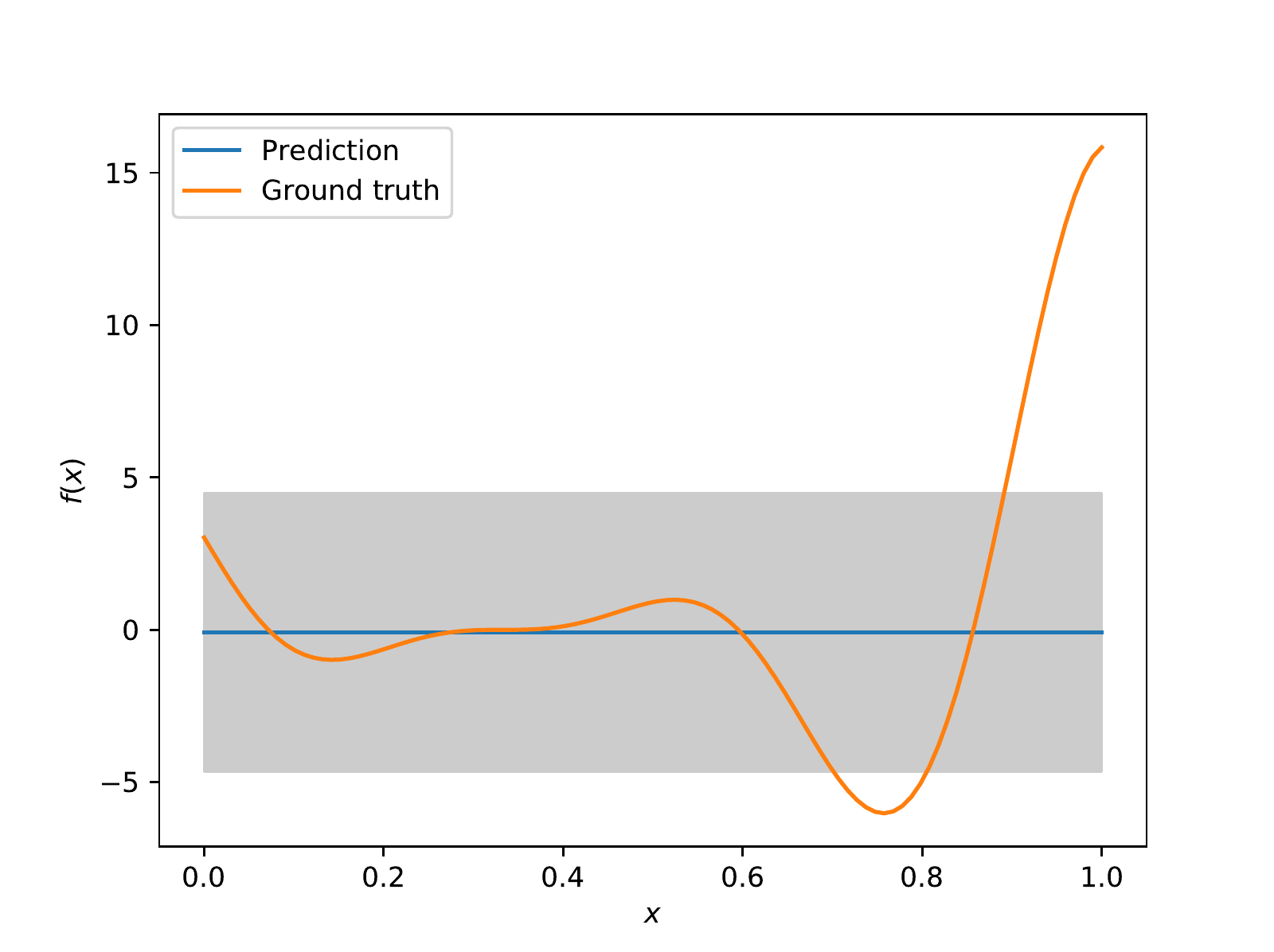}
	\includegraphics[width=0.3\textwidth]{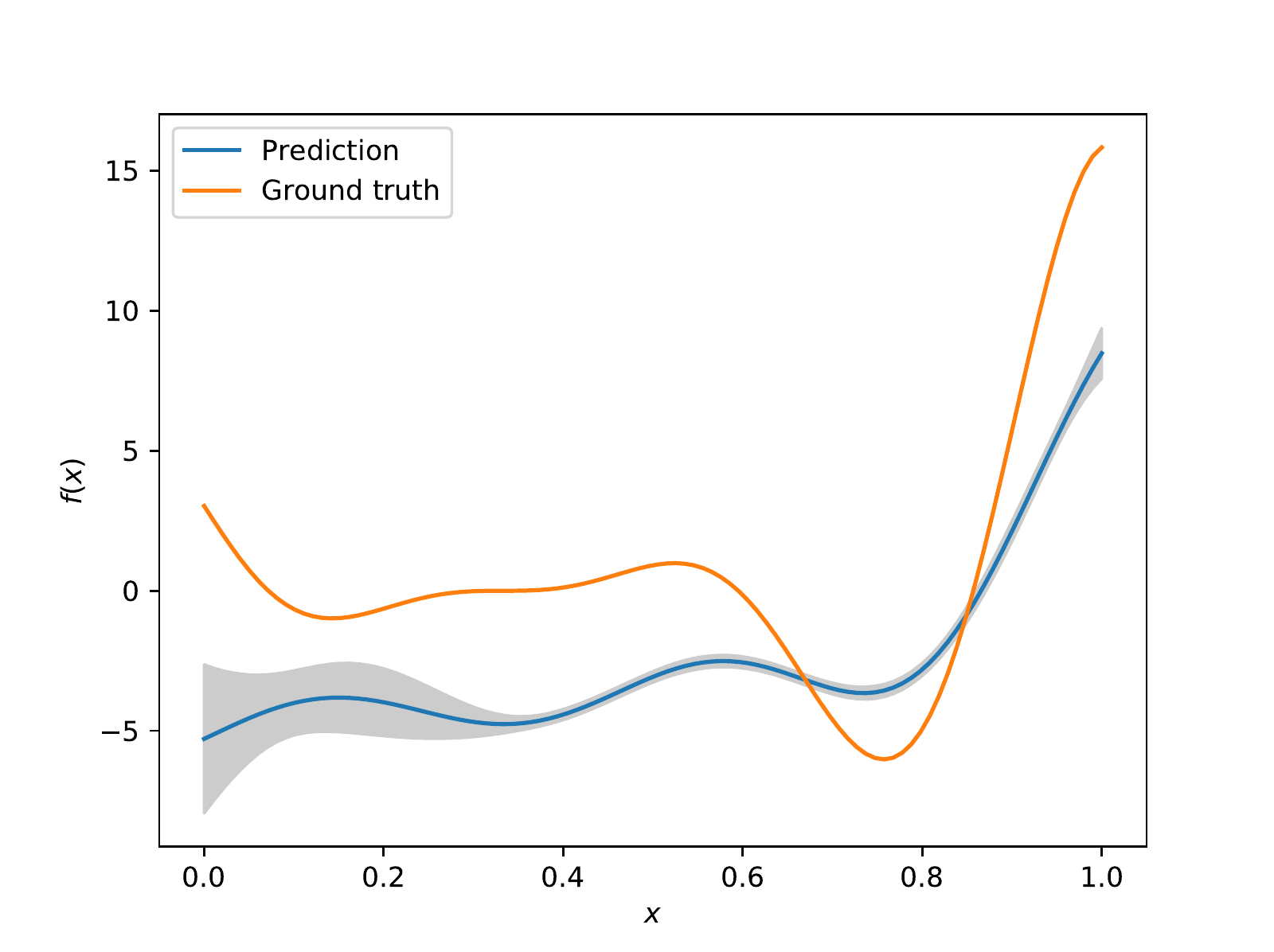}
	\includegraphics[width=0.3\textwidth]{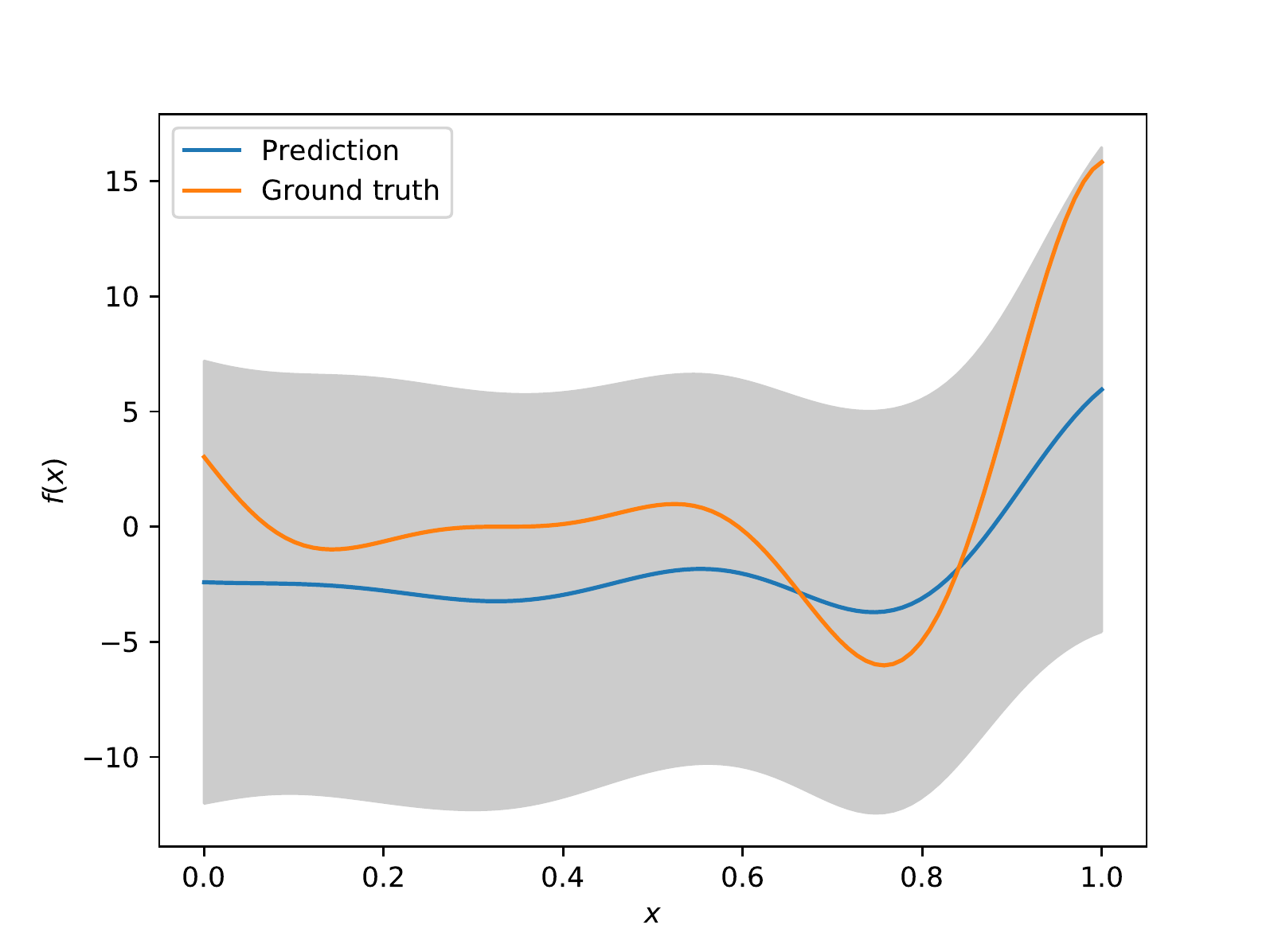}
	\\
	\includegraphics[width=0.3\textwidth]{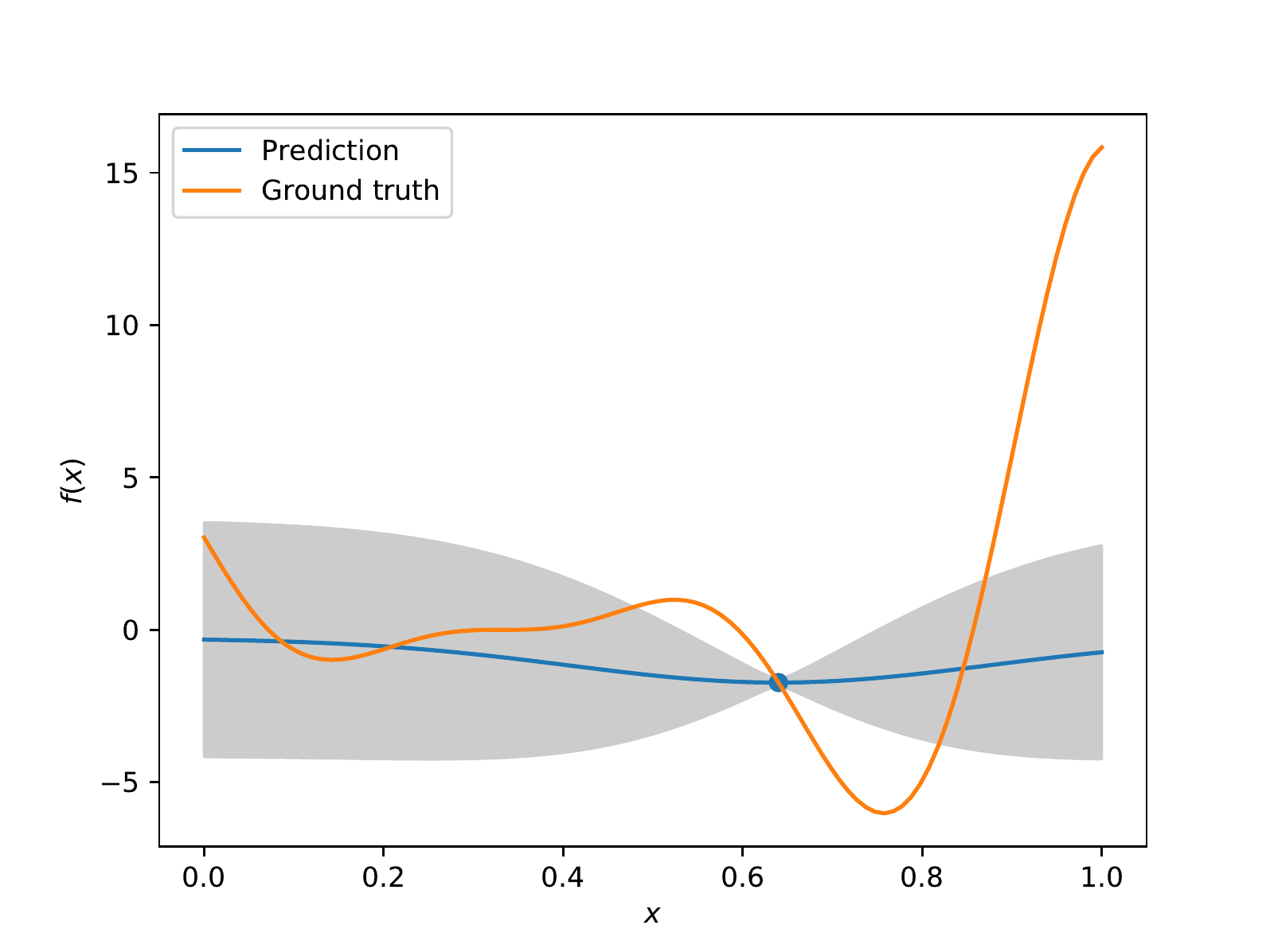}
	\includegraphics[width=0.3\textwidth]{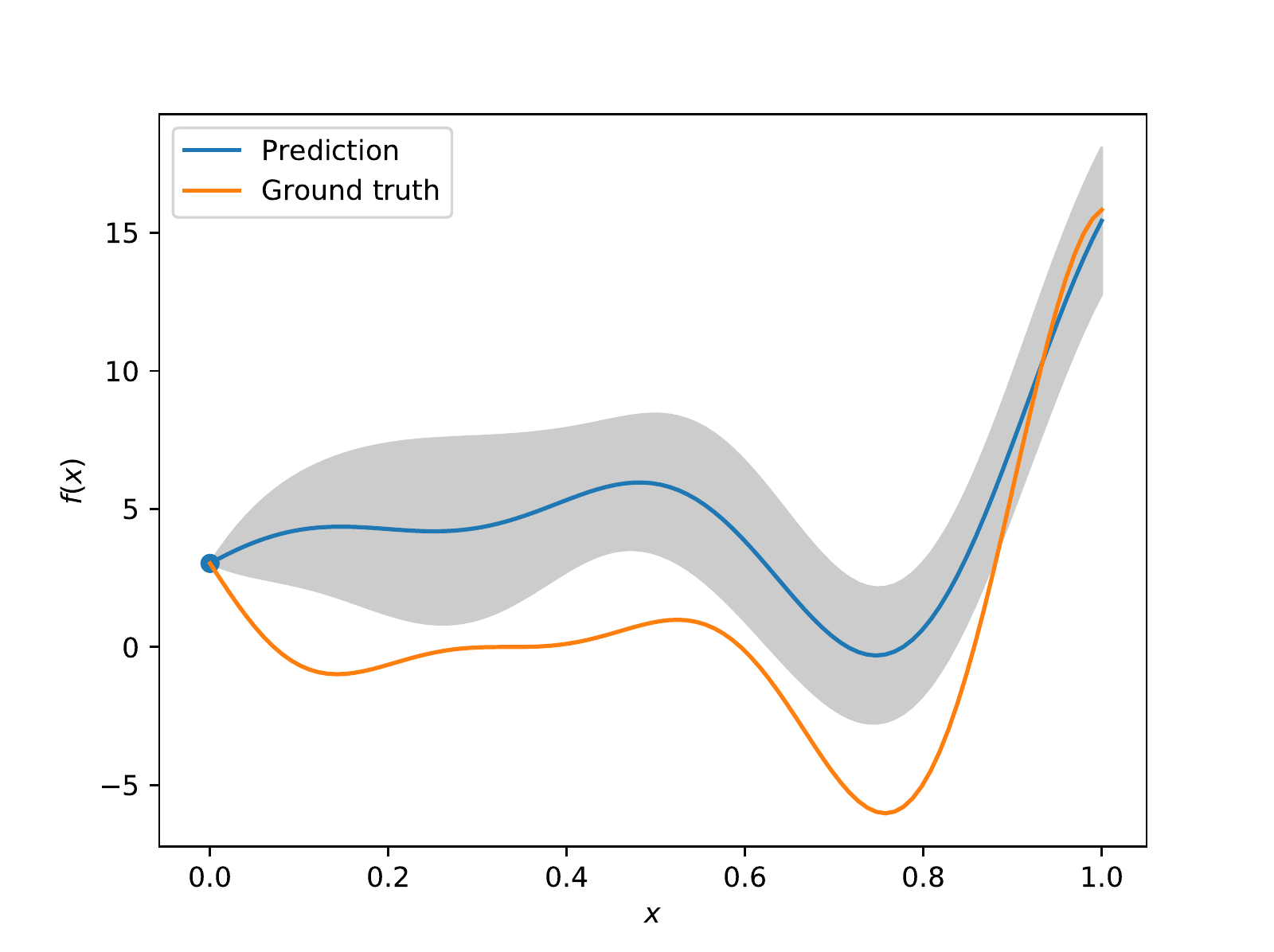}
	\includegraphics[width=0.3\textwidth]{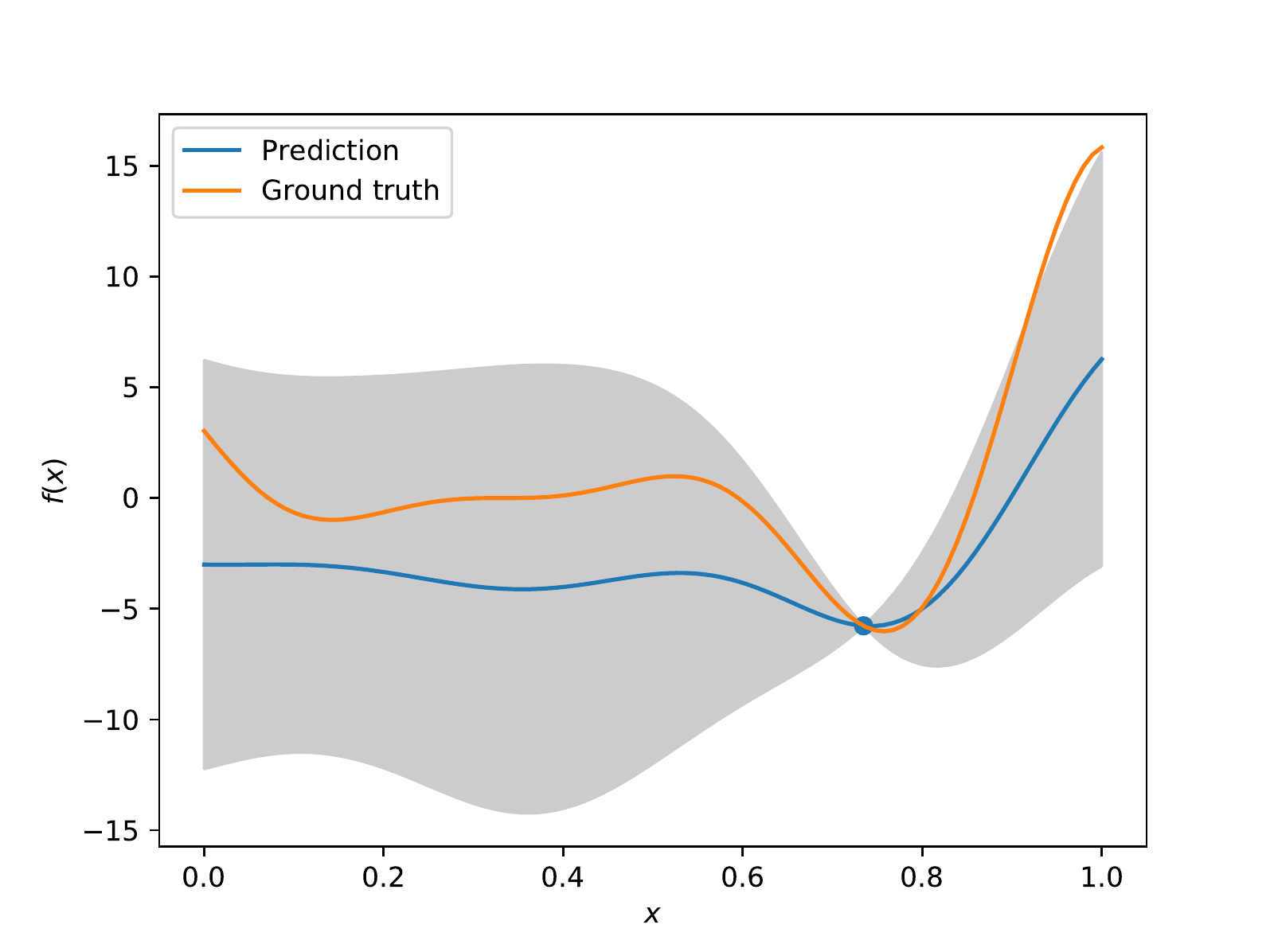}
	\caption{
		Predictive posteriors for the metamodels on the Forrester system.
		Top row: zero-shot predictions.
		Bottom row: one-shot predictions.
		From left to right: GP, EGP, BEGP metamodels.
	}
	\label{fig:examples:optimization:forrester:posterior}
\end{figure}

Figures \ref{fig:examples:optimization:synthetic:posterior} and \ref{fig:examples:optimization:forrester:posterior} show the zero- and one-shot posteriors of all three methods on the toy and Forrester systems, respectively.
We see that the legacy task data equip the embedding GPs with helpful priors that aid them in identifying the response surface.
By contrast, the Bayesian GP guesses randomly in the zero-shot setting and has minimal insight for its one-shot predictions.
We also notice that while the deterministic embedding model seems to fortunately pick good designs, its model of the response surface is highly overconfident.
By contrast, we see that the Bayesian embedding endows our models with well-calibrated uncertainty estimates that simultaneously allows is to find good designs while retaining a credible metamodel.
Thus, while the EGP performs comparably to the BEGP, we must be cautious about generalizing these results.
We also point out that while it seems like the BGP has picked a very good first point in Fig.\ \ref{fig:examples:optimization:synthetic:posterior}, this selection is by pure luck since BGP's predictive distribution with zero training data (i.e.\ the prior) is flat in its inputs.  Indeed, we see that it picks the same first point for the Forrester function in Fig.\ \ref{fig:examples:optimization:forrester:posterior}, which is not nearly so close to optimal.

\begin{figure}[hbt]
	\centering
	\includegraphics[width=0.45\textwidth]{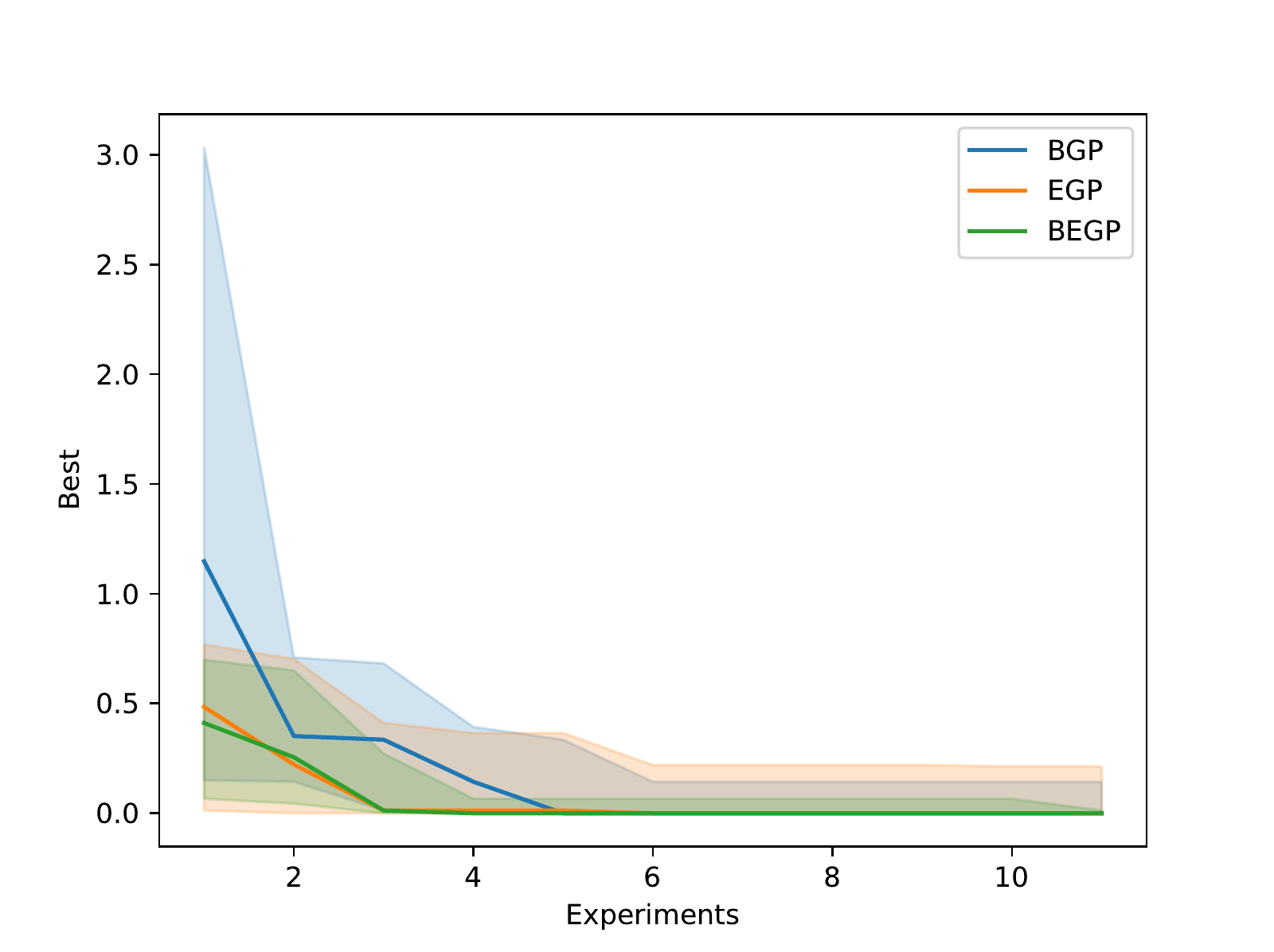}
	\includegraphics[width=0.45\textwidth]{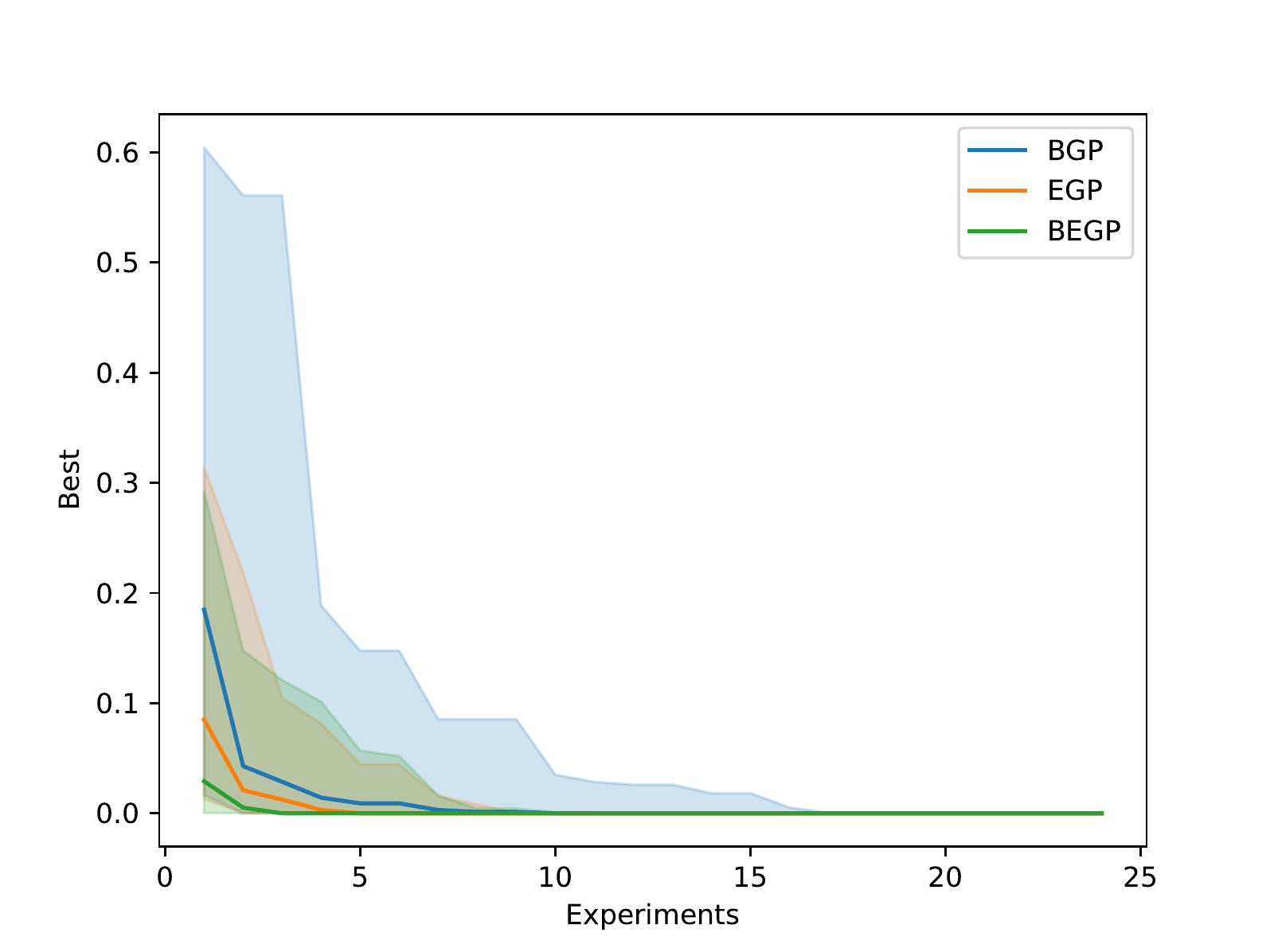}
	\caption{
		Bayesian optimization for the pump and additive systems.
		Results are shown relative to the per-task best design.
		The solid curves show the median performance over all repeats, and the shaded region shows the 80\% coverage interval.
	}
	\label{fig:examples:optimization:real:running_best}
\end{figure}

Figure \ref{fig:examples:optimization:real:running_best} shows the running best design for the pump and additive systems as a function of the number of current task evaluations.
Since each legacy system has a different optimum, results show the performance relative to the best design for the current task.
Again, we see that the embedding GPs vastly outperform the vanilla GP approach, usually finding good designs on their first attempts.
However, here we see that the Bayesian embedding gives better results over a deterministic embedding; this can be directly attributed to our earlier observation that the deterministic embedding tends to overfit, particularly when there are very few data to inform the embedding.
By contrast, the Bayesian embedding safely guards against overfitting and provides robust performance.

\section{Conclusion}
\label{sec:conclusion}
We have introduced a method for Bayesian optimization applicable to settings where one has access to a variety of related datasets but does not know how they relate to each other \textit{a priori}.
The Bayesian embedding GP automatically learns latent representations that enable a single metamodel to share information across multiple tasks, thereby improving the predictive performance in all cases, particularly in novel settings where limited data is available.
We observe that our method enables Bayesian optimization to more quickly and reliably find good solutions compared to traditional methods.

We find that variational inference with rudimentary variational approximations (factorized Gaussians) over the latent variables provides sufficient uncertainty information that credible estimates may be obtained in practice and is robust with respect to the number of latent dimensions.
Our experiments show that this Bayesian embedding is critical for obtaining credible uncertainty quantification.

There are a number of ways in which our work might be extended.
For one, our method might be applied to optimization settings with unknown constraints.
In such cases, one might simultaneously model the probability of violating a constraint with a multi-task model.
Modeling binary outputs such as constraint violation with GPs requires non-conjugate inference, which can be enabled using techniques from \cite{hensman2015scalable}.
Second, our approach is straightforward to apply to problems of robust design.
In fact, we have already shown how uncertainty in our metamodel can be propagated efficiently during design selection through stochastic backpropagation, eliminating the need for expensive double-loop optimization.
Other approaches \cite{ryan2018gaussian, ling2018efficient} exploit analytic properties to solve the inner-loop optimization, though our work suggests that it might be approximately solved in general using stochastic optimization.
Finally, our model learns a single embedding for each general input (task) as a latent vector.
We would like to explore the effect of a hierarchical latent representation in which the embedding of each task is allowed to vary with $\vc x^{(r)}$.
Such a conditional multi-task model might enable one to exploit partial similarities between tasks in certain regions of input space while still providing for the possibility that their trends might differ elsewhere.
While this was studied in \cite{ghosh2018bayesian}, we expect that our Bayesian latent representation might improve performance in few-shot settings such as those considered in this work.

\section*{Funding Sources}
Funding for this work was provided internally by GE Research.

\clearpage
% Added from bbl:

\end{document}